\documentclass{article} 
\usepackage{iclr2027_conference,times}

\usepackage{amsmath,amsfonts,bm}

\def\eqref#1{equation~\ref{#1}}

\def\1{\bm{1}}

\DeclareMathAlphabet{\mathsfit}{\encodingdefault}{\sfdefault}{m}{sl}
\SetMathAlphabet{\mathsfit}{bold}{\encodingdefault}{\sfdefault}{bx}{n}

\usepackage[table]{xcolor}
\definecolor{gain}{RGB}{0,120,40}\definecolor{cost}{RGB}{190,30,30}
\usepackage{hyperref}
\usepackage{url}
\usepackage{graphicx}
\usepackage{cleveref}
\usepackage{wrapfig}
\iclrfinalcopy 

\title{Mind the RefGAP: Correcting Reference Attention in Diffusion-Based Visual Editing}

\author{
Yanan Wang\textsuperscript{1,2} \quad
Shengcai Liao\textsuperscript{3} \quad
Guangyi Liu\textsuperscript{1,2} \quad
Xiaodan Liang\textsuperscript{1}%
\thanks{Emails: \texttt{\{yanan.wang, guangyi.liu, xiaodan.liang\}@mbzuai.ac.ae}, \texttt{scliao@uaeu.ac.ae}}
\\[1mm]
{\normalfont \textsuperscript{1}Mohamed bin Zayed University of Artificial Intelligence \quad
\textsuperscript{2}Institute of Foundation Models}
\\
{\normalfont \textsuperscript{3}United Arab Emirates University}
\\[1mm]
{\normalfont\small Project page: \href{https://yanan-wang-cs.github.io/RefGAP/}{\texttt{https://yanan-wang-cs.github.io/RefGAP/}}}
}

\newcommand{\gap}{RefGAP\xspace}
\usepackage{xspace}
\usepackage{booktabs,multirow,makecell}
\usepackage{pifont}
\usepackage{float}
\begin{document}

\maketitle
\begin{abstract}
    Reference-guided diffusion editors struggle to faithfully reproduce user-provided references. We observe that many diffusion editors allocate limited attention to the reference. For example, in LoomVideo, edit-region queries assign less than $1\%$ of their attention mass to the reference. We introduce RefGAP, a training-free correction that determines logit-offset magnitudes online at each layer from the reference-attention mass measured during the forward pass. Positive offsets to reference logits strengthen reference usage by edit-region queries, while negative offsets for keep-region queries limit reference-induced changes outside the edit. Two global coefficients control the correction; they are selected once on validation data from four development diffusion editors and held fixed. Across seven diffusion-based image/video editors, RefGAP improves identity fidelity in head swapping and face swapping, where identity takes priority; on some editors this raises pose error, which delaying the correction reduces. RefGAP achieves a fidelity-preservation trade-off comparable to separately tuned constant edit-side biases, without per-approach strength sweeps. Additional experiments on virtual try-on and background replacement evaluate transfer beyond identity editing.
        \end{abstract}
\section{Introduction}
\label{sec:intro}
Modern diffusion editors can condition on a user-provided reference image to edit a driving image or video, supporting tasks such as head swapping, face swapping, virtual try-on, and background replacement. Many recent systems adopt an \emph{in-context} design~\citep{vace,qwenimageedit2509,wu2025omnigen2,flux-2-2025}, in which reference tokens share the denoiser's attention sequence with the content being edited. Yet direct access to the reference does not guarantee fidelity.

Existing methods strengthen reference conditioning through task-specific training~\citep{wang2025directswap,simswapplusplus,fang2024vivid} or training-free manipulation of reference attention or features~\citep{shin2025large,fan2024refdrop}. Training-free interventions typically rely on fixed, empirically chosen coefficients that may require retuning across models or tasks. SaaS~\citep{zhou2025saas} sets its scale online, but it rescales text tokens, derives the scale from a separate group of input-image tokens, which some in-context editors do not provide (e.g., VACE and LoomVideo), and has no keep-region term. Region-aware methods either rely on trained modules~\citep{chen2024anyscene,zhang2025boow} or copy the source background~\citep{zhu2025kvedit}. None of these sets reference-attention strength from the reference mass alone, and none couples reference-attention enhancement in edit regions with suppression in keep regions.

We partition each attention layer's keys into reference and rest tokens and compute the fraction $\rho$ of each query's attention mass assigned to the reference (\cref{fig:rho_profiles}). Using an edit mask, we average $\rho$ separately over content queries in the edit and keep regions. The profiles reveal two patterns: some approaches exhibit substantially higher reference attention in edit versus keep regions within a localized layer band, typically in the network's second half; others assign little reference attention to edit queries across many layers, falling below $1\%$ on LoomVideo~\citep{wu2026loomvideo}. These patterns motivate selective layer intervention, while $\rho$ provides an online signal for correction strength.

Based on this measurement, we introduce \gap, a training-free correction named after the reference-attention gap it aims to bridge. Within a single forward pass, using statistics already produced by the attention computation, \gap reads the reference-attention masses in both regions and adds one scalar to every reference-key logit: a positive offset for edit-region queries, so they read the reference more, and a negative offset for keep-region queries, so the stronger reference signal does not pull the preserved content with it. The offset adapts to the measured reference-attention mass at each intervened layer and forward pass, reducing the need for model- or task-specific tuning.

We calibrate two global coefficients and establish a shared layer-selection rule on four development approaches (JoyAI-Video-Edit~\citep{xiao2026joyai}, LoomVideo~\citep{wu2026loomvideo}, FLUX.2-klein and FLUX.2-klein-base~\citep{flux-2-2025}) using 40 head-swap validation clips. We freeze the coefficients, thresholds, and rule before applying them to three held-out approaches. Each approach requires one validation profiling run to determine its layer set under the fixed rule; this set is reused across tasks without tuning. Identity similarity improves on head-swap and face-swap tasks. Without task-specific retuning, virtual try-on fidelity improves across seven approaches, while background replacement yields mixed results, including a clear degradation on Qwen-Image-Edit (\cref{sec:limitations}).

\noindent To summarize, our contributions are:
\begin{itemize}
    \item We measure reference-attention mass for edit and keep queries, revealing approach-dependent allocation patterns and limited reference attention in several approaches.
    \item We propose \gap, a training-free correction of reference-key logits that amplifies reference attention for edit queries and suppresses it for keep queries, with per-layer strength set online from the measured reference mass.
    \item We demonstrate cross-approach and cross-task transfer using a single configuration calibrated once on 40 head-swap validation clips. Without retuning, it consistently boosts identity fidelity across seven approaches in head and face swapping (alongside an attribute-preservation trade-off), transfers to virtual try-on, and delineates transfer boundaries with mixed results on background replacement.
\end{itemize}

\section{Related Work}
\label{sec:related}

\paragraph{Reference-conditioned generation and editing.}
Reference-guided editors incorporate visual conditions either as tokens in a shared attention sequence or through a separate conditioning branch. In-context systems such as VACE~\citep{vace}, LoomVideo~\citep{wu2026loomvideo}, JoyAI-Video-Edit~\citep{xiao2026joyai}, Qwen-Image-Edit~\citep{qwenimageedit2509}, OmniGen2~\citep{wu2025omnigen2}, and FLUX.2-klein~\citep{flux-2-2025} allow reference and content tokens to interact within joint attention. Decoupled methods inject reference features through cross-attention or additive branches, as in IP-Adapter~\citep{ye2023ip}. \gap targets the in-context setting, where reference and content tokens compete within the same softmax and their relative mass can be modified directly.

\paragraph{Training-free reference and attention steering.}
Training-free diffusion methods manipulate attention to control conditioning and source consistency~\citep{hertz2022prompt,cao2023masactrl}. Diptych Prompting~\citep{shin2025large} enhances attention from queries in the generated panel to keys in the reference panel. FreeCustom~\citep{ding2024freecustom} combines multi-reference self-attention with weighted concept masks to integrate and emphasize reference concepts for multi-concept composition. RefDrop~\citep{fan2024refdrop} controls consistency across generated images or video frames by mixing reference and self-attention outputs with a user-specified scalar shared across queries.

These methods control reference influence through fixed or user-specified coefficients selected for a particular model or task. FreeCustom and RefDrop primarily target generation or consistency, without defining a region to preserve in an existing source image. Diptych also supports subject-driven editing through an edit mask, but its reference-attention boost remains uniform across queries in the edited panel. GRAG~\citep{zhang2026group} and DCAG~\citep{li2026dual} provide separate representation-level controls for editing strength through externally selected coefficients. KV-Edit~\citep{zhu2025kvedit} preserves the background by replacing its keys and values with those of the source, and SaaS~\citep{zhou2025saas} rescales attention to text sub-instructions inside the edit region to match attention on the input image, without a keep-region term. AnyScene~\citep{chen2024anyscene} injects the foreground through a trained ControlNet-style module and, at inference, multiplies the text cross-attention of foreground words by the binary foreground mask; BooW-VTON~\citep{zhang2025boow} learns region control through a training loss. Minimal-change learning~\citep{chen2026scalable} instead fine-tunes an instruction-based editor with reinforcement learning, using VLM rewards that penalize unintended changes outside the instructed edit; \gap pursues this locality for reference-guided editing without training, by suppressing reference attention in keep regions.

\gap determines logit-offset magnitudes online from reference attention measured separately for edit and keep queries. Positive offsets enhance reference attention in edit regions, while negative offsets suppress it in keep regions. After a single calibration, the same coefficients transfer to held-out approaches and additional tasks without further manual tuning.

\section{Method}
\label{sec:method}

\begin{figure*}[t]
    \centering
    \includegraphics[width=\textwidth]{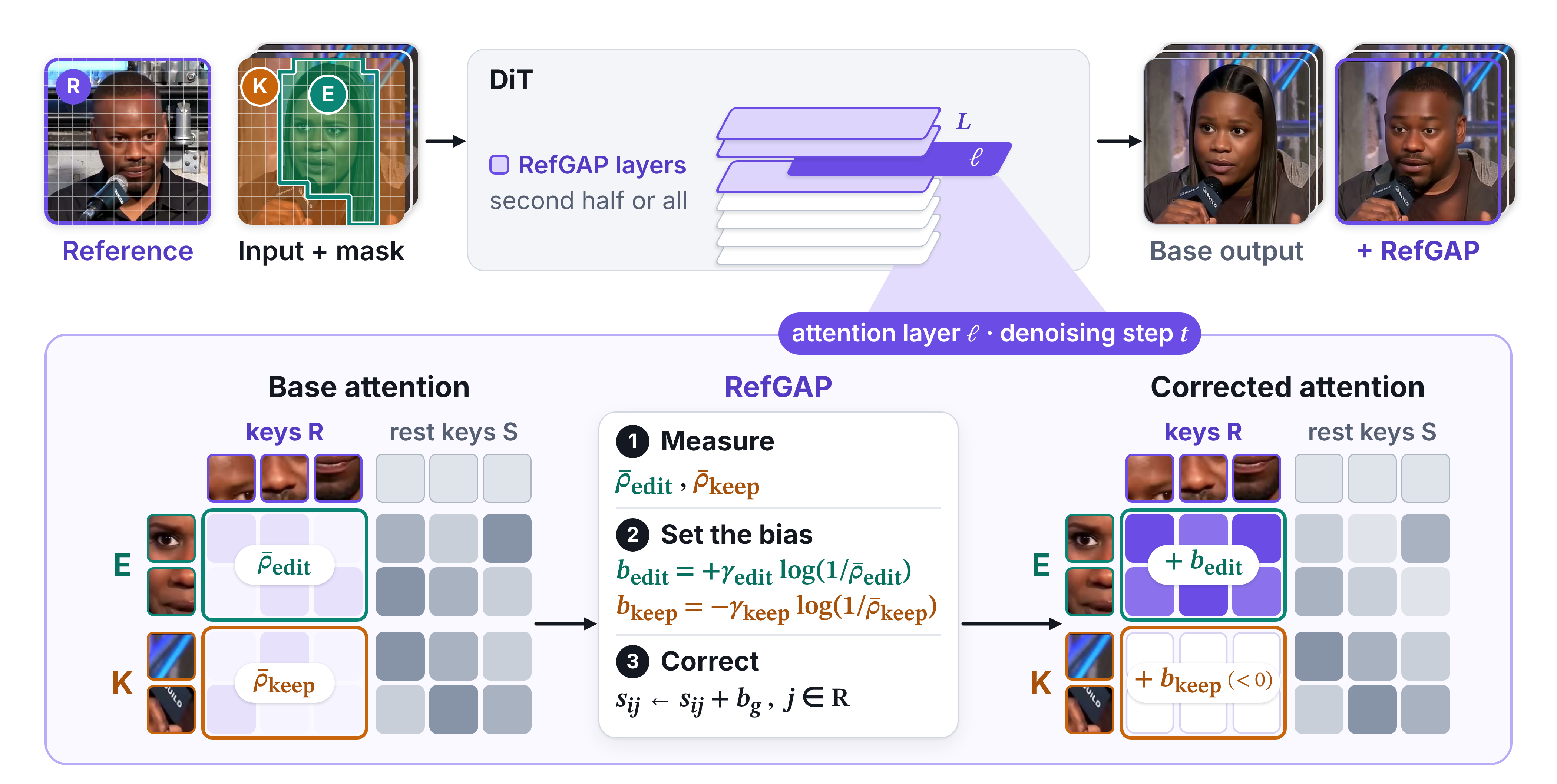}
    \caption{\textbf{\gap overview.} At each selected layer and denoising step, \gap (1) measures the mean reference-attention mass $\bar\rho_{\mathrm{edit}}$ and $\bar\rho_{\mathrm{keep}}$ of edit queries (E) and keep queries (K), (2) converts them into a positive offset for E and a negative offset for K, and (3) adds the offset to the reference-key logits, which raises $\rho_i$ for E and lowers it for K. Reference queries and rest keys are unchanged. The two coefficients are shared across approaches and tasks.}
    \label{fig:gap_overview}
\end{figure*}

\subsection{Overview}

\gap defines the masked area as the edit region and its complement as the keep region. At each attention layer and denoising step, it measures reference-attention mass in both regions (\cref{sec:rho}) to determine reference-logit shifts (\cref{sec:correction}), applied according to a layer-selection rule (\cref{sec:layer_rule}). \Cref{fig:gap_overview} summarizes this inference-time pipeline, which requires no weight updates.

\subsection{Reference attention mass}
\label{sec:rho}

Consider attention layer $\ell$ at denoising step $t$, with $H$ attention heads indexed by $h$. Let $s_{ij}^{(\ell,t,h)}$ denote the pre-softmax score between query $i$ and key $j$ in head $h$. We partition the keys into reference tokens $\mathcal{R}$ and all rest tokens $\mathcal{S}$. The \emph{reference attention mass} of a content query $i$ is the fraction of its attention allocated to the reference group:
\begin{equation}
    \rho_i^{(\ell,t,h)}=
    \frac{\sum_{j\in\mathcal{R}}\exp(s_{ij}^{(\ell,t,h)})}
    {\sum_{j\in\mathcal{R}\cup\mathcal{S}}\exp(s_{ij}^{(\ell,t,h)})},
    \qquad i \in S,
    \qquad \rho_i\in(0,1).
    \label{eq:rho}
\end{equation}

Let the edit mask, mapped to the latent token grid, partition content queries into an \emph{edit set} $\mathcal{E}$ of tokens to be modified and a \emph{keep set} $\mathcal{K}$ of tokens to be preserved. We average the per-head mass over the relevant queries and $H$ heads, yielding region-level statistics per layer and denoising step:
\begin{equation}
\bar\rho_{\mathrm{edit}}^{(\ell,t)}
=\frac{1}{|\mathcal{E}|H}\sum_{i\in\mathcal{E}}\sum_{h=1}^{H}
\rho_i^{(\ell,t,h)},
\qquad
\bar\rho_{\mathrm{keep}}^{(\ell,t)}
=\frac{1}{|\mathcal{K}|H}\sum_{i\in\mathcal{K}}\sum_{h=1}^{H}
\rho_i^{(\ell,t,h)}.
\label{eq:region_rho}
\end{equation}
The first quantity measures how strongly queries in the edit region attend to the reference, while the second measures reference influence on queries in the keep region. Each statistic is a scalar for a given input, layer, and denoising step, recomputed from the current denoising trajectory immediately before that layer's offset is applied. This online computation requires no additional base-model forward pass; the offline profiling used for layer selection is described in \cref{sec:layer_rule}.

\paragraph{Token-count-normalized diagnosis.}
To account for reference-token count, we normalize reference-attention mass by the share assigned under uniform attention:
\begin{equation}
\pi_{\mathcal R}
=\frac{|\mathcal R|}{|\mathcal R|+|\mathcal S|},
\qquad
u_g^{(\ell,t)}
=\frac{\bar\rho_g^{(\ell,t)}}{\pi_{\mathcal R}},
\quad g\in\{\mathrm{edit},\mathrm{keep}\}.
\label{eq:uniform_share}
\end{equation}
Values below one indicate less reference attention than uniform allocation. This is a diagnostic baseline, not a target for the correction.

We average base-model statistics over all layers, denoising steps, and head-swap validation clips. Edit-region reference attention falls below the baseline in five approaches, with VACE the exception (\cref{tab:normalized_diag}). JoyAI-Video-Edit is excluded because its reference-token share varies with the streaming KV cache. Qwen-Image-Edit and OmniGen2 assign more reference attention to keep than edit queries, motivating separate regional measurement and correction. Diagnostics appear in \cref{sec:diag}. We use reference mass to set the correction strength, not to predict failure or the size of the identity gain; VACE, for example, exceeds the uniform share yet gains $+.324$ identity (\cref{tab:main}).

\begin{table}[t]
\centering
\setlength{\tabcolsep}{4pt}
\resizebox{\linewidth}{!}{%
\begin{tabular}{lcccccc}
\toprule
& LoomVideo & FLUX.2-klein & FLUX.2-klein-base
& VACE & Qwen-Image-Edit & OmniGen2 \\
\midrule
$u_{\mathrm{edit}}$ & 0.22 & 0.35 & 0.22 & 1.79 & 0.30 & 0.24 \\
$u_{\mathrm{keep}}$ & 0.13 & 0.10 & 0.15 & 0.69 & 0.35 & 0.28 \\
\bottomrule
\end{tabular}%
}
\caption{\textbf{Token-count-normalized reference attention.}
Measured using head-swap validation data.}
\label{tab:normalized_diag}
\end{table}

\subsection{Two-sided reference correction}
\label{sec:correction}

To control reference-attention allocation, \gap adds a shared offset to the reference-key logits for each query group. This changes the total attention assigned to the reference while preserving the relative attention weights within each key group. The offset is positive for edit queries and negative for keep queries; its magnitude adapts to the corresponding measured reference mass. 

 For query $i$ in layer $\ell$ at step $t$, \gap adds a region-dependent offset to logits with keys in $\mathcal{R}$:
\begin{equation}
\tilde s_{ij}=s_{ij}+b_{\ell,t}^{(i)},
\qquad j\in\mathcal{R},
\label{eq:modified_logit}
\end{equation}
with
\begin{equation}
b_{\ell,t}^{(i)}=
\begin{cases}
+\gamma_{\mathrm{edit}}\log\!\left(1/\bar\rho_{\mathrm{edit}}^{(\ell,t)}\right),
    & i\in\mathcal{E},\\[2pt]
-\gamma_{\mathrm{keep}}\log\!\left(1/\bar\rho_{\mathrm{keep}}^{(\ell,t)}\right),
    & i\in\mathcal{K}.
\end{cases}
\label{eq:gap}
\end{equation}
For edit queries, lower reference mass yields a larger positive offset, strengthening reference attention. For keep queries, we use a heuristic negative offset whose magnitude decreases as reference mass increases. This design aims to accommodate potentially necessary changes beyond the mask, such as hair extending past its boundary; reference mass alone does not distinguish these changes from unwanted leakage. 

The coefficients $\gamma_{\mathrm{edit}}$ and $\gamma_{\mathrm{keep}}$ are calibrated once during development and shared across approaches and tasks. Applying the same offset to every reference-key logit for a query changes total reference mass while preserving relative weights within the reference and non-reference groups.

\begin{figure*}[t]
\centering
\includegraphics[width=\textwidth]{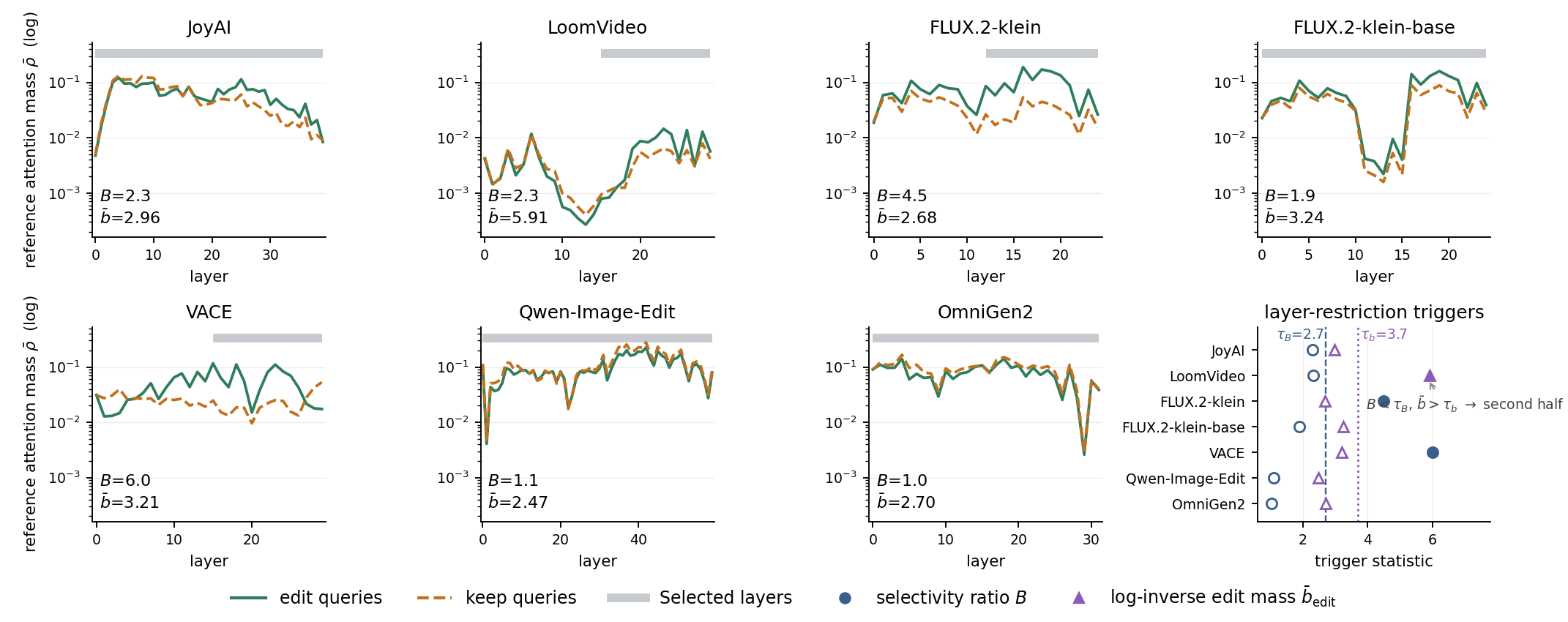}
\caption{\textbf{Reference-attention profiles and layer selection.} Reference-attention masses across layers on the head-swap validation split, shown on a log scale. Gray bars mark layers selected by \cref{eq:layer_rule}. The bottom-right panels report complementary statistics $B$ and $\bar b_{\mathrm{edit}}$, with their thresholds.}
\label{fig:rho_profiles}
\end{figure*}
    
\subsection{Layer selection}
\label{sec:layer_rule}

We derive the layer-selection rule from two recurring patterns in the reference-attention profiles of the four development approaches. \Cref{fig:rho_profiles} shows profiles and selected layers for seven approaches. First, some approaches exhibit clear edit-keep separation: $\bar\rho_{\mathrm{edit}}^{(\ell)}$ dominates $\bar\rho_{\mathrm{keep}}^{(\ell)}$ within a localized layer band, typically in the network's latter half, as in FLUX.2-klein~\citep{flux-2-2025}. Second, some assign little reference mass to edit queries across many layers, with $\bar\rho_{\mathrm{edit}}^{(\ell)}$ remaining around $10^{-2}$, as in LoomVideo~\citep{wu2026loomvideo}. The first pattern suggests a selective reference-reading band; the second indicates a risk of accumulating large corrections when intervening throughout the network.

We capture these behaviors with two complementary statistics. For each group $g\in\{\mathrm{edit},\mathrm{keep}\}$, let $\bar\rho_g^{(\ell)}$ denote the corresponding region mass from \cref{eq:region_rho}, averaged over denoising steps and validation clips. We define the edit-to-keep selectivity ratio
    \begin{equation}
    B=
    \max_{\ell}
    \frac{\bar\rho_{\mathrm{edit}}^{(\ell)}}
    {\bar\rho_{\mathrm{keep}}^{(\ell)}+\epsilon_B},
    \qquad
    \epsilon_B=10^{-6}.
    \label{eq:band_ratio}
    \end{equation}
A large $B$ indicates at least one layer assigns substantially more reference attention to the edit region than to the keep region. To summarize the potential strength of the edit-side correction, we compute
    \begin{equation}
    \bar b_{\mathrm{edit}} =
    \frac{1}{L}
    \sum_{\ell=1}^{L}
    \log\left(
    \frac{1}{\bar\rho_{\mathrm{edit}}^{(\ell)}}
    \right).
    \label{eq:mean_bias}
    \end{equation}
Whereas $B$ measures layer-wise selectivity, $\bar b_{\mathrm{edit}}$ averages the log-inverse edit mass across layers and serves as a proxy for reference-amplification strength. It is used only for layer selection; the deployed offsets remain those computed online at each attention layer and denoising step by \cref{eq:gap}.

Given thresholds $\tau_B$ and $\tau_b$, we define the set of correction layers as:
    
\begin{equation}
\mathcal{L}_{\mathrm{RefGAP}}=
\begin{cases}
\{\lfloor L/2\rfloor+1,\ldots,L\},
& B\ge \tau_B
\ \text{or}\
\bar b_{\mathrm{edit}}\ge \tau_b,\\[2pt]
\{1,\ldots,L\},
& \text{otherwise}.
\end{cases}
\label{eq:layer_rule}
\end{equation}

The thresholds and decision rule are fixed using only the four development approaches. For each approach, including the three held-out ones, we compute $B$ and $\bar b_{\mathrm{edit}}$ from its base-model head-swap validation profiles and apply the fixed rule to select correction layers. The resulting layer set is reused unchanged across tasks. This requires approach-specific profiling but no approach-specific strength or threshold tuning. Layer-selection ablations are reported in \cref{tab:abl_two}.

\section{Experiments}
\label{sec:exp}

\subsection{Experimental setup}
\label{sec:setup}

\paragraph{Approaches.}
We evaluate seven diffusion editors: JoyAI-Video-Edit~\citep{xiao2026joyai}, LoomVideo~\citep{wu2026loomvideo}, FLUX.2-klein and FLUX.2-klein-base~\citep{flux-2-2025}, VACE~\citep{vace}, Qwen-Image-Edit~\citep{qwenimageedit2509}, and OmniGen2~\citep{wu2025omnigen2}.

\paragraph{Tasks and datasets.}
Our primary evaluation focuses on head and face swapping. Head swapping uses the 1,040 test clips of HeadSwapBench~\citep{wang2025directswap}, and face swapping uses the 1,000 official target-source pairs from the FaceSwap subset of FaceForensics++~\citep{roessler2018faceforensics}.

We further evaluate transfer beyond identity editing through virtual try-on and background replacement. Try-on uses 90 ViViD clips~\citep{fang2024vivid}, with reference garments permuted within category. Background replacement uses 150 ground-truth videos from the HeadSwapBench test split as inputs, paired with 30 SUN397 scene references~\citep{Xiao:2010}. Across all four tasks, video approaches process video clips, while image approaches edit the middle frame of each clip. 

\paragraph{Metrics and comparison protocol.}
\label{sec:metrics}

\paragraph{Identity editing.}
For head and face swapping, we report AdaFace identity similarity~\citep{kim2022adaface}, keypoint error~\citep{lugaresi2019mediapipe}, and LPIPS~\citep{zhang2018unreasonable}.

For head swapping, $\mathrm{Repl.}$ denotes the percentage of outputs whose generated face is closer in identity to the reference than to the input subject. Identity similarity is measured against the ground-truth video, and keypoint error is measured against the corresponding ground-truth frames.

For face swapping, $\mathrm{Repl.}$ is the percentage of outputs whose nearest identity in the $978$-identity FF++ gallery is no longer the input subject, and $\mathrm{Retr.}$ is the percentage whose nearest identity is the reference. VACE's masked driving video hides the input identity, so we omit $\mathrm{Repl.}$ for VACE in both identity tasks. Identity similarity is measured against the reference face, whereas keypoint error is measured against the input (driving) video. 

For the main results, LPIPS uses task-specific preservation targets. For head swapping, it is computed over the whole frame against the ground-truth video. For face swapping, it is computed outside the face mask against the input video.

\paragraph{Transfer tasks.}
For virtual try-on and background replacement, we report region-specific DINOv2 similarity~\citep{oquab2023dinov2} to the reference image over the edited garment and background regions, respectively. LPIPS is computed outside the garment mask against the input for virtual try-on, and inside the person mask against the input for background replacement.

\paragraph{Common evaluation protocol.}
Base and $\gap$ use identical inputs, random seeds, and input-derived masks. No generated outputs are used during mask construction. All metrics are macro-averaged over clips, and identity comparisons use only clips with valid scores from both methods. Implementation details are provided in \cref{app:protocols}.

\subsection{Parameter calibration and transfer}
\label{sec:calib}

We jointly explore the two global coefficients and the layer coverage using only a 40-clip head-swap validation split evaluated on four development approaches: JoyAI-Video-Edit, LoomVideo, FLUX.2-klein, and FLUX.2-klein-base. Specifically, we enumerate
\[
\gamma_{\mathrm{edit}}\in\{0.4,0.6,0.8\},\qquad
\gamma_{\mathrm{keep}}\in\{0,0.4,0.8\},\qquad
\mathcal{L}\in\{\mathcal{L}_{\mathrm{all}},
\mathcal{L}_{\mathrm{half}}\},
\]
where $\mathcal{L}_{\mathrm{all}}$ denotes correction at all layers and $\mathcal{L}_{\mathrm{half}}$ denotes correction at the second half of the network. This results in $3\times3\times2=18$ candidate settings, allowing the coefficient and layer-coverage choices to be explored jointly.

For each combination of correction coefficients and layer coverage, we measure identity performance together with the change in non-edit LPIPS ($\mathrm{LPIPS}_{\mathrm{ne}}$). We retain settings satisfying
\[
\Delta\mathrm{LPIPS}_{\mathrm{ne}}(\gap)
=
\mathrm{LPIPS}_{\mathrm{ne}}(\gap)
-
\mathrm{LPIPS}_{\mathrm{ne}}(\mathrm{Base})\leq 0.005,
\]
and select the best admissible coefficient pair according to validation identity performance. 

In parallel, we validate whether the threshold-based rule correctly selects between all-layer and second-half correction. We set $\tau_B=2.7$ and $\tau_b=3.7$ to the arithmetic means of $B$ and $\bar b_{\mathrm{edit}}$, respectively, across the four development approaches. For each approach, we apply these fixed thresholds to its statistics and compare the resulting coverage with the corresponding all-layer and second-half results from the joint sweep. Full results are reported in \cref{tab:supp_calib}.

After this joint validation, we fix $\gamma_{\mathrm{edit}}=0.6$, $\gamma_{\mathrm{keep}}=0.8$, $\tau_B=2.7$, and $\tau_b=3.7$. We apply the shared rule to each approach's validation profile and reuse the selected layers across test sets and tasks, including the three held-out approaches. No further tuning is performed for these evaluations.

\subsection{Results across approaches and tasks}
\label{sec:tasks}

\begin{figure*}[t]
    \centering
    \includegraphics[width=.95\textwidth]{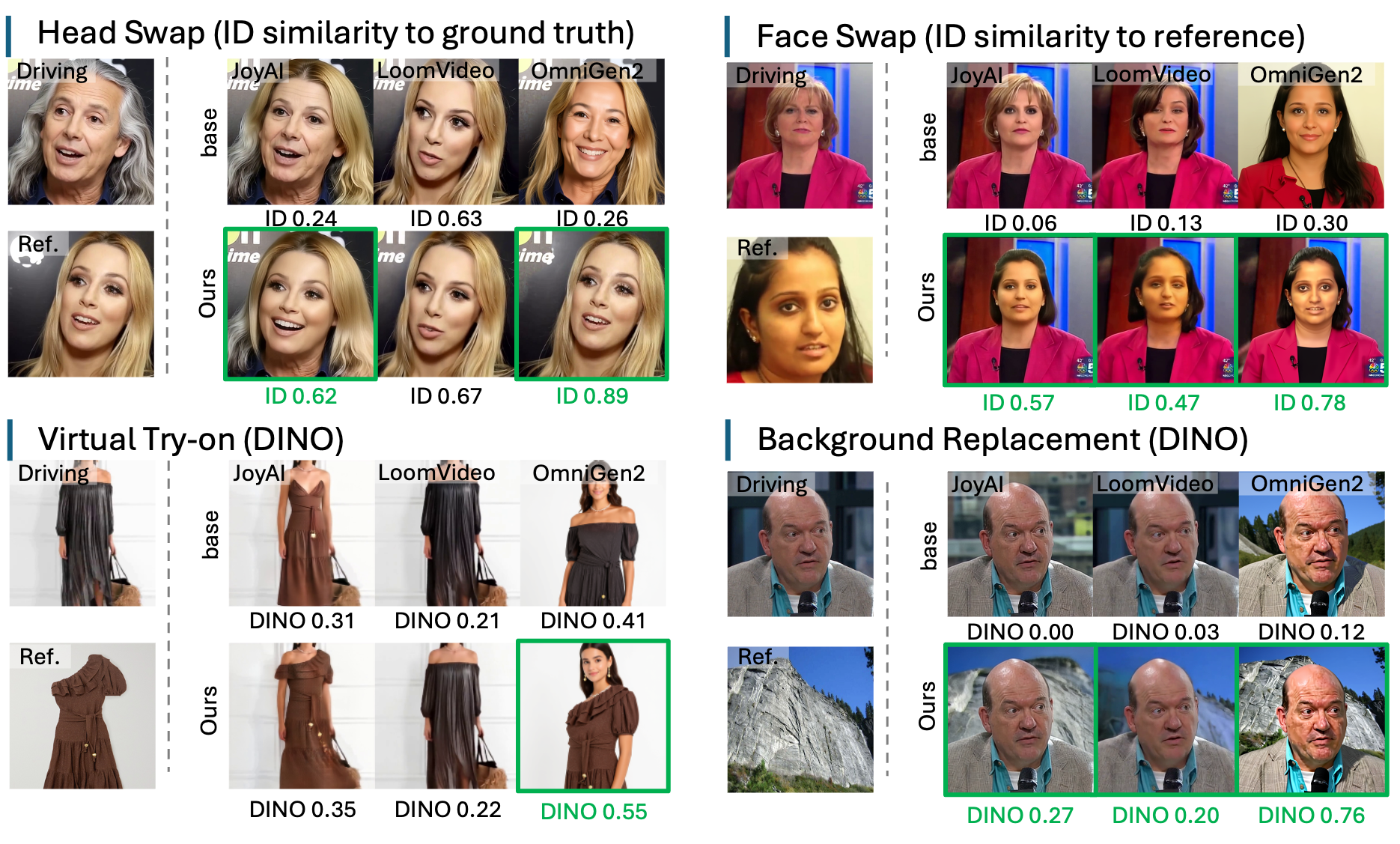}
    \caption{\textbf{Qualitative results across four tasks.} Base vs.\ Ours on JoyAI-Video-Edit (JoyAI), LoomVideo, and OmniGen2. Scores are shown below, with green borders marking gains $\geq0.05$. }
    \label{fig:result}
\end{figure*}

\Cref{tab:main} summarizes quantitative results across the seven approaches and four tasks, while \cref{fig:result} presents qualitative examples of head swap, face swap, virtual try-on, and background replacement on three approaches (all seven in \cref{fig:supp_whole}). 

\begin{table}[t]
\centering\scriptsize
\setlength{\tabcolsep}{2.0pt}
\renewcommand{\arraystretch}{1.0}
\resizebox{\textwidth}{!}{%
\begin{tabular}{l*{13}{c}}
\toprule
& \multicolumn{4}{c}{Head swap}
& \multicolumn{5}{c}{Face swap}
& \multicolumn{2}{c}{Virtual try-on}
& \multicolumn{2}{c}{Background replacement} \\
\cmidrule(lr){2-5}\cmidrule(lr){6-10}
\cmidrule(lr){11-12}\cmidrule(lr){13-14}
Approach
& Repl.$\uparrow$ & ID$\uparrow$ & Kpt.$\downarrow$ & LPIPS$\downarrow$
& Repl.$\uparrow$ & ID$\uparrow$ & Retr.$\uparrow$ & Kpt.$\downarrow$ & LPIPS$\downarrow$
& DINO$\uparrow$ & LPIPS$\downarrow$
& DINO$\uparrow$ & LPIPS$\downarrow$ \\
\midrule

JoyAI
& \makecell{68$\to$99\%\\{\tiny\textcolor{gain}{+31}}}
& \makecell{.507$\to$.644\\{\tiny\textcolor{gain}{+.137}}}
& \makecell{.043$\to$.085\\{\tiny\textcolor{cost}{+.042}}}
& \makecell{.218$\to$.215\\{\tiny\textcolor{gray}{$-$.003}}}
& \makecell{60$\to$97\%\\{\tiny\textcolor{gain}{+37}}}
& \makecell{.282$\to$.591\\{\tiny\textcolor{gain}{+.309}}}
& \makecell{39$\to$88\%\\{\tiny\textcolor{gain}{+49}}}
& \makecell{.057$\to$.105\\{\tiny\textcolor{cost}{+.048}}}
& \makecell{.086$\to$.108\\{\tiny\textcolor{cost}{+.022}}}
& \makecell{.283$\to$.298\\{\tiny\textcolor{gain}{+.015}}}
& \makecell{.099$\to$.101\\{\tiny\textcolor{gray}{+.002}}}
& \makecell{.242$\to$.286\\{\tiny\textcolor{gain}{+.044}}}
& \makecell{.063$\to$.065\\{\tiny\textcolor{gray}{+.002}}} \\

LoomVideo
& \makecell{79$\to$92\%\\{\tiny\textcolor{gain}{+13}}}
& \makecell{.608$\to$.665\\{\tiny\textcolor{gain}{+.057}}}
& \makecell{.107$\to$.098\\{\tiny\textcolor{gain}{$-$.009}}}
& \makecell{.281$\to$.275\\{\tiny\textcolor{gain}{$-$.006}}}
& \makecell{71$\to$84\%\\{\tiny\textcolor{gain}{+13}}}
& \makecell{.449$\to$.466\\{\tiny\textcolor{gain}{+.017}}}
& \makecell{63$\to$75\%\\{\tiny\textcolor{gain}{+12}}}
& \makecell{.083$\to$.086\\{\tiny\textcolor{gray}{+.003}}}
& \makecell{.180$\to$.178\\{\tiny\textcolor{gray}{$-$.002}}}
& \makecell{.174$\to$.183\\{\tiny\textcolor{gain}{+.009}}}
& \makecell{.105$\to$.105\\{\tiny\textcolor{gray}{.000}}}
& \makecell{.135$\to$.258\\{\tiny\textcolor{gain}{+.123}}}
& \makecell{.051$\to$.058\\{\tiny\textcolor{cost}{+.007}}} \\

Klein
& \makecell{91$\to$100\%\\{\tiny\textcolor{gain}{+9}}}
& \makecell{.726$\to$.790\\{\tiny\textcolor{gain}{+.064}}}
& \makecell{.097$\to$.153\\{\tiny\textcolor{cost}{+.056}}}
& \makecell{.222$\to$.217\\{\tiny\textcolor{gain}{$-$.005}}}
& \makecell{98$\to$100\%\\{\tiny\textcolor{gain}{+2}}}
& \makecell{.704$\to$.820\\{\tiny\textcolor{gain}{+.116}}}
& \makecell{90$\to$96\%\\{\tiny\textcolor{gain}{+6}}}
& \makecell{.065$\to$.082\\{\tiny\textcolor{cost}{+.017}}}
& \makecell{.064$\to$.076\\{\tiny\textcolor{cost}{+.012}}}
& \makecell{.363$\to$.372\\{\tiny\textcolor{gain}{+.009}}}
& \makecell{.093$\to$.092\\{\tiny\textcolor{gray}{$-$.001}}}
& \makecell{.629$\to$.630\\{\tiny\textcolor{gray}{+.001}}}
& \makecell{.038$\to$.037\\{\tiny\textcolor{gray}{$-$.001}}} \\

Klein-base
& \makecell{77$\to$100\%\\{\tiny\textcolor{gain}{+23}}}
& \makecell{.588$\to$.799\\{\tiny\textcolor{gain}{+.211}}}
& \makecell{.169$\to$.236\\{\tiny\textcolor{cost}{+.067}}}
& \makecell{.326$\to$.261\\{\tiny\textcolor{gain}{$-$.065}}}
& \makecell{95$\to$100\%\\{\tiny\textcolor{gain}{+5}}}
& \makecell{.557$\to$.831\\{\tiny\textcolor{gain}{+.274}}}
& \makecell{74$\to$96\%\\{\tiny\textcolor{gain}{+22}}}
& \makecell{.129$\to$.223\\{\tiny\textcolor{cost}{+.094}}}
& \makecell{.173$\to$.179\\{\tiny\textcolor{cost}{+.006}}}
& \makecell{.409$\to$.465\\{\tiny\textcolor{gain}{+.056}}}
& \makecell{.150$\to$.160\\{\tiny\textcolor{cost}{+.010}}}
& \makecell{.632$\to$.651\\{\tiny\textcolor{gain}{+.019}}}
& \makecell{.162$\to$.136\\{\tiny\textcolor{gain}{$-$.026}}} \\

\midrule
VACE
& ---
& \makecell{.405$\to$.729\\{\tiny\textcolor{gain}{+.324}}}
& \makecell{.198$\to$.200\\{\tiny\textcolor{gray}{+.002}}}
& \makecell{.228$\to$.214\\{\tiny\textcolor{gain}{$-$.014}}}
& ---
& \makecell{.541$\to$.818\\{\tiny\textcolor{gain}{+.277}}}
& \makecell{87$\to$98\%\\{\tiny\textcolor{gain}{+11}}}
& \makecell{.187$\to$.193\\{\tiny\textcolor{cost}{+.006}}}
& \makecell{.018$\to$.019\\{\tiny\textcolor{gray}{+.001}}}
& \makecell{.171$\to$.226\\{\tiny\textcolor{gain}{+.055}}}
& \makecell{.056$\to$.056\\{\tiny\textcolor{gray}{.000}}}
& \makecell{.008$\to$.014\\{\tiny\textcolor{gain}{+.006}}}
& \makecell{.008$\to$.008\\{\tiny\textcolor{gray}{.000}}} \\

Qwen
& \makecell{70$\to$86\%\\{\tiny\textcolor{gain}{+16}}}
& \makecell{.609$\to$.723\\{\tiny\textcolor{gain}{+.114}}}
& \makecell{.179$\to$.179\\{\tiny\textcolor{gray}{.000}}}
& \makecell{.449$\to$.237\\{\tiny\textcolor{gain}{$-$.212}}}
& \makecell{65$\to$82\%\\{\tiny\textcolor{gain}{+17}}}
& \makecell{.518$\to$.687\\{\tiny\textcolor{gain}{+.169}}}
& \makecell{59$\to$79\%\\{\tiny\textcolor{gain}{+20}}}
& \makecell{.097$\to$.118\\{\tiny\textcolor{cost}{+.021}}}
& \makecell{.069$\to$.065\\{\tiny\textcolor{gray}{$-$.004}}}
& \makecell{.213$\to$.248\\{\tiny\textcolor{gain}{+.035}}}
& \makecell{.077$\to$.094\\{\tiny\textcolor{cost}{+.017}}}
& \makecell{.299$\to$.042\\{\tiny\textcolor{cost}{$-$.257}}}
& \makecell{.027$\to$.024\\{\tiny\textcolor{gray}{$-$.003}}} \\

OmniGen2
& \makecell{34$\to$99\%\\{\tiny\textcolor{gain}{+65}}}
& \makecell{.207$\to$.693\\{\tiny\textcolor{gain}{+.486}}}
& \makecell{.255$\to$.258\\{\tiny\textcolor{gray}{+.003}}}
& \makecell{.464$\to$.365\\{\tiny\textcolor{gain}{$-$.099}}}
& \makecell{79$\to$100\%\\{\tiny\textcolor{gain}{+21}}}
& \makecell{.157$\to$.656\\{\tiny\textcolor{gain}{+.499}}}
& \makecell{20$\to$91\%\\{\tiny\textcolor{gain}{+71}}}
& \makecell{.238$\to$.249\\{\tiny\textcolor{cost}{+.011}}}
& \makecell{.323$\to$.279\\{\tiny\textcolor{gain}{$-$.044}}}
& \makecell{.277$\to$.355\\{\tiny\textcolor{gain}{+.078}}}
& \makecell{.422$\to$.404\\{\tiny\textcolor{gain}{$-$.018}}}
& \makecell{.245$\to$.515\\{\tiny\textcolor{gain}{+.270}}}
& \makecell{.259$\to$.295\\{\tiny\textcolor{cost}{+.036}}} \\

\midrule
mean $\Delta$
& \textcolor{gain}{+26}
& \textcolor{gain}{+.199}
& \textcolor{cost}{+.023}
& \textcolor{gain}{$-$.058}
& \textcolor{gain}{+16}
& \textcolor{gain}{+.237}
& \textcolor{gain}{+27}
& \textcolor{cost}{+.029}
& \textcolor{gray}{$-$.001}
& \textcolor{gain}{+.037}
& \textcolor{gray}{+.001}
& \textcolor{gain}{+.029}
& \textcolor{gray}{+.002} \\
\bottomrule
\end{tabular}%
}
\caption{\textbf{Quantitative results across seven approaches and four tasks.}
JoyAI, Klein, Klein-base, and Qwen denote JoyAI-Video-Edit, FLUX.2-klein,
FLUX.2-klein-base, and Qwen-Image-Edit, respectively.
Cells show Base$\to$Ours, with differences shown below.
Mean differences average all seven approaches (six for Repl., which is undefined for VACE).
Green and red indicate improvements and degradations, respectively;
gray denotes $|\Delta|<0.005$. Differences are computed from the displayed values; unrounded paired estimates are given in \cref{tab:supp_main_ci}.}

\label{tab:main}
\end{table}

\paragraph{Identity editing: head and face swapping.}
\label{sec:blind}

For head swapping, \gap raises the replacement rate to at least \(86\%\) on every applicable approach, improves identity similarity on all seven approaches, and does not increase whole-frame LPIPS on any approach, reducing it on four with intervals excluding zero (\cref{tab:supp_main_ci}). On Qwen-Image-Edit, $57.7\%$ of base outputs are near-copies of the reference (whole-frame LPIPS to the reference below $0.2$), which inflates the base identity similarity; \gap lowers this rate to $9.1\%$ (\cref{tab:supp_leak}). Keypoint error increases mainly on JoyAI-Video-Edit, FLUX.2-klein, and FLUX.2-klein-base. \Cref{sec:supp_knob} shows that delaying correction to later denoising steps substantially reduces keypoint degradation while retaining identity gains over the base model.

For face swapping, \gap improves identity similarity across all approaches, with replacement and correct retrieval rates reaching at least \(82\%\) and \(75\%\), respectively. Keypoint error rises on most approaches, while non-face LPIPS varies by approach. Without ground-truth outputs, LPIPS compares against the input video, potentially counting identity-related geometry differences as preservation errors. Larger identity gains generally accompany larger keypoint changes (\cref{sec:face_tradeoff}).

\paragraph{Transfer beyond identity editing.}
With the same coefficients, \gap improves virtual try-on reference fidelity on all seven approaches (\cref{tab:main}, intervals excluding zero on four). Background replacement has a different objective from identity editing and try-on: the surrounding scene should change, while the foreground subject should remain unchanged. It improves on JoyAI-Video-Edit, LoomVideo, and OmniGen2, changes little on the FLUX.2 variants and VACE, and degrades on Qwen-Image-Edit (\cref{tab:supp_main_ci}). These mixed results highlight that reference transfer remains task- and approach-dependent.

\subsection{Human evaluation}
\label{sec:user}

\begin{table*}[t]
\centering\scriptsize
\setlength{\tabcolsep}{2.2pt}
\resizebox{\textwidth}{!}{%
\begin{tabular}{l*{12}{c}}
\toprule
& \multicolumn{3}{c}{Head swap} & \multicolumn{3}{c}{Face swap} & \multicolumn{3}{c}{Virtual try-on} & \multicolumn{3}{c}{Background replacement} \\
\cmidrule(lr){2-4}\cmidrule(lr){5-7}\cmidrule(lr){8-10}\cmidrule(lr){11-13}
Approach & Ref. & Pose & Qual. & Ref. & Pose & Qual. & Ref. & Pose & Qual. & Ref. & Pose & Qual. \\
\midrule
JoyAI-Video-Edit & \textcolor{gain}{28}/0/\textcolor{cost}{0} & \textcolor{gain}{3}/25/\textcolor{cost}{0} & \textcolor{gain}{5}/23/\textcolor{cost}{0} & \textcolor{gain}{50}/7/\textcolor{cost}{0} & \textcolor{gain}{3}/50/\textcolor{cost}{4} & \textcolor{gain}{5}/50/\textcolor{cost}{2} & \textcolor{gain}{37}/5/\textcolor{cost}{12} & \textcolor{gain}{0}/52/\textcolor{cost}{2} & \textcolor{gain}{5}/46/\textcolor{cost}{3} & \textcolor{gain}{19}/25/\textcolor{cost}{6} & \textcolor{gain}{0}/50/\textcolor{cost}{0} & \textcolor{gain}{1}/46/\textcolor{cost}{3} \\
LoomVideo & \textcolor{gain}{7}/18/\textcolor{cost}{2} & \textcolor{gain}{12}/14/\textcolor{cost}{2} & \textcolor{gain}{8}/14/\textcolor{cost}{6} & \textcolor{gain}{18}/31/\textcolor{cost}{3} & \textcolor{gain}{3}/44/\textcolor{cost}{5} & \textcolor{gain}{1}/33/\textcolor{cost}{18} & \textcolor{gain}{36}/15/\textcolor{cost}{0} & \textcolor{gain}{0}/51/\textcolor{cost}{0} & \textcolor{gain}{3}/48/\textcolor{cost}{0} & \textcolor{gain}{39}/12/\textcolor{cost}{0} & \textcolor{gain}{0}/50/\textcolor{cost}{1} & \textcolor{gain}{1}/49/\textcolor{cost}{1} \\
FLUX.2-klein & \textcolor{gain}{20}/5/\textcolor{cost}{0} & \textcolor{gain}{0}/25/\textcolor{cost}{0} & \textcolor{gain}{7}/18/\textcolor{cost}{0} & \textcolor{gain}{33}/16/\textcolor{cost}{2} & \textcolor{gain}{0}/48/\textcolor{cost}{3} & \textcolor{gain}{1}/48/\textcolor{cost}{2} & \textcolor{gain}{1}/11/\textcolor{cost}{1} & \textcolor{gain}{1}/12/\textcolor{cost}{0} & \textcolor{gain}{0}/12/\textcolor{cost}{1} & \textcolor{gain}{0}/10/\textcolor{cost}{0} & \textcolor{gain}{0}/10/\textcolor{cost}{0} & \textcolor{gain}{0}/10/\textcolor{cost}{0} \\
FLUX.2-klein-base & \textcolor{gain}{24}/1/\textcolor{cost}{0} & \textcolor{gain}{0}/25/\textcolor{cost}{0} & \textcolor{gain}{3}/22/\textcolor{cost}{0} & \textcolor{gain}{45}/8/\textcolor{cost}{0} & \textcolor{gain}{2}/45/\textcolor{cost}{6} & \textcolor{gain}{7}/44/\textcolor{cost}{2} & \textcolor{gain}{20}/16/\textcolor{cost}{18} & \textcolor{gain}{0}/53/\textcolor{cost}{1} & \textcolor{gain}{1}/51/\textcolor{cost}{2} & \textcolor{gain}{15}/40/\textcolor{cost}{1} & \textcolor{gain}{0}/56/\textcolor{cost}{0} & \textcolor{gain}{10}/45/\textcolor{cost}{1} \\
VACE & \textcolor{gain}{24}/0/\textcolor{cost}{1} & \textcolor{gain}{0}/24/\textcolor{cost}{1} & \textcolor{gain}{6}/15/\textcolor{cost}{4} & \textcolor{gain}{35}/19/\textcolor{cost}{1} & \textcolor{gain}{5}/48/\textcolor{cost}{2} & \textcolor{gain}{2}/50/\textcolor{cost}{3} & \textcolor{gain}{57}/0/\textcolor{cost}{0} & \textcolor{gain}{5}/52/\textcolor{cost}{0} & \textcolor{gain}{8}/49/\textcolor{cost}{0} & --- & --- & --- \\
Qwen-Image-Edit & \textcolor{gain}{29}/0/\textcolor{cost}{0} & \textcolor{gain}{2}/27/\textcolor{cost}{0} & \textcolor{gain}{8}/21/\textcolor{cost}{0} & \textcolor{gain}{41}/15/\textcolor{cost}{1} & \textcolor{gain}{3}/39/\textcolor{cost}{15} & \textcolor{gain}{5}/50/\textcolor{cost}{2} & \textcolor{gain}{48}/2/\textcolor{cost}{5} & \textcolor{gain}{1}/54/\textcolor{cost}{0} & \textcolor{gain}{2}/53/\textcolor{cost}{0} & \textcolor{gain}{5}/25/\textcolor{cost}{21} & \textcolor{gain}{0}/50/\textcolor{cost}{1} & \textcolor{gain}{0}/49/\textcolor{cost}{2} \\
OmniGen2 & \textcolor{gain}{25}/0/\textcolor{cost}{0} & \textcolor{gain}{10}/15/\textcolor{cost}{0} & \textcolor{gain}{0}/25/\textcolor{cost}{0} & \textcolor{gain}{56}/0/\textcolor{cost}{1} & \textcolor{gain}{3}/49/\textcolor{cost}{5} & \textcolor{gain}{11}/45/\textcolor{cost}{1} & \textcolor{gain}{50}/1/\textcolor{cost}{4} & \textcolor{gain}{2}/53/\textcolor{cost}{0} & \textcolor{gain}{5}/50/\textcolor{cost}{0} & \textcolor{gain}{49}/9/\textcolor{cost}{1} & \textcolor{gain}{5}/53/\textcolor{cost}{1} & \textcolor{gain}{5}/51/\textcolor{cost}{2} \\
\midrule
All items & \textcolor{gain}{157}/24/\textcolor{cost}{3} & \textcolor{gain}{27}/155/\textcolor{cost}{3} & \textcolor{gain}{37}/138/\textcolor{cost}{10} & \textcolor{gain}{278}/96/\textcolor{cost}{8} & \textcolor{gain}{19}/323/\textcolor{cost}{40} & \textcolor{gain}{32}/320/\textcolor{cost}{30} & \textcolor{gain}{249}/50/\textcolor{cost}{40} & \textcolor{gain}{9}/327/\textcolor{cost}{3} & \textcolor{gain}{24}/309/\textcolor{cost}{6} & \textcolor{gain}{127}/121/\textcolor{cost}{29} & \textcolor{gain}{5}/269/\textcolor{cost}{3} & \textcolor{gain}{17}/250/\textcolor{cost}{9} \\
\bottomrule
\end{tabular}}
\caption{\textbf{Human evaluation across four tasks.} Cells report judgment counts as \textcolor{gain}{\gap}/tie/\textcolor{cost}{base} for reference fidelity (Ref.), pose/expression consistency with the input (Pose), and visual quality (Qual.). Ties include ``both succeed'' and ``both fail''; ``---'' denotes an unevaluated setting.}
\label{tab:user}
\end{table*}

We evaluate $219$ items with $54$ raters retained after attention checks. Each item presents the input, reference, and base/\gap outputs in randomized A/B order. Raters assess reference fidelity, pose/expression consistency with the input, and visual quality, choosing A, B, both succeed, or both fail. To prioritize informative comparisons, we use outputs with visible differences; results describe these selected comparisons. An unscreened audit of 56 randomly sampled comparisons by five raters favors \gap in reference fidelity (\cref{app:qual}). \Cref{tab:user} reports preferences for \gap in reference fidelity across head swap, face swap and virtual try-on, consistent with the generally improved identity similarity in head and face swapping and DINO scores in virtual try-on (\cref{tab:main}).

Preservation results are mixed. Pose/expression judgments are mostly ties (both-success or both-failure). Non-tied counts (\gap versus base) are $27$ to $3$ for head swapping and $19$ to $40$ for face swapping. Reference-fidelity judgments for background replacement vary across approaches. Quantitative results vary across approaches, including clear degradation on Qwen-Image-Edit.

\subsection{Ablation studies}
\label{sec:ablations}

We ablate three \gap components: attention-conditioned strength, keep-side correction, and layer coverage, using the four development approaches on head swapping. We use non-edit LPIPS (LPIPS$_{\mathrm{ne}}$) for validation and whole-frame LPIPS for the fixed-bias test comparison. Unlike the whole-frame LPIPS in \cref{tab:main}, LPIPS$_{\mathrm{ne}}$ is computed outside the dilated head mask to detect reference leakage into the keep region. For image approaches, validation metrics use the middle frame. 

\begin{table}[H]
\centering\scriptsize
\begin{tabular}{l|ccc|cccc}
\toprule
& \multicolumn{3}{c|}{\gap} & & \multicolumn{3}{c}{best admissible edit-side constant} \\
Approach & ID$\uparrow$ & Kpt.$\downarrow$ & LPIPS$\downarrow$ & $b$ & ID$\uparrow$ & Kpt.$\downarrow$ & LPIPS$\downarrow$ \\
\midrule
JoyAI-Video-Edit  & .644 & .085 & .215 & 1.5 & .644 & .075 & .207 \\
LoomVideo         & .665 & .098 & .275 & 2.5 & .671 & .098 & .271 \\
FLUX.2-klein      & .790 & .153 & .217 & 2.0 & .794 & .171 & .229 \\
FLUX.2-klein-base & .799 & .236 & .261 & 2.0 & .801 & .236 & .260 \\
\bottomrule
\end{tabular}
\caption{\textbf{\gap versus approach-specific constant
edit-side biases.} Constants are selected on validation data
under $\Delta\mathrm{LPIPS}_{\mathrm{ne}}\leq0.005$.
Only the edit-side bias differs between settings.
Results use the full $1{,}040$-clip head-swap test set;
LPIPS is measured over the whole frame.}
\label{tab:abl_fixed}
\end{table}

\paragraph{Online versus fixed strength.}
\label{sec:handtuning}

Diptych Prompting~\citep{shin2025large} boosts reference attention in the edited region by a uniform constant.
As its editing code is unreleased, we compare against a constant-bias baseline inspired by Diptych Prompting: a constant edit-side bias $b$ combined with our keep-side correction and layer sets, tuned separately for each approach.
On the 40-clip validation split, we sweep $b \in \{0.5, 1, \dots, 3\}$ and select the highest-identity value with $\Delta\mathrm{LPIPS}_{\mathrm{ne}} \le 0.005$ (\cref{tab:supp_matrix}).
The selected constants range from 1.5 to 2.5, confirming that the optimal fixed strength is approach-dependent.
On the 1,040-clip test set, they improve identity over RefGAP by only 0.000--0.006, with mixed keypoint and LPIPS trade-offs (\cref{tab:abl_fixed}).
RefGAP thus matches a per-approach-tuned Diptych-style baseline without per-approach strength sweeps.
We also compare against a single constant edit/keep pair shared across approaches (\cref{app:shared_const}). Since copying the reference inflates identity similarity, we count near-copies (\cref{sec:blind}) and score their identity as $0$. The shared constant produces more near-copies than \gap on all four image approaches, and with near-copies scored $0$, \gap's identity is higher on four of seven approaches, tied on JoyAI-Video-Edit and OmniGen2, and $.007$ lower on VACE (\cref{tab:supp_leak}).

\paragraph{One-sided versus two-sided correction.}
\label{sec:twosided}

With $\gamma_{\mathrm{edit}}=0.6$, we vary $\gamma_{\mathrm{keep}}\in\{0,0.4,0.8\}$, where zero gives the one-sided variant. Omitting keep-side correction severely degrades non-edit preservation (\cref{tab:abl_two}, left): FLUX.2-klein-base's $\text{LPIPS}_{\mathrm{ne}}$ rises from $0.1206$ (base) to $0.2115$. Increasing $\gamma_{\mathrm{keep}}$ to $0.8$ lowers $\text{LPIPS}_{\mathrm{ne}}$ across most models, reaching $0.0674$ on FLUX.2-klein-base, while largely preserving identity similarity. These results support keep-side correction as a means to contain strong edit-side interventions and limit reference-induced changes in preserved content.

\paragraph{Layer coverage.}
\label{sec:layersel}

As specified in \cref{sec:layer_rule}, \gap applies the correction only to the second half of the network when \(B\ge2.7\) or \(\bar b_{\mathrm{edit}}\ge3.7\), and to all layers otherwise. With \(\gamma_{\mathrm{edit}}=0.6\) and \(\gamma_{\mathrm{keep}}=0.8\), \cref{tab:abl_two} supports this rule. JoyAI-Video-Edit and FLUX.2-klein-base do not satisfy either threshold and perform better with all-layer correction. FLUX.2-klein is selected by its large \(B\), and the second-half setting substantially lowers LPIPS\(_{\mathrm{ne}}\) from \(0.055\) to \(0.041\) with only a small ID decrease from \(0.893\) to \(0.885\). LoomVideo is selected by its large \(\bar b_{\mathrm{edit}}\): applying the correction to all layers gives higher ID (\(0.826\)) but increases LPIPS\(_{\mathrm{ne}}\) from the base value \(0.070\) to \(0.278\), whereas the second-half setting gives \(0.767\) ID and \(0.073\) LPIPS\(_{\mathrm{ne}}\). These results support using \(B\) to target selective reference-reading bands and \(\bar b_{\mathrm{edit}}\) to prevent excessive accumulation of large corrections.

\begin{table}[t]
\centering\scriptsize
\setlength{\tabcolsep}{2.6pt}
\sbox0{\begin{tabular}{l|c|ccc|ccc}
\toprule
\multicolumn{8}{c}{Keep side: $\gamma_{\mathrm{edit}}=0.6$, rule-selected layers, $\gamma_{\mathrm{keep}}\in\{0,0.4,0.8\}$} \\
\midrule
 & & \multicolumn{3}{c|}{LPIPS$_{\mathrm{ne}}$$\downarrow$} & \multicolumn{3}{c}{ID$\uparrow$} \\
Approach & base & $\gamma_{\mathrm{keep}}=0$ & $0.4$ & $0.8$ & $0$ & $0.4$ & $0.8$ \\
\midrule
JoyAI-Video-Edit  & .0663 & \textbf{.0635} & .0644 & .0641 & .705 & .705 & .706 \\
LoomVideo         & .0700 & .0781$^{\times}$ & .0746 & \textbf{.0727} & .768 & .767 & .767 \\
FLUX.2-klein      & .0435 & .0524$^{\times}$ & .0416 & \textbf{.0406} & .885 & .885 & .885 \\
FLUX.2-klein-base & .1206 & .2115$^{\times}$ & .0764 & \textbf{.0674} & .879 & .877 & .878 \\
\bottomrule
\end{tabular}}%
\sbox2{\begin{tabular}{l|ccc|cc|cc}
\toprule
\multicolumn{8}{c}{layer coverage: $\gamma_{\mathrm{edit}}=0.6$, $\gamma_{\mathrm{keep}}=0.8$} \\
\midrule
 & \multicolumn{3}{c|}{layer rule} & \multicolumn{2}{c|}{all layers} & \multicolumn{2}{c}{second half} \\
Approach & $B$ & $\bar b_{\mathrm{edit}}$ & pred. & ID & LPIPS$_{\text{ne}}$ & ID & LPIPS$_{\text{ne}}$ \\
\midrule
JoyAI-Video-Edit  & 2.30 & 2.96 & all  & \textbf{.706} & \textbf{.064} & .674 & .067 \\
LoomVideo         & 2.32 & 5.91 & half & .826 & .278 & \textbf{.767} & \textbf{.073} \\
FLUX.2-klein      & 4.48 & 2.68 & half & .893 & .055 & \textbf{.885} & \textbf{.041} \\
FLUX.2-klein-base & 1.88 & 3.24 & all  & \textbf{.878} & \textbf{.067} & .826 & .093 \\
\bottomrule
\end{tabular}}%
\ifdim\wd0>\wd2
  \sbox4{\resizebox{0.485\textwidth}{!}{\usebox0}}\sbox6{\resizebox*{!}{\dimexpr\ht4+\dp4\relax}{\usebox2}}%
\else
  \sbox6{\resizebox{0.485\textwidth}{!}{\usebox2}}\sbox4{\resizebox*{!}{\dimexpr\ht6+\dp6\relax}{\usebox0}}%
\fi
\noindent\begin{minipage}[t]{0.485\textwidth}\centering\usebox4\end{minipage}\hfill
\begin{minipage}[t]{0.485\textwidth}\centering\usebox6\end{minipage}
\caption{\textbf{Keep-side and layer-coverage ablations.} Only $\gamma_{\mathrm{keep}}$ (left) or corrected layers (right) vary on the validation split. $B$ and $\bar b_{\mathrm{edit}}$ are computed from the base model's validation passes; pred.\ denotes the layer set selected by \cref{eq:layer_rule}. $^{\times}$ marks LPIPS$_{\mathrm{ne}}$ increases $>0.005$ over the base. Bold marks the lowest LPIPS$_{\mathrm{ne}}$ (left) and the rule-selected setting (right).}
\label{tab:abl_two}
\end{table}

\section{Limitations}
\label{sec:limitations}

\paragraph{Identity--attribute trade-off.}
The all-step default prioritizes reference identity and can compromise pose and expression preservation, because one offset applies to all reference tokens and does not separate identity from the reference's pose and expression. Delaying correction can mitigate these costs: on JoyAI-Video-Edit and FLUX.2-klein, skipping the first correction step substantially reduces keypoint error while retaining identity improvements over the base. However, the required delay varies across sampling schedules, and stronger geometric preservation reduces identity gains (\cref{tab:supp_knob}).

\paragraph{An approach-task-specific failure.}
Despite broad gains in identity editing and virtual try-on, \gap reduces background fidelity on Qwen-Image-Edit. Restricting layer coverage or using a constant edit-side bias does not resolve this degradation (\cref{tab:supp_qwen_bg}), and we have not identified its cause. This approach-task pair may require a different correction strategy.

\paragraph{Image quality.}
MUSIQ~\citep{ke2021musiq}, an automatic no-reference quality score from $0$ to $100$, changes by at most $2.5$ points on $11$ of the $14$ head- and face-swap approach-task pairs. It drops more on LoomVideo (head swap $-4.0$, face swap $-8.2$) and on FLUX.2-klein-base face swap ($-5.1$), where the identity gain comes with lower image quality.

\paragraph{Validation profiling overhead.}
Applying \gap to a new approach requires one base-model profiling run on a validation split to select correction layers under the fixed rule. The coefficients and thresholds remain unchanged, and the selected layers are reused across tasks. This adds setup cost, although it avoids repeated evaluations over a grid of correction strengths.

\section{Conclusion}

We introduce \gap, a training-free method that uses measured reference-attention mass to dynamically strengthen reference influence in edit regions and suppress it in keep regions. With two coefficients calibrated on four development approaches and a shared layer-selection rule, \gap improves identity fidelity in head and face swapping across seven approaches, including three held-out ones. It achieves a fidelity-preservation trade-off comparable to approach-specific constant edit-side biases without per-approach strength sweeps. Transfer without task-specific retuning yields positive fidelity gains in virtual try-on but mixed results in background replacement. Together, these results support measured attention as a practical signal for controlling reference influence, while leaving attribute preservation and reliable cross-task transfer as directions for further work.

\subsection*{AI use statement}
Generative AI tools were used to polish the writing, prepare comparison figures and user study materials, and check numerical consistency between tables and the underlying result files. The authors developed the method, designed the experiments, and interpreted the results.

\subsection*{Ethics statement}
This work studies reference-guided visual editing, including head and face swapping, which can be misused to fabricate media of real individuals. All images and videos used in our evaluation are drawn from public benchmarks (HeadSwapBench, FaceForensics++, ViViD, VITON-HD, SUN397, VFHQ, DreamEditBench, and DreamSwapV). We release no new identity data and neither train nor release a new model; our method is an inference-time modification of publicly released editors. 

For the human evaluation, participants were undergraduate, master's, and doctoral students with a background in computer vision. All participants provided informed consent before taking part. We collected their judgments of the editing results without collecting personally identifiable information.

\subsection*{Reproducibility statement}
All seven approaches are publicly available and used through their released pipelines. RefGAP modifies attention logits at inference time and requires no training. The correction rule is described in \cref{sec:method}, and its coefficients and calibration procedure are reported in \cref{sec:calib}. \Cref{app:protocols} provides implementation details and numerical safeguards; \cref{app:ablations} reports calibration results and additional ablations; \cref{app:results} describes task-specific evaluation protocols and additional results; \cref{app:extra} reports comparisons with shared constants, SaaS, and KV-Edit, and the object-replacement and multi-person experiments; and \cref{app:qual} details the human evaluation and its statistical analysis. Code and per-clip result files will be released.

\bibliography{iclr2027_conference}

@String(CVPR= {IEEE Conf. Comput. Vis. Pattern Recog.})

@String(ICCV= {Int. Conf. Comput. Vis.})

@String(CVPRW= {IEEE Conf. Comput. Vis. Pattern Recog. Worksh.})

@String(CVPR  = {CVPR})

@String(ICCV  = {ICCV})

@String(CVPRW= {CVPRW})

@inproceedings{vace,
    title = {{VACE}: All-in-One Video Creation and Editing},
    author = {Jiang, Zeyinzi and Han, Zhen and Mao, Chaojie and Zhang, Jingfeng and Pan, Yulin and Liu, Yu},
    booktitle = {Proceedings of the IEEE/CVF International Conference on Computer Vision},
    pages = {17191-17202},
    year = {2025}
}

@misc{qwenimageedit2509,
      title={{Qwen-Image} Technical Report},
      author={Chenfei Wu and Jiahao Li and Jingren Zhou and Junyang Lin and Kaiyuan Gao and Kun Yan and Sheng-ming Yin and Shuai Bai and Xiao Xu and Yilei Chen and Yuxiang Chen and Zecheng Tang and Zekai Zhang and Zhengyi Wang and An Yang and Bowen Yu and Chen Cheng and Dayiheng Liu and Deqing Li and Hang Zhang and Hao Meng and Hu Wei and Jingyuan Ni and Kai Chen and Kuan Cao and Liang Peng and Lin Qu and Minggang Wu and Peng Wang and Shuting Yu and Tingkun Wen and Wensen Feng and Xiaoxiao Xu and Yi Wang and Yichang Zhang and Yongqiang Zhu and Yujia Wu and Yuxuan Cai and Zenan Liu},
      year={2025},
      eprint={2508.02324},
      archivePrefix={arXiv},
      primaryClass={cs.CV},
      url={https://arxiv.org/abs/2508.02324},
}

@article{wu2025omnigen2,
  title={{OmniGen2}: Exploration to Advanced Multimodal Generation},
  author={Chenyuan Wu and Pengfei Zheng and Ruiran Yan and Shitao Xiao and Xin Luo and Yueze Wang and Wanli Li and Xiyan Jiang and Yexin Liu and Junjie Zhou and Ze Liu and Ziyi Xia and Chaofan Li and Haoge Deng and Jiahao Wang and Kun Luo and Bo Zhang and Defu Lian and Xinlong Wang and Zhongyuan Wang and Tiejun Huang and Zheng Liu},
  journal={arXiv preprint arXiv:2506.18871},
  year={2025}
}

@misc{flux-2-2025,
    author={{Black Forest Labs}},
    title={{FLUX.2: Frontier Visual Intelligence}},
    year={2025},
    howpublished={\url{https://bfl.ai/blog/flux-2}},
}

@article{ye2023ip,
  title={{IP-Adapter}: Text compatible image prompt adapter for text-to-image diffusion models},
  author={Ye, Hu and Zhang, Jun and Liu, Sibo and Han, Xiao and Yang, Wei},
  journal={arXiv preprint arXiv:2308.06721},
  year={2023}
}

@Article{simswapplusplus,
    author  = {Xuanhong Chen and
              Bingbing Ni and
              Yutian Liu and
              Naiyuan Liu and
              Zhilin Zeng and
              Hang Wang},
    title   = {{SimSwap++}: Towards Faster and High-Quality Identity Swapping},
    journal = {{IEEE} Trans. Pattern Anal. Mach. Intell.},
    volume  = {46},
    number  = {1},
    pages   = {576--592},
    year    = {2024}
}

@article{wu2026loomvideo,
  title={{LoomVideo}: Unifying Multimodal Inputs into Video Generation and Editing},
  author={Wu, Jianzong and Lian, Hao and Yang, Jiongfan and Hao, Dachao and Tian, Ye and Tong, Yunhai and Zhu, Jingyuan and Chen, Biaolong and Qi, Qiaosong and Zhang, Aixi and He, Wanggui and Liu, Mushui and Huang, Pipei and Jiang, Hao},
  journal={arXiv preprint arXiv:2606.06042},
  year={2026}
}

@article{xiao2026joyai,
  title={{JoyAI-Video-Edit}: Real-Time Open-Ended Video Editing with Autoregressive Diffusion},
  author={Xiao, Yicheng and Dai, Wenxun and Qin, Xinran and Song, Lin and Zhang, Maoquan and Xu, Hang and Chen, Yukang and Li, Yitong and Zhang, Guohui and Zhang, Yuan and Zhang, Xuying and Zhang, Tommy and Yuan, Jianlong and Li, Peihao and Lu, Shuai and Fu, Siming and Zhao, Chuyang and Han, Xin and Huang, Jie and Li, Wenbo and Ma, Guoqing and Huang, Wei and Qi, Xiaojuan and Huang, Haoyang and Duan, Nan},
  journal={arXiv preprint arXiv:2608.03974},
  year={2026}
}

@article{wang2025directswap,
  title={{DirectSwap}: Mask-Free Cross-Identity Training and Benchmarking for Expression-Consistent Video Head Swapping},
  author={Wang, Yanan and Liao, Shengcai and Hu, Panwen and Li, Xin and Yang, Fan and Liang, Xiaodan},
  journal={arXiv preprint arXiv:2512.09417},
  year={2025}
}

@misc{fang2024vivid,
        title={{ViViD}: Video Virtual Try-on using Diffusion Models},
        author={Zixun Fang and Wei Zhai and Aimin Su and Hongliang Song and Kai Zhu and Mao Wang and Yu Chen and Zhiheng Liu and Yang Cao and Zheng-Jun Zha},
        year={2024},
        eprint={2405.11794},
        archivePrefix={arXiv},
        primaryClass={cs.CV}
  }

@inproceedings{shin2025large,
  title={Large-scale text-to-image model with inpainting is a zero-shot subject-driven image generator},
  author={Shin, Chaehun and Choi, Jooyoung and Kim, Heeseung and Yoon, Sungroh},
  booktitle={2025 IEEE/CVF Conference on Computer Vision and Pattern Recognition (CVPR)},
  pages={7986--7996},
  year={2025},
  organization={IEEE}
}

@inproceedings{ding2024freecustom,
  title={{FreeCustom}: Tuning-Free Customized Image Generation for Multi-Concept Composition},
  author={Ganggui Ding and Canyu Zhao and Wen Wang and Zhen Yang and Zide Liu and Hao Chen and Chunhua Shen},
  booktitle={Proceedings of the IEEE/CVF Conference on Computer Vision and Pattern Recognition},
  year={2024}
}

@article{fan2024refdrop,
    title={{RefDrop}: Controllable Consistency in Image or Video Generation via Reference Feature Guidance},
    author={Fan, Jiaojiao and Xue, Haotian and Zhang, Qinsheng and Chen, Yongxin},
    journal={arXiv preprint arXiv:2405.17661},
    year={2024}
}

@article{hertz2022prompt,
  title={Prompt-to-prompt image editing with cross attention control},
  author={Hertz, Amir and Mokady, Ron and Tenenbaum, Jay and Aberman, Kfir and Pritch, Yael and Cohen-Or, Daniel},
  journal={arXiv preprint arXiv:2208.01626},
  year={2022}
}

@inproceedings{cao2023masactrl,
  title={{MasaCtrl}: Tuning-free mutual self-attention control for consistent image synthesis and editing},
  author={Cao, Mingdeng and Wang, Xintao and Qi, Zhongang and Shan, Ying and Qie, Xiaohu and Zheng, Yinqiang},
  booktitle={2023 IEEE/CVF International Conference on Computer Vision (ICCV)},
  pages={22503--22513},
  year={2023},
  organization={IEEE}
}

@inproceedings{zhang2026group,
  title={Group Relative Attention Guidance for Image Editing},
  author={Zhang, Xuanpu and Niu, Xuesong and Chen, Ruidong and Song, Dan and Zeng, Jianhao and Du, Penghui and Cao, Haoxiang and Wu, Kai and Liu, An-an},
  booktitle={Proceedings of the IEEE/CVF Conference on Computer Vision and Pattern Recognition},
  pages={3840--3850},
  year={2026}
}

@article{li2026dual,
  title={Dual-Channel Attention Guidance for Training-Free Image Editing Control in Diffusion Transformers},
  author={Li, Guandong},
  journal={arXiv preprint arXiv:2602.18022},
  year={2026}
}

@inproceedings{kim2022adaface,
  title={{AdaFace}: Quality adaptive margin for face recognition},
  author={Kim, Minchul and Jain, Anil K and Liu, Xiaoming},
  booktitle={2022 IEEE/CVF conference on computer vision and pattern recognition (CVPR)},
  pages={18729--18738},
  year={2022},
  organization={IEEE}
}

@article{oquab2023dinov2,
  title={{DINOv2}: Learning robust visual features without supervision},
  author={Oquab, Maxime and Darcet, Timoth{\'e}e and Moutakanni, Th{\'e}o and Vo, Huy and Szafraniec, Marc and Khalidov, Vasil and Fernandez, Pierre and Haziza, Daniel and Massa, Francisco and El-Nouby, Alaaeldin and others},
  journal={arXiv preprint arXiv:2304.07193},
  year={2023}
}

@inproceedings{roessler2018faceforensics,
  title={Faceforensics++: Learning to detect manipulated facial images},
  author={Rossler, Andreas and Cozzolino, Davide and Verdoliva, Luisa and Riess, Christian and Thies, Justus and Nie{\ss}ner, Matthias},
  booktitle={Proceedings of the IEEE/CVF international conference on computer vision},
  pages={1--11},
  year={2019}
}

@inproceedings{choi2021viton,
  title={{VITON-HD}: High-Resolution Virtual Try-On via Misalignment-Aware Normalization},
  author={Choi, Seunghwan and Park, Sunghyun and Lee, Minsoo and Choo, Jaegul},
  booktitle={Proc. of the IEEE conference on computer vision and pattern recognition (CVPR)},
  year={2021}
}

@INPROCEEDINGS{Xiao:2010,
author={J. {Xiao} and J. {Hays} and K. A. {Ehinger} and A. {Oliva} and A. {Torralba} },
booktitle={2010 IEEE Computer Society Conference on Computer Vision and Pattern Recognition},
title={{SUN} database: Large-scale scene recognition from abbey to zoo},
year={2010},
volume={},
number={},
pages={3485-3492},
doi={10.1109/CVPR.2010.5539970},
ISSN={1063-6919},
month={June},}

@article{wang2026head,
  title={Head Similarity: Modeling Structured Whole-Head Appearance Beyond Face Recognition},
  author={Wang, Yingfeng and Xiao, Yuxuan and Liao, Shengcai},
  journal={arXiv preprint arXiv:2605.07766},
  year={2026}
}

@article{lugaresi2019mediapipe,
  title={{MediaPipe}: A Framework for Building Perception Pipelines},
  author={Lugaresi, Camillo and Tang, Jiuqiang and Nash, Hadon and McGuire, Mark and Lee, Jongmin and Chang, Chuo-Ling and Yong, Ming Guang and Grundmann, Matthias and Kwatra, Vivek},
  journal={arXiv preprint arXiv:1906.08172},
  year={2019}
}

@inproceedings{zhang2018unreasonable,
  title={The unreasonable effectiveness of deep features as a perceptual metric},
  author={Zhang, Richard and Isola, Phillip and Efros, Alexei A and Shechtman, Eli and Wang, Oliver},
  booktitle={Proceedings of the IEEE conference on computer vision and pattern recognition},
  pages={586--595},
  year={2018}
}

@misc{carion2025sam3segmentconcepts,
      title={{SAM} 3: Segment Anything with Concepts},
      author={Nicolas Carion and Laura Gustafson and Yuan-Ting Hu and Shoubhik Debnath and Ronghang Hu and Didac Suris and Chaitanya Ryali and Kalyan Vasudev Alwala and Haitham Khedr and Andrew Huang and Jie Lei and Tengyu Ma and Baishan Guo and Arpit Kalla and Markus Marks and Joseph Greer and Meng Wang and Peize Sun and Roman Rädle and Triantafyllos Afouras and Effrosyni Mavroudi and Katherine Xu and Tsung-Han Wu and Yu Zhou and Liliane Momeni and Rishi Hazra and Shuangrui Ding and Sagar Vaze and Francois Porcher and Feng Li and Siyuan Li and Aishwarya Kamath and Ho Kei Cheng and Piotr Dollár and Nikhila Ravi and Kate Saenko and Pengchuan Zhang and Christoph Feichtenhofer},
      year={2025},
      eprint={2511.16719},
      archivePrefix={arXiv},
      primaryClass={cs.CV},
      url={https://arxiv.org/abs/2511.16719},
}

@article{Owen_2007,
   title={The pigeonhole bootstrap},
   volume={1},
   ISSN={1932-6157},
   url={http://dx.doi.org/10.1214/07-AOAS122},
   DOI={10.1214/07-aoas122},
   number={2},
   journal={The Annals of Applied Statistics},
   publisher={Institute of Mathematical Statistics},
   author={Owen, Art B.},
   year={2007},
   month=Dec }

@inproceedings{ke2021musiq,
   title={{MUSIQ}: Multi-scale Image Quality Transformer},
   author={Ke, Junjie and Wang, Qifei and Wang, Yilin and Milanfar, Peyman and Yang, Feng},
   booktitle={2021 IEEE/CVF International Conference on Computer Vision (ICCV)},
   pages={5128--5137},
   year={2021},
   organization={IEEE}
 }

@article{li2023dreamedit,
 title={{DreamEdit}: Subject-driven Image Editing},
 author={Li, Tianle and Ku, Max and Wei, Cong and Chen, Wenhu},
 journal={arXiv preprint arXiv:2306.12624},
 year={2023}
}

@article{wang2026dreamswapv,
title={{DreamSwapV}: Mask-guided Subject Swapping for Any Customized Video Editing},
  author={Wang, Weitao and Wang, Zichen and Shen, Hongdeng and Lu, Yulei and Fan, Xirui and Wu, Suhui and Zhang, Jun and Wang, Haoqian and Zhang, Hao},
  journal={arXiv preprint arXiv:2508.14465},
  year={2025}
}

@article{chen2026scalable,
  title={Scalable Minimal-Change Learning for Controllable Image Editing},
  author={Chen, Shuo and Huang, Fengming and Yao, Yu and Gong, Mingming and Liu, Tongliang},
  journal={arXiv preprint arXiv:2610.06021},
  year={2026}
}

@inproceedings{zhou2025saas,
  title={Scale Your Instructions: Enhance the Instruction-Following Fidelity of Unified Image Generation Model by Self-Adaptive Attention Scaling},
  author={Zhou, Chao and Wei, Tianyi and Yu, Nenghai},
  booktitle={2025 IEEE/CVF International Conference on Computer Vision (ICCV)},
  pages={15171--15181},
  year={2025},
  organization={IEEE}
}

@inproceedings{chen2024anyscene,
  title={{AnyScene}: Customized Image Synthesis with Composited Foreground},
  author={Chen, Ruidong and Wang, Lanjun and Nie, Weizhi and Zhang, Yongdong and Liu, An-An},
  booktitle={2024 IEEE/CVF Conference on Computer Vision and Pattern Recognition (CVPR)},
  pages={8724--8733},
  year={2024},
  organization={IEEE}
}

@inproceedings{zhu2025kvedit,
  title={{KV-Edit}: Training-Free Image Editing for Precise Background Preservation},
  author={Zhu, Tianrui and Zhang, Shiyi and Shao, Jiawei and Tang, Yansong},
  booktitle={2025 IEEE/CVF International Conference on Computer Vision (ICCV)},
  pages={16607--16617},
  year={2025},
  organization={IEEE}
}

@inproceedings{zhang2025boow,
  title={{BooW-VTON}: Boosting In-the-Wild Virtual Try-On via Mask-Free Pseudo Data Training},
  author={Zhang, Xuanpu and Song, Dan and Zhan, Pengxin and Chang, Tianyu and Zeng, Jianhao and Chen, Qingguo and Luo, Weihua and Liu, An-An},
  booktitle={2025 IEEE/CVF Conference on Computer Vision and Pattern Recognition (CVPR)},
  pages={26399--26408},
  year={2025},
  organization={IEEE}
}

@inproceedings{xie2022vfhq,
  title={{VFHQ}: A High-Quality Dataset and Benchmark for Video Face Super-Resolution},
  author={Xie, Liangbin and Wang, Xintao and Zhang, Honglun and Dong, Chao and Shan, Ying},
  booktitle={2022 IEEE/CVF Conference on Computer Vision and Pattern Recognition Workshops (CVPRW)},
  pages={656--665},
  year={2022},
  organization={IEEE}
}
\bibliographystyle{iclr2027_conference}
\clearpage
\appendix
\section*{Appendix}

\section{Implementation and statistical details}
\label{app:protocols}
\paragraph{Default implementation settings.}
Unless otherwise specified in ablations or diagnostic experiments, we use \cref{eq:gap} with $(\gamma_{\mathrm{edit}},\gamma_{\mathrm{keep}})=(0.6,0.8)$. Using zero-based indices, correction is applied to layers 15--29 on LoomVideo and VACE, and 12--24 on FLUX.2-klein (25 layers). All layers are corrected on JoyAI-Video-Edit (40), FLUX.2-klein-base (25), Qwen-Image-Edit (60) and OmniGen2 (32). The edit mask partitions content queries only; no additional mask is applied within the reference-key group. Base and \gap use identical pipeline settings and seeds, with the correction disabled for the base.

\begin{table}[!htb]
\centering\scriptsize
\setlength{\tabcolsep}{3pt}
\begin{tabular}{lc|ccc|ccc|cc}
\toprule
 & & \multicolumn{3}{c|}{Denoising (s/sample)} & \multicolumn{3}{c|}{End-to-end (s/sample)} & \multicolumn{2}{c}{Peak memory} \\
Approach & Steps & Base & \gap & $\Delta$ & Base & \gap & $\Delta$ & Base (GB) & $\Delta$ (MiB) \\
\midrule
VACE & 30 & 158.2 & 159.0 & $+0.5\%$ & 159.4 & 160.0 & $+0.4\%$ & 26.30 & $+0$ \\
LoomVideo & 50 & 106.2 & 108.9 & $+2.5\%$ & 112.0 & 116.6 & $+4.1\%$ & 43.95 & $+1.5$ \\
Qwen-Image-Edit & 20 & 62.48 & 66.00 & $+5.6\%$ & 64.0 & 67.4 & $+5.3\%$ & 57.95 & $+0$ \\
OmniGen2 & 50 & 21.68 & 23.39 & $+7.9\%$ & 22.4 & 24.2 & $+8.0\%$ & 15.36 & $+0$ \\
FLUX.2-klein-base & 50 & 16.35 & 20.34 & $+24\%$ & 17.0 & 21.2 & $+25\%$ & 15.55 & $+0$ \\
FLUX.2-klein & 4 & 0.53 & 0.89 & $+68\%$ & 1.4 & 1.6 & $+14\%$ & 15.54 & $+75$ \\
\midrule
JoyAI-Video-Edit$^{\dagger}$ & 2 & --- & --- & --- & 12.2 & 18.8 & $+54\%$ & 55.9$^{\ddagger}$ & $+1400^{\ddagger}$ \\
\bottomrule
\end{tabular}
\caption{\textbf{Inference cost of \gap on head swapping.} Means over five timed samples after one warm-up, Base and \gap back to back on the same A100-80GB GPU with the deployed settings. Denoising time excludes encoding, decoding, and saving; end-to-end time excludes model loading. Peak memory is the PyTorch allocator's maximum allocation. $^{\dagger}$Unfused reference implementation that materializes the attention matrix in corrected layers; its streaming server reports no per-step timing. $^{\ddagger}$Whole-GPU peak from \texttt{nvidia-smi}.}
\label{tab:supp_compute}
\end{table}

\paragraph{Compute.}
Our fused implementation computes attention separately over reference and non-reference keys, then combines the outputs using log-sum-exp statistics that also yield the reference mass, without materializing the attention matrix. \Cref{tab:supp_compute} reports costs for all seven approaches under the head-swap settings in \cref{tab:main}, including deployed layers, coefficients, resolutions, and step counts. Base and \gap run consecutively on the same otherwise idle A100-80GB GPU, with one warm-up clip followed by five timed validation clips. Denoising time comes from sampler logs; end-to-end time is measured between consecutive saved outputs, excluding model loading, at one-second resolution. Except for JoyAI-Video-Edit, peak memory is measured in-process as PyTorch's peak allocated bytes.

Fused denoising overhead is $0.5$--$7.9\%$ on VACE, LoomVideo, Qwen-Image-Edit, and OmniGen2, and $24\%$ and $68\%$ on FLUX.2-klein-base and FLUX.2-klein, respectively. For the FLUX.2 variants, short image sequences make attention inexpensive relative to the additional kernel call and output combination; FLUX.2-klein's overhead is $0.37$\,s per image over four steps. Peak allocated memory is unchanged on four of these six approaches and increases by 1.5 MiB on LoomVideo and 75 MiB on FLUX.2-klein.

JoyAI-Video-Edit instead uses an unfused hook that materializes per-head fp32 attention matrices. This implementation incurs $54\%$ end-to-end overhead and $1.4$\,GB additional peak memory. Because its serving process was not instrumented, memory is measured at the whole-GPU level using \texttt{nvidia-smi}, rather than PyTorch allocation statistics. 

\paragraph{Setup cost.}
We estimate setup cost from the end-to-end times in \cref{tab:supp_compute}, using the $40$-clip head-swap validation split. Both \gap and constant-bias alternatives need parameter selection; \gap performs it once and keeps the coefficients fixed. Its calibration evaluates $18$ configurations on the four development approaches, about $32$ GPU hours ($18\times40$ clips $\times$ the summed \gap time of $158.2$\,s per clip over the four approaches). Adding a new approach then requires one base profiling run to select layers, $0.02$ to $1.8$ GPU hours from FLUX.2-klein to VACE ($40$ clips $\times$ $1.4$ to $159.4$\,s); the layer-rule statistics come from the same run and need no extra generation.

A per-approach constant instead repeats its sweep on every approach: the $6$-value edit-side sweep of \cref{tab:abl_fixed} costs $0.1$ to $11$ GPU hours per approach ($6\times40$ clips $\times$ $1.6$ to $160.0$\,s), and a $25$-pair edit/keep grid (\cref{app:shared_const}) costs $0.4$ to $44$. A shared constant is selected once, like \gap, at about $44$ GPU hours for the same $25$-pair grid on the four development approaches ($25\times40$ clips $\times$ $158.2$\,s), and needs the same profiling run.

\paragraph{Numerical safeguards.}
For numerical stability, we floor both region-averaged reference-attention masses, $\bar\rho_{\mathrm{edit}}$ and $\bar\rho_{\mathrm{keep}}$, at $10^{-6}$ before computing the logarithms in the bias rule. We then clip the edit-side bias to $[0,8]$; the keep-side bias is not clipped. The $\epsilon_B$ in \cref{eq:band_ratio} stabilizes only the layer-selection ratio and is not used in computing the attention biases.

\section{Reference-mass diagnosis across approaches}
\label{sec:diag}

\Cref{tab:supp_diag} supplements the normalized diagnostics
in \cref{tab:normalized_diag} with absolute reference-attention
masses and the layer-selection statistics $B$ and
$\bar b_{\mathrm{edit}}$ for all seven base models on the
head-swap validation split. The uniform reference-token share
$\pi_{\mathcal R}$ and normalized masses $u_g$ follow
\cref{eq:uniform_share}. We omit these quantities for
JoyAI-Video-Edit because its reference-token share varies
with the streaming KV cache.

The mean log-inverse edit mass $\bar b_{\mathrm{edit}}$
ranges from $2.47$ to $5.91$, indicating substantial variation
in reference-attention allocation across approaches.
This statistic summarizes the potential correction strength
for layer selection; it is not the mean deployed offset,
which is computed online using \cref{eq:gap}.

\begin{table}[b]
\centering
\setlength{\tabcolsep}{4pt}
\begin{tabular}{lccccccc}
\toprule
Approach & $\pi_{\mathcal R}$
& $\bar\rho_{\mathrm{edit}}$ & $u_{\mathrm{edit}}$
& $\bar\rho_{\mathrm{keep}}$ & $u_{\mathrm{keep}}$
& $B$ & $\bar b_{\mathrm{edit}}$ \\
\midrule
JoyAI-Video-Edit  & n/a  & .0615 & ---  & .0555 & ---  & 2.30 & 2.96 \\
LoomVideo        & .031 & .0069 & 0.22 & .0040 & 0.13 & 2.32 & 5.91 \\
FLUX.2-klein     & .286 & .0988 & 0.35 & .0290 & 0.10 & 4.48 & 2.68 \\
FLUX.2-klein-base & .286 & .0635 & 0.22 & .0433 & 0.15 & 1.88 & 3.24 \\
VACE             & .033 & .0590 & 1.79 & .0228 & 0.69 & 5.99 & 3.21 \\
Qwen-Image-Edit   & .332 & .0990 & 0.30 & .1157 & 0.35 & 1.11 & 2.47 \\
OmniGen2         & .327 & .0783 & 0.24 & .0916 & 0.28 & 1.04 & 2.70 \\
\bottomrule
\end{tabular}
\caption{\textbf{Full reference-attention diagnostics.}
Measurements from seven base models on the head-swap
validation split. $B$ is the peak edit-to-keep
reference-mass ratio (\cref{eq:band_ratio});
$\bar b_{\mathrm{edit}}$ is the mean log-inverse edit mass
(\cref{eq:mean_bias}). JoyAI-Video-Edit's uniform baseline
and normalized masses are omitted because its key
composition varies during streaming.}
\label{tab:supp_diag}
\end{table}
\section{Calibration and additional ablations}
\label{app:ablations}

\subsection{Calibration grids}
\label{app:moved}

We jointly evaluate layer coverage $\{\text{all},\text{second half}\}$, $\gamma_{\mathrm{edit}}\in\{0.4,0.6,0.8\}$, and $\gamma_{\mathrm{keep}}\in\{0,0.4,0.8\}$ on the head-swap validation split, yielding $18$ configurations per development approach (\cref{tab:supp_calib}). The coefficients are selected jointly and shared across approaches and tasks.

Under the rule-selected layer sets, four coefficient pairs satisfy the preservation criterion and improve identity on all four approaches: $\gamma_{\mathrm{edit}}\in\{0.4,0.6\}$ and $\gamma_{\mathrm{keep}}\in\{0.4,0.8\}$. Increasing $\gamma_{\mathrm{edit}}$ to $0.8$ degrades JoyAI-Video-Edit below its base identity score and violates preservation on JoyAI-Video-Edit and LoomVideo, while removing keep-side correction ($\gamma_{\mathrm{keep}}=0$) violates preservation at both remaining edit coefficients on FLUX.2 variants, and at $\gamma_{\mathrm{edit}}=0.6$ on LoomVideo.

Among the four feasible pairs, $\gamma_{\mathrm{edit}}=0.6$ gives higher identity similarity than $\gamma_{\mathrm{edit}}=0.4$ on every approach at either keep coefficient. At $\gamma_{\mathrm{edit}}=0.6$, increasing $\gamma_{\mathrm{keep}}$ from $0.4$ to $0.8$ lowers non-edit LPIPS on all four approaches. Identity changes by at most $0.001$ on all four. We therefore select $(\gamma_{\mathrm{edit}},\gamma_{\mathrm{keep}})=(0.6,0.8)$ as a shared fidelity-preservation compromise. The lower block reports transfer to approaches excluded from coefficient selection. These ablations support keep-side suppression but do not isolate its mass-dependent form; \cref{app:shared_const} tests the online form of both sides jointly against a shared constant edit/keep pair.

\begin{table}[t]\centering\scriptsize
\setlength{\tabcolsep}{4.5pt}
\begin{tabular}{@{}l|cc|cc|cc@{}}
\toprule
 & \multicolumn{2}{c|}{$\gamma_{\mathrm{keep}}{=}0$} & \multicolumn{2}{c|}{$\gamma_{\mathrm{keep}}{=}0.4$} & \multicolumn{2}{c}{$\gamma_{\mathrm{keep}}{=}0.8$} \\
$\gamma_{\mathrm{edit}}$ & ID$\uparrow$ & LPIPS$_{\mathrm{ne}}$$\downarrow$ & ID & LPIPS$_{\mathrm{ne}}$ & ID & LPIPS$_{\mathrm{ne}}$ \\
\midrule
\multicolumn{7}{@{}l}{\textbf{JoyAI-Video-Edit} ($B{=}2.3$); base ID $.557$, LPIPS$_{\mathrm{ne}}$ $.0663$} \\
\multicolumn{7}{@{}l}{\quad \emph{all layers}~$\star$} \\
$0.4$ & .689 & .0615 & .683 & .0625 & .682 & .0618 \\
$0.6$ & .705 & .0635 & .705 & .0644 & \textbf{.706} & \textbf{.0641} \\
$0.8$ & .548 & .0913$^{\times}$ & .543 & .0914$^{\times}$ & .549 & .0917$^{\times}$ \\
\multicolumn{7}{@{}l}{\quad \emph{second half}} \\
$0.4$ & .652 & .0639 & .651 & .0638 & .652 & .0644 \\
$0.6$ & .675 & .0678 & .675 & .0678 & .674 & .0674 \\
$0.8$ & .675 & .0784$^{\times}$ & .675 & .0780$^{\times}$ & .676 & .0788$^{\times}$ \\
\midrule
\multicolumn{7}{@{}l}{\textbf{LoomVideo} ($B{=}2.3$, $\bar b{=}5.91$); base ID $.653$, LPIPS$_{\mathrm{ne}}$ $.0700$} \\
\multicolumn{7}{@{}l}{\quad \emph{second half}~$\star$} \\
$0.4$ & .742 & .0696 & .739 & .0681 & .739 & .0666 \\
$0.6$ & .768 & .0781$^{\times}$ & .767 & .0746 & \textbf{.767} & \textbf{.0727} \\
$0.8$ & .771 & .1317$^{\times}$ & .769 & .1281$^{\times}$ & .769 & .1262$^{\times}$ \\
\multicolumn{7}{@{}l}{\quad \emph{all layers}} \\
$0.4$ & .766 & .1306$^{\times}$ & .768 & .1250$^{\times}$ & .789 & .1139$^{\times}$ \\
$0.6$ & .824 & .3187$^{\times}$ & .825 & .3094$^{\times}$ & .826 & .2781$^{\times}$ \\
$0.8$ & .847 & .3334$^{\times}$ & .847 & .3150$^{\times}$ & .847 & .2854$^{\times}$ \\
\midrule
\multicolumn{7}{@{}l}{\textbf{FLUX.2-klein} ($B{=}4.48$); base ID $.807$, LPIPS$_{\mathrm{ne}}$ $.0435$} \\
\multicolumn{7}{@{}l}{\quad \emph{second half}~$\star$} \\
$0.4$ & .882 & .0497$^{\times}$ & .882 & .0395 & .882 & .0387 \\
$0.6$ & .885 & .0524$^{\times}$ & .885 & .0416 & \textbf{.885} & \textbf{.0406} \\
$0.8$ & .889 & .0561$^{\times}$ & .889 & .0438 & .889 & .0428 \\
\multicolumn{7}{@{}l}{\quad \emph{all layers}} \\
$0.4$ & .888 & .1079$^{\times}$ & .892 & .0484 & .892 & .0445 \\
$0.6$ & .888 & .1779$^{\times}$ & .893 & .0644$^{\times}$ & .893 & .0545$^{\times}$ \\
$0.8$ & .890 & .1918$^{\times}$ & .891 & .0780$^{\times}$ & .891 & .0686$^{\times}$ \\
\midrule
\multicolumn{7}{@{}l}{\textbf{FLUX.2-klein-base} ($B{=}1.88$); base ID $.590$, LPIPS$_{\mathrm{ne}}$ $.1206$} \\
\multicolumn{7}{@{}l}{\quad \emph{all layers}~$\star$} \\
$0.4$ & .797 & .1900$^{\times}$ & .866 & .0687 & .864 & .0654 \\
$0.6$ & .879 & .2115$^{\times}$ & .877 & .0764 & \textbf{.878} & \textbf{.0674} \\
$0.8$ & .873 & .2273$^{\times}$ & .877 & .0979 & .875 & .0835 \\
\multicolumn{7}{@{}l}{\quad \emph{second half}} \\
$0.4$ & .704 & .1373$^{\times}$ & .773 & .0941 & .773 & .0914 \\
$0.6$ & .791 & .1439$^{\times}$ & .812 & .0958 & .826 & .0932 \\
$0.8$ & .824 & .1434$^{\times}$ & .844 & .0996 & .842 & .0957 \\
\midrule

\multicolumn{7}{@{}l}{\emph{Held out from coefficient selection:
$(\gamma_{\mathrm{edit}},\gamma_{\mathrm{keep}})=(0.6,0.8)$}} \\
\midrule
VACE ($6.0$; half) & \multicolumn{3}{c|}{ID $0.484\rightarrow\textbf{0.810}$} & \multicolumn{3}{c}{LPIPS$_{\mathrm{ne}}$ $0.045\rightarrow0.044$\ \checkmark} \\
Qwen-Image-Edit  ($1.11$; all) & \multicolumn{3}{c|}{ID $0.455\rightarrow\textbf{0.787}$} & \multicolumn{3}{c}{LPIPS$_{\mathrm{ne}}$ $0.296\rightarrow0.167$\ \checkmark} \\
OmniGen2 ($1.04$; all) & \multicolumn{3}{c|}{ID $0.413\rightarrow\textbf{0.841}$} & \multicolumn{3}{c}{LPIPS$_{\mathrm{ne}}$ $0.237\rightarrow0.124$\ \checkmark} \\
\bottomrule
\end{tabular}
\caption{\textbf{Joint calibration of correction coefficients and layer coverage.} Head-swap validation results. $\star$ marks rule-selected layers; bold marks the chosen configuration. LPIPS$_{\mathrm{ne}}$ denotes non-edit LPIPS, and $^{\times}$ marks an increase over the base exceeding $0.005$. }
\label{tab:supp_calib}
\end{table}

\subsection{Online versus fixed edit-side strength}
\label{app:fixedb}

We compare online correction with approach-specific fixed
edit-side biases on the 40-clip head-swap validation split
using the four development approaches. We sweep
$b\in\{0.5,1,1.5,2,2.5,3\}$ while keeping masks, keep-side
correction, and layer sets unchanged. For each approach,
we select the bias with the highest validation identity
similarity satisfying
$\Delta\mathrm{LPIPS}_{\mathrm{ne}}\leq0.005$
(\cref{tab:supp_matrix}).
The selected biases are $1.5$ for JoyAI-Video-Edit,
$2.5$ for LoomVideo, and $2.0$ for both FLUX.2 variants.
Their test-set results are reported in \cref{tab:abl_fixed}. \Cref{app:shared_const} makes both sides constant and shares them across approaches.

\begin{table}
\centering\small
\setlength{\tabcolsep}{3.2pt}
{\begin{tabular}{l|cc|cc|cc|cc}
\toprule
& \multicolumn{2}{c|}{JoyAI-Video-Edit}
& \multicolumn{2}{c|}{LoomVideo}
& \multicolumn{2}{c|}{FLUX.2-klein}
& \multicolumn{2}{c}{FLUX.2-klein-base} \\

& \multicolumn{2}{c|}{all}
& \multicolumn{2}{c|}{second half}
& \multicolumn{2}{c|}{second half}
& \multicolumn{2}{c}{all} \\
$b$
& ID$\uparrow$ & LPIPS$_{\mathrm{ne}}$$\downarrow$
& ID$\uparrow$ & LPIPS$_{\mathrm{ne}}$$\downarrow$
& ID$\uparrow$ & LPIPS$_{\mathrm{ne}}$$\downarrow$
& ID$\uparrow$ & LPIPS$_{\mathrm{ne}}$$\downarrow$ \\
\midrule
base
& .557 & .0663
& .653 & .0700
& .807 & .0435
& .590 & .1206 \\
\midrule
0.5
& .647 & .0641
& .661 & .0681
& .851 & .0377
& .806 & .0667 \\
1.0
& .683 & .0625
& .689 & .0676
& .883 & .0394
& .861 & .0659 \\
1.5
& \textbf{.703} & \textbf{.0616}
& .717 & .0669
& .888 & .0410
& .880 & .0667 \\
2.0
& .625 & .0761$^{\times}$
& .744 & .0670
& \textbf{.889} & \textbf{.0432}
& \textbf{.881} & \textbf{.0698} \\
2.5
& --- & ---
& \textbf{.761} & \textbf{.0697}
& .888 & .0474
& .871 & .0763 \\
3.0
& --- & ---
& .774 & .0790$^{\times}$
& .888 & .0544$^{\times}$
& .873 & .0898 \\
\bottomrule
\end{tabular}}
\caption{\textbf{Fixed edit-side sweeps that select each approach's constant}
(head-swap validation, $40$ clips). $^{\times}$ marks
$\Delta\mathrm{LPIPS}_{\mathrm{ne}}>0.005$ relative to the base; bold marks the
best admissible constant, the highest-identity value that satisfies the
criterion, which is the constant used in
\cref{tab:abl_fixed}. }
\label{tab:supp_matrix}
\end{table}

\section{Task protocols and additional results}
\label{app:results}

\subsection{Head swapping}
\label{app:headswap}

\paragraph{Protocol and metrics.}
HeadSwapBench~\citep{wang2025directswap} contains $1040$ test clips and a disjoint validation split of $40$ clips. Each clip contains $121$ frames at $512\times512$ resolution. Inputs comprise a driving video and a reference head image; VACE receives a head-masked driving video. Image approaches are edited and evaluated on the middle frame.

Because head swapping replaces the entire head, including the hair, whereas AdaFace focuses only on the face, we additionally report whole-head similarity~\citep{wang2026head}, which measures the cosine similarity between head-appearance embeddings of the generated result and the ground truth. Keypoint error uses $478$ landmarks and is normalized by inter-ocular distance.

\paragraph{Task-fine-tuned model: DirectSwap.}
While the main paper evaluates general-purpose editors, we further apply RefGAP to a task-specific fine-tuned model. We evaluate DirectSwap~\citep{wang2025directswap} in its released inference pipeline, including its attention-based background compositing. For this mask-free model, we define edit queries from the reference-attention mass of the first denoising step, averaged over all attention heads: in each of layers 8–26 (zero-based), a query casts a vote if its mass lies above the 0.7 quantile of its latent frame, and it is an edit query if more than 20\% of these layers vote for it. These are the defaults of DirectSwap's released background-compositing mask, so RefGAP adds no threshold of its own. The first step runs unmodified; the region is then fixed, and RefGAP acts on layers 15–29 for the remaining 29 of 30 sampling steps.

As shown in \cref{tab:supp_directswap_test}, \gap increases identity similarity from $0.7019$ to $0.7450$. These gains accompany increased keypoint error, while whole-frame LPIPS changes little. The results show that reference-attention correction can also improve identity fidelity in a task-fine-tuned model, with remaining attribute-preservation trade-offs.

\begin{table}[t]\centering\small\setlength{\tabcolsep}{4pt}
\begin{tabular}{l|ccc}
\toprule
& ID (AdaFace)$\uparrow$
& Kpt.$\downarrow$
& LPIPS$\downarrow$ \\
\midrule
DirectSwap
& $0.7019 \rightarrow 0.7450$
& $0.0289 \rightarrow 0.0371$
& $0.1333 \rightarrow 0.1339$ \\
\bottomrule
\end{tabular}
\caption{\textbf{DirectSwap on the HeadSwapBench test split.}
Scores are reported as base $\rightarrow$ \gap using clip-level averaging.
ID and Kpt.\ denote identity similarity and keypoint error, respectively.
LPIPS is measured over the whole frame.}
\label{tab:supp_directswap_test}
\end{table}

\paragraph{Whole-head similarity.}
\Cref{tab:supp_headsim} complements the identity similarity results in \cref{tab:main}. Whole-head similarity increases on all seven general-purpose approaches.

\begin{table}[H]
\centering
\scriptsize
\setlength{\tabcolsep}{3pt}
\resizebox{\linewidth}{!}{
\begin{tabular}{l|ccccccc|c}
\toprule
 & JoyAI-Video-Edit & LoomVideo & FLUX.2-klein & FLUX.2-klein-base
 & VACE & Qwen-Image-Edit & OmniGen2 & DirectSwap \\
\midrule
base
 & 0.809 & 0.816 & 0.861 & 0.757
 & 0.786 & 0.734 & 0.581 & 0.894 \\
RefGAP
 & \textbf{0.876} & \textbf{0.861} & \textbf{0.887} & \textbf{0.875}
 & \textbf{0.888} & \textbf{0.802} & \textbf{0.851} & 0.894 \\
win \%
 & 73 & 73 & 61 & 76
 & 96 & 63 & 96 & 49 \\
\bottomrule
\end{tabular}
}
\caption{\textbf{Whole-head similarity on HeadSwapBench.}
Win \% is the percentage of paired valid clips on which RefGAP outperforms the base.
DirectSwap is included as a task-fine-tuned reference.}
\label{tab:supp_headsim}
\end{table}

\subsection{Face swapping}
\label{app:faceswap}

\paragraph{Inputs and preprocessing.}
We use the c23-compressed real videos from FaceForensics++~\citep{roessler2018faceforensics}, following the $1000$ official target--source pairings of its FaceSwap subset. Both inputs are real videos; manipulated videos are used only to identify the official pairings. VACE receives a face-masked driving video.

Each target clip contains the first $33$ frames. We apply a fixed face-centered square crop with side length twice the median face height across the clip and resize it to $512\times512$. The edit mask undergoes the same transformation. Image approaches edit the middle frame.

We select a reference frame with a near-neutral expression from the first $300$ source frames, using MediaPipe blendshape scores to penalize facial expressions, particularly jaw opening and blinking. If no face is detected ($21$ pairs), we use the first frame.

\paragraph{Why face-centered crops.}
In the full $640\times480$ frames the face occupies a small part of the image, so at the resolutions the approaches process it spans few latent tokens: the edit has little room to render the source identity, and mouth and eye details degrade. The crop places the face at roughly a quarter of the image area, matching the head scale of the head-swap benchmark. We compared the two inputs on JoyAI-Video-Edit with the same neutral references (\cref{tab:supp_ffpp_crop}). On these pairs the crop raises retrieval and identity similarity for both arms and lowers keypoint. All face-swap results in the paper use the crops.

\begin{table}[H]\centering\scriptsize\setlength{\tabcolsep}{4pt}
\begin{tabular}{l|cc|cc|cc}
\toprule
& \multicolumn{2}{c|}{ID-Retr.\ (\%) $\uparrow$}
& \multicolumn{2}{c|}{ID-sim $\uparrow$}
& \multicolumn{2}{c}{Kpt.\ $\downarrow$} \\
input & base & \gap & base & \gap & base & \gap \\
\midrule
full frame ($640\times480$)
& 25.0 & 50.0
& .109 & .240
& .085 & .108 \\
face-centered crop ($512^2$)
& 40.0 & \textbf{80.0}
& .248 & \textbf{.434}
& .058 & .074 \\
\bottomrule
\end{tabular}
\caption{\textbf{Full-frame versus face-centered inputs for face swapping}}
\label{tab:supp_ffpp_crop}
\end{table}

\subsection{Virtual try-on}
\label{app:tryonbg}

\paragraph{Datasets and inputs.}
Our main evaluation uses $90$ ViViD videos~\citep{fang2024vivid}. Each clip contains $65$ frames at $624\times832$ resolution ($640\times832$ for LoomVideo). We additionally evaluate image approaches on $100$ VITON-HD images~\citep{choi2021viton}.

Both evaluations use an unpaired setting: reference garments are permuted within category, and the model is instructed to replace the input garment with the reference garment. Models receive the unmasked input, except VACE, which uses its masked-input interface. For VITON-HD, inputs to FLUX.2-klein and FLUX.2-klein-base are padded to a square canvas, and outputs are cropped back before evaluation.

\paragraph{Additional evaluation on VITON-HD.}
\label{app:tryonmid}
We further evaluate the four image approaches on $100$ VITON-HD images using the same coefficients and layer sets. As shown in \cref{tab:supp_tryonmid}, DINO improves on all four approaches by $0.009$--$0.046$, with LPIPS increases of at most $0.0056$. These results support transfer to an additional image dataset without retuning.

\begin{table}[H]\centering\scriptsize\setlength{\tabcolsep}{3pt}
{\begin{tabular}{lcc}
\toprule
Approach & DINO $\uparrow$ & LPIPS $\downarrow$ \\
\midrule
FLUX.2-klein
& 0.382 $\rightarrow$ 0.391
& 0.0302 $\rightarrow$ 0.0318 \\
FLUX.2-klein-base
& 0.407 $\rightarrow$ 0.426
& 0.0415 $\rightarrow$ 0.0417 \\
Qwen-Image-Edit
& 0.216 $\rightarrow$ 0.225
& 0.0158 $\rightarrow$ 0.0160 \\
OmniGen2
& 0.397 $\rightarrow$ 0.443
& 0.1339 $\rightarrow$ 0.1395 \\
\bottomrule
\end{tabular}}
\caption{\textbf{Virtual try-on on $100$ VITON-HD images.}
Scores are shown as base $\rightarrow$ \gap.}
\label{tab:supp_tryonmid}
\end{table}

\subsection{Background replacement}
\label{app:bgswap}

\paragraph{Inputs and preprocessing.}
We use $150$ ground-truth videos from $58$ subjects in the HeadSwapBench test split, selecting static-camera clips containing a single person. Each clip is paired with a person-free scene reference from a set of $30$ SUN397 images~\citep{Xiao:2010}. SAM3~\citep{carion2025sam3segmentconcepts} person masks define the subject and background regions. Video approaches edit the original clips, while image approaches edit their middle frames. All approaches use the same clip-reference pairs.

\section{Uncertainty of the main quantitative results}
\label{app:uncertainty}

We quantify uncertainty in the Base-to-Ours differences in
\cref{tab:main} using paired bootstrap intervals that account for
shared identities and reference content. Metric definitions and
task-specific protocols are provided in \cref{app:results}.

\paragraph{Paired estimates.}
Base and \gap use identical inputs, seeds, and masks.
For each metric, we compute the clip-level difference
$d_i = m_i^{\mathrm{Ours}} - m_i^{\mathrm{Base}}$ and average
over the eligible paired observations. Frames within a video are
aggregated before resampling; image approaches are evaluated on
the middle frame. Repl.\ and Retr.\ differences are expressed in
percentage points (pp). Estimates and intervals are computed from
unrounded per-clip scores, rather than from the rounded values
displayed in \cref{tab:main}.

\paragraph{Resampling units.}
We use task-specific units to represent dependencies between clips:
\begin{itemize}
    \item \textbf{Head swap:}
    subjects, with 104 identities across 1,040 clips.
    OmniGen2's face-based metrics retain 103 identities.

    \item \textbf{Face swap:}
    a shared pool of target and source identities, accounting
    for identities that occur in both roles.

    \item \textbf{Virtual try-on:}
    input clips and reference garments, with garments assigned
    from other clips in the same category.

    \item \textbf{Background replacement:}
    subject identities and scene references, comprising
    58 people and 30 distinct scene images.
\end{itemize}
Repl.\ is omitted for VACE in both identity tasks because its
masked input does not retain the input identity.

\paragraph{Bootstrap procedure.}
Following the factor-wise resampling construction of the
pigeonhole bootstrap~\citep{Owen_2007}, we resample
the units of each factor with replacement. For crossed factors,
the draws are independent, and observation $i$ receives weight
\[
    w_i = \prod_f c_f(u_{i,f}),
\]
where $c_f(u_{i,f})$ is the sampled multiplicity of its unit
in factor $f$. Each replicate estimates
\[
    \widehat{\Delta}^{*}
    = \frac{\sum_i w_i d_i}{\sum_i w_i}.
\]
For head swap, this reduces to a subject-level cluster bootstrap.
For face swap, we draw multiplicities once from the shared
identity pool and weight each pair by the product of its target
and source multiplicities. We generate 10,000 replicates and
report the 2.5th and 97.5th percentiles. These are pointwise
intervals, not simultaneous intervals across all comparisons.
We also compute an ordinary paired clip bootstrap for comparison.

\begin{table}[t]
\centering\scriptsize
\setlength{\tabcolsep}{2.5pt}
\renewcommand{\arraystretch}{1.05}
\resizebox{\linewidth}{!}{%
\begin{tabular}{ll*{7}{c}}
\toprule
Task & Metric & JoyAI & LoomVideo & Klein & Klein-base & VACE & Qwen & OmniGen2 \\
\midrule
\multirow{4}{*}{Head swap} & Repl.\ (pp)$\uparrow$ & \makecell{\textcolor{gain}{+30.9}\\{\tiny[+21.5, +40.1]}} & \makecell{\textcolor{gain}{+12.2}\\{\tiny[+6.6, +17.6]}} & \makecell{\textcolor{gain}{+8.6}\\{\tiny[+5.1, +12.3]}} & \makecell{\textcolor{gain}{+22.8}\\{\tiny[+17.5, +27.4]}} & --- & \makecell{\textcolor{gain}{+16.1}\\{\tiny[+10.1, +24.1]}} & \makecell{\textcolor{gain}{+65.1}\\{\tiny[+56.7, +71.4]}} \\
 & ID$\uparrow$ & \makecell{\textcolor{gain}{+.137}\\{\tiny[+.111, +.169]}} & \makecell{\textcolor{gain}{+.058}\\{\tiny[+.045, +.069]}} & \makecell{\textcolor{gain}{+.064}\\{\tiny[+.049, +.088]}} & \makecell{\textcolor{gain}{+.211}\\{\tiny[+.176, +.247]}} & \makecell{\textcolor{gain}{+.323}\\{\tiny[+.290, +.362]}} & \makecell{\textcolor{gain}{+.114}\\{\tiny[+.071, +.167]}} & \makecell{\textcolor{gain}{+.486}\\{\tiny[+.449, +.525]}} \\
 & Kpt.$\downarrow$ & \makecell{\textcolor{cost}{+.042}\\{\tiny[+.027, +.059]}} & \makecell{\textcolor{gray}{$-$.009}\\{\tiny[$-$.025, +.002]}} & \makecell{\textcolor{cost}{+.055}\\{\tiny[+.046, +.065]}} & \makecell{\textcolor{cost}{+.068}\\{\tiny[+.048, +.096]}} & \makecell{\textcolor{gray}{+.001}\\{\tiny[$-$.009, +.012]}} & \makecell{\textcolor{gray}{+.000}\\{\tiny[$-$.017, +.015]}} & \makecell{\textcolor{gray}{+.002}\\{\tiny[$-$.019, +.021]}} \\
 & LPIPS$\downarrow$ & \makecell{\textcolor{gray}{$-$.003}\\{\tiny[$-$.010, +.006]}} & \makecell{\textcolor{gray}{$-$.005}\\{\tiny[$-$.013, +.000]}} & \makecell{\textcolor{gray}{$-$.004}\\{\tiny[$-$.011, +.005]}} & \makecell{\textcolor{gain}{$-$.066}\\{\tiny[$-$.093, $-$.037]}} & \makecell{\textcolor{gain}{$-$.013}\\{\tiny[$-$.016, $-$.011]}} & \makecell{\textcolor{gain}{$-$.212}\\{\tiny[$-$.246, $-$.171]}} & \makecell{\textcolor{gain}{$-$.099}\\{\tiny[$-$.117, $-$.079]}} \\
\midrule
\multirow{5}{*}{Face swap} & Repl.\ (pp)$\uparrow$ & \makecell{\textcolor{gain}{+36.8}\\{\tiny[+31.4, +42.6]}} & \makecell{\textcolor{gain}{+13.0}\\{\tiny[+9.4, +17.1]}} & \makecell{\textcolor{gain}{+2.2}\\{\tiny[+0.8, +4.1]}} & \makecell{\textcolor{gain}{+5.1}\\{\tiny[+2.9, +7.8]}} & --- & \makecell{\textcolor{gain}{+16.7}\\{\tiny[+11.7, +21.8]}} & \makecell{\textcolor{gain}{+20.7}\\{\tiny[+16.7, +24.9]}} \\
 & ID$\uparrow$ & \makecell{\textcolor{gain}{+.310}\\{\tiny[+.293, +.327]}} & \makecell{\textcolor{gain}{+.017}\\{\tiny[+.003, +.034]}} & \makecell{\textcolor{gain}{+.116}\\{\tiny[+.102, +.131]}} & \makecell{\textcolor{gain}{+.274}\\{\tiny[+.244, +.305]}} & \makecell{\textcolor{gain}{+.277}\\{\tiny[+.262, +.294]}} & \makecell{\textcolor{gain}{+.169}\\{\tiny[+.137, +.202]}} & \makecell{\textcolor{gain}{+.498}\\{\tiny[+.474, +.521]}} \\
 & Retr.\ (pp)$\uparrow$ & \makecell{\textcolor{gain}{+49.1}\\{\tiny[+43.6, +54.7]}} & \makecell{\textcolor{gain}{+11.3}\\{\tiny[+7.6, +15.3]}} & \makecell{\textcolor{gain}{+6.1}\\{\tiny[+3.6, +9.1]}} & \makecell{\textcolor{gain}{+22.4}\\{\tiny[+17.3, +27.6]}} & \makecell{\textcolor{gain}{+11.7}\\{\tiny[+8.3, +15.4]}} & \makecell{\textcolor{gain}{+20.1}\\{\tiny[+15.0, +25.3]}} & \makecell{\textcolor{gain}{+71.6}\\{\tiny[+66.5, +76.4]}} \\
 & Kpt.$\downarrow$ & \makecell{\textcolor{cost}{+.048}\\{\tiny[+.045, +.052]}} & \makecell{\textcolor{gray}{+.003}\\{\tiny[$-$.002, +.008]}} & \makecell{\textcolor{cost}{+.017}\\{\tiny[+.015, +.019]}} & \makecell{\textcolor{cost}{+.093}\\{\tiny[+.081, +.106]}} & \makecell{\textcolor{cost}{+.006}\\{\tiny[+.001, +.012]}} & \makecell{\textcolor{cost}{+.022}\\{\tiny[+.013, +.030]}} & \makecell{\textcolor{gray}{+.011}\\{\tiny[$-$.001, +.023]}} \\
 & LPIPS$\downarrow$ & \makecell{\textcolor{cost}{+.022}\\{\tiny[+.019, +.026]}} & \makecell{\textcolor{gray}{$-$.001}\\{\tiny[$-$.006, +.003]}} & \makecell{\textcolor{cost}{+.013}\\{\tiny[+.011, +.015]}} & \makecell{\textcolor{cost}{+.006}\\{\tiny[+.001, +.012]}} & \makecell{\textcolor{cost}{+.001}\\{\tiny[+.001, +.001]}} & \makecell{\textcolor{gray}{$-$.004}\\{\tiny[$-$.011, +.001]}} & \makecell{\textcolor{gain}{$-$.044}\\{\tiny[$-$.057, $-$.031]}} \\
\midrule
\multirow{2}{*}{Virtual try-on} & DINO$\uparrow$ & \makecell{\textcolor{gray}{+.016}\\{\tiny[$-$.006, +.039]}} & \makecell{\textcolor{gray}{+.010}\\{\tiny[$-$.001, +.026]}} & \makecell{\textcolor{gray}{+.009}\\{\tiny[$-$.001, +.021]}} & \makecell{\textcolor{gain}{+.056}\\{\tiny[+.032, +.090]}} & \makecell{\textcolor{gain}{+.055}\\{\tiny[+.024, +.092]}} & \makecell{\textcolor{gain}{+.035}\\{\tiny[+.003, +.069]}} & \makecell{\textcolor{gain}{+.078}\\{\tiny[+.024, +.136]}} \\
 & LPIPS$\downarrow$ & \makecell{\textcolor{gray}{+.001}\\{\tiny[$-$.009, +.009]}} & \makecell{\textcolor{gray}{+.001}\\{\tiny[$-$.006, +.004]}} & \makecell{\textcolor{gray}{$-$.001}\\{\tiny[$-$.006, +.002]}} & \makecell{\textcolor{gray}{+.010}\\{\tiny[$-$.001, +.023]}} & \makecell{\textcolor{gray}{$-$.000}\\{\tiny[$-$.000, +.000]}} & \makecell{\textcolor{gray}{+.017}\\{\tiny[$-$.002, +.043]}} & \makecell{\textcolor{gain}{$-$.017}\\{\tiny[$-$.034, $-$.003]}} \\
\midrule
\multirow{2}{*}{Background} & DINO$\uparrow$ & \makecell{\textcolor{gain}{+.045}\\{\tiny[+.014, +.086]}} & \makecell{\textcolor{gain}{+.123}\\{\tiny[+.064, +.187]}} & \makecell{\textcolor{gray}{+.001}\\{\tiny[$-$.008, +.013]}} & \makecell{\textcolor{gray}{+.019}\\{\tiny[$-$.017, +.048]}} & \makecell{\textcolor{gray}{+.006}\\{\tiny[$-$.005, +.020]}} & \makecell{\textcolor{cost}{$-$.257}\\{\tiny[$-$.374, $-$.153]}} & \makecell{\textcolor{gain}{+.270}\\{\tiny[+.151, +.346]}} \\
 & LPIPS$\downarrow$ & \makecell{\textcolor{cost}{+.002}\\{\tiny[+.001, +.003]}} & \makecell{\textcolor{cost}{+.007}\\{\tiny[+.004, +.011]}} & \makecell{\textcolor{gray}{$-$.002}\\{\tiny[$-$.004, +.000]}} & \makecell{\textcolor{gray}{$-$.026}\\{\tiny[$-$.050, +.001]}} & \makecell{\textcolor{gray}{+.000}\\{\tiny[$-$.000, +.000]}} & \makecell{\textcolor{gray}{$-$.004}\\{\tiny[$-$.021, +.011]}} & \makecell{\textcolor{cost}{+.036}\\{\tiny[+.006, +.068]}} \\
\bottomrule
\end{tabular}}
\caption{\textbf{Paired differences with pointwise 95\% bootstrap
confidence intervals.}
Cells show Ours$-$Base and the corresponding interval from
10,000 task-specific resamples.
Green/red indicate intervals excluding zero in the
improving/degrading direction; gray indicates intervals containing
zero. Colors are determined before rounding.
Repl.\ and Retr.\ differences are in percentage points.
VACE Repl.\ is undefined and omitted.
Sample counts vary by metric and detection availability.}
\label{tab:supp_main_ci}
\end{table}

\paragraph{Overall uncertainty.}
\Cref{tab:supp_main_ci} reports 89 paired estimates and intervals.
Relative to clip-level resampling, the dependence-aware procedure
increases interval widths by a median factor of 1.85.
The number of intervals excluding zero decreases from 76 to 62,
indicating that treating clips as independent can overstate the
precision of the estimated effects.

\paragraph{Identity fidelity.}
All 27 intervals for head-swap Repl.\ and ID and face-swap ID
and Retr.\ exclude zero in the improving direction.
This includes LoomVideo's face-swap ID gain of
$+.017$ $[+.003,+.034]$.
Face-swap Repl.\ improves with an interval excluding zero on all six approaches for which it is defined.

\paragraph{Preservation trade-offs.}
The intervals also identify preservation costs.
Head-swap keypoint error increases on JoyAI-Video-Edit and both
FLUX.2 variants. Face-swap keypoint error increases on these
three approaches, Qwen-Image-Edit, and VACE; face-swap LPIPS
increases on JoyAI-Video-Edit, both FLUX.2 variants, and VACE.
Background LPIPS increases on JoyAI-Video-Edit, LoomVideo,
and OmniGen2. Conversely, head-swap LPIPS improves on
Qwen-Image-Edit, FLUX.2-klein-base, VACE, and OmniGen2,
while OmniGen2 also improves face-swap and try-on LPIPS.
Each change listed here has an interval excluding zero.

\paragraph{Transfer beyond identity editing.}
Try-on DINO gains have intervals excluding zero on four
approaches, while the intervals for JoyAI-Video-Edit,
FLUX.2-klein, and LoomVideo include zero.
For background replacement, DINO improves on JoyAI-Video-Edit,
LoomVideo, and OmniGen2 with intervals excluding zero.
The decrease on Qwen-Image-Edit also excludes zero
($-.257$ $[-.374,-.153]$), whereas the intervals for both
FLUX.2 variants and VACE include zero.
Thus, the transfer results support gains on several approaches,
rather than uniform improvements across all settings.

The 14 comparisons whose intervals exclude zero only under
clip-level resampling concern preservation or transfer metrics.
For these comparisons, we report the estimated direction without
claiming a reliably established improvement or degradation.

\section{Human evaluation}
\label{app:qual}

\paragraph{Sampling details.}
For each approach and task, we first form a candidate pool of samples showing large visible differences between the two methods, with method labels hidden, and then randomly select items from this pool using a fixed seed while rotating over subject identities. We evaluate five items per approach for head swapping and ten for each other task, except that FLUX.2-klein contributes only two items each for try-on and background replacement because the two methods produce similar outputs on most samples. VACE background replacement is excluded because neither method performs the edit.

\paragraph{Unscreened audit.}
To measure what the screening leaves out, we repeated the protocol on a form built without it: two items per approach and task drawn at random from each evaluation set ($56$ items, VACE background replacement included), plus two attention checks and two A/B-swapped repeats, answered by five raters. All five passed both checks, agreed with one another on $95$--$99\%$ of answers, and repeated their own judgment on $87\%$ of the swapped repeats. For reference fidelity, $27.9\%$ of judgments favored
\gap, $3.6\%$ favored the base, and $68.6\%$ were ties, giving \gap an
$88.6\%$ share among decided judgments. For visual quality, $92.1\%$ were ties,
while all $22$ decided judgments favored the base. Full results are reported
in \cref{tab:supp_audit}.

The 22 base-favoring quality judgments are concentrated in five items, rather than 22 distinct examples. This is consistent with the base-favoring quality preference observed for LoomVideo face swapping in the screened evaluation (1/33/18, RefGAP/tie/base). However, \cref{tab:main} shows mixed changes for LoomVideo face swapping: identity similarity improves, keypoint error increases slightly, and non-face LPIPS decreases slightly. These metrics therefore do not fully capture the perceived quality differences.
\begin{table}
\centering\small\setlength{\tabcolsep}{5pt}
\begin{tabular}{l|cc|cc|c}
\toprule
& \gap & base & both succeed & both fail & tie \% \\
\midrule
Reference fidelity
& 78 \tiny(27.9\%) & 10 \tiny(3.6\%) & 144 & 48 & 68.6 \\
Pose / expression
& 20 \tiny(7.1\%) & 10 \tiny(3.6\%) & 178 & 72 & 89.3 \\
Visual quality
& 0 \tiny(0.0\%) & 22 \tiny(7.9\%) & 246 & 12 & 92.1 \\
\bottomrule
\end{tabular}
\caption{\textbf{Unscreened audit form.}
Judgment counts from five raters on $56$ randomly sampled items
(two per approach and task), with no screening for visible differences.}
\label{tab:supp_audit}
\end{table}

\paragraph{Quality controls.}
Each form includes two attention checks with an unedited input as one candidate. One of $55$ raters fails a check and is excluded. Three additional identical-output controls receive tie responses in all $45$ judgments.

\paragraph{Additional statistics.}
\Cref{tab:supp_user_pooled} reports item-level preference rates with $95\%$ confidence intervals from $10{,}000$ bootstrap resamples of items. Preference rates exclude ties separately for each question; items with only tied responses to a question are omitted from its preference estimate. The reference-fidelity tie rate is reported separately.

\Cref{tab:supp_user_ties} separates the two responses pooled as ties in \cref{tab:user}. Pose/expression receives $46$ both-fail judgments on head swapping and $138$ on face swapping, showing why ties should not be interpreted as successful preservation.

\begin{table}[htbp]\centering\small
\begin{tabular}{@{}ll|cccc|c@{}}
\toprule
Task & Question & \gap & both succeed & both fail & base & total \\
\midrule
\multirow{3}{*}{Head swap}
 & closer to reference & 157 & 23 & 1 & 3 & 184 \\
 & pose / expression   & 27 & 109 & 46 & 3 & 185 \\
 & visual quality      & 37 & 131 & 7 & 10 & 185 \\
\midrule
\multirow{3}{*}{Face swap}
 & closer to reference & 278 & 89 & 7 & 8 & 382 \\
 & pose / expression   & 19 & 185 & 138 & 40 & 382 \\
 & visual quality      & 32 & 312 & 8 & 30 & 382 \\
\midrule
\multirow{3}{*}{Try-on}
 & closer to reference & 249 & 37 & 13 & 40 & 339 \\
 & pose / expression   & 9 & 274 & 53 & 3 & 339 \\
 & visual quality      & 24 & 298 & 11 & 6 & 339 \\
\midrule
\multirow{3}{*}{Background}
 & closer to reference & 127 & 101 & 20 & 29 & 277 \\
 & pose / expression   & 5 & 217 & 52 & 3 & 277 \\
 & visual quality      & 17 & 242 & 8 & 9 & 276 \\
\bottomrule
\end{tabular}
\caption{\textbf{Breakdown of tied judgments.} The ``both succeed'' and ``both fail'' responses are reported separately and sum to the tie counts in \cref{tab:user}. Entries report the number of rater responses.}
\label{tab:supp_user_ties}
\end{table}

\paragraph{Face-swap subgroup analysis.}
We further group face-swap items by whether the base output is closer to the reference identity or the input identity, excluding two items without a detected face in the base. Reference-fidelity judgments favor \gap in both groups. For pose/expression, the base receives $25$ preferences versus $3$ for \gap when its output is closer to the input identity. When it is closer to the reference identity, preferences are nearly balanced ($16$ for \gap versus $15$ for the base, with $234$ ties). This analysis provides the breakdown underlying the observation in \cref{sec:user}.

\begin{table}[htbp]\centering\small
\begin{tabular}{@{}lcccc@{}}
\toprule
Task & Ref. & Pose/expr. & Quality & Ties \\
\midrule
Head swap  & 93.5 \tiny[84.7,100] & 86.0 \tiny[66.0,100] & 86.2 \tiny[71.9,97.9] & 12.6 \tiny[3.6,23.4] \\
Face swap  & 93.6 \tiny[87.2,98.7] & 39.7 \tiny[23.0,57.0] & 56.1 \tiny[40.0,72.2] & 26.0 \tiny[16.9,35.4] \\
Try-on     & 85.2 \tiny[75.4,93.8] & 71.4 \tiny[42.9,100] & 85.0 \tiny[70.0,100] & 14.6 \tiny[6.8,23.5] \\
Background & 82.2 \tiny[68.8,93.8] & 40.0 \tiny[0.0,80.0] & 52.2 \tiny[30.0,74.4] & 45.7 \tiny[32.7,58.6] \\
\bottomrule
\end{tabular}
\caption{\textbf{Item-level human preferences pooled across approaches (\%).} Ref., Pose/expr., and Quality report preference for \gap after excluding ties. Ties reports the reference-fidelity tie rate. Brackets show $95\%$ item-bootstrap confidence intervals.}
\label{tab:supp_user_pooled}
\end{table}

\section{Attribute trade-offs and failure analysis}
\label{app:limits}

In head swapping, keypoint degradation is concentrated in JoyAI-Video-Edit, FLUX.2-klein, and FLUX.2-klein-base, whereas face-swap keypoint error increases on most approaches. This difference partly reflects the face-swap evaluation protocol, in which the output is compared with the input geometry while being encouraged to adopt the reference identity. We first disentangle this identity-geometry confound and show that the remaining keypoint and expression costs are concentrated in the same three approaches as in head swapping. We then examine whether delaying the attention correction reduces these costs. Finally, we analyze the background-replacement failure of Qwen-Image-Edit and the occasional blurred-face failures of OmniGen2.

\subsection{Identity-attribute trade-offs in face swapping}
\label{sec:face_tradeoff}

FF++ does not provide ground-truth edited frames, so face-swap keypoints are evaluated against the driving input rather than the reference identity. Although the landmarks are scale-normalized, they remain sensitive to identity-specific face shape. Consequently, successful identity transfer can increase keypoint error even when the input pose is preserved.

To quantify this effect, \cref{tab:supp_fs_kpt_id} relates the clip-level change in keypoint error to the corresponding identity gain. For Qwen-Image-Edit, LoomVideo, and OmniGen2, the increase in keypoint error is concentrated among clips with large identity gains: their lowest-gain quartiles have near-zero or negative changes, and only small costs remain when $|\Delta\mathrm{ID}|<0.05$. These patterns are consistent with identity-related geometry changes contributing to the measured keypoint increase.

JoyAI-Video-Edit and FLUX.2-klein-base retain keypoint increases of $0.029$ and $0.052$, respectively, when identity similarity changes by less than $0.05$. Together with the smaller residual increase on FLUX.2-klein, this matches the approach-level pattern observed in head swapping, where a ground-truth target removes the identity confound. After accounting for identity gain, keypoint degradation is likewise concentrated in these three approaches. VACE is listed for completeness but not discussed, since its masked input provides no expression signal.

\begin{table}[H]\centering\scriptsize\setlength{\tabcolsep}{4pt}
\begin{tabular}{l|cc|ccc}
\toprule
Approach & $\Delta$kpt & $\Delta$ID
& lowest $\Delta$ID quartile
& highest $\Delta$ID quartile
& $|\Delta\mathrm{ID}|<0.05$ ($n$) \\
\midrule
JoyAI-Video-Edit  & $+0.048$ & $+0.310$ & $+0.037$ & $+0.064$ & $+0.029$ (44) \\
LoomVideo         & $+0.003$ & $+0.017$ & $-0.010$ & $+0.012$ & $+0.006$ (350) \\
FLUX.2-klein      & $+0.017$ & $+0.116$ & $+0.011$ & $+0.026$ & $+0.012$ (348) \\
FLUX.2-klein-base & $+0.094$ & $+0.274$ & $+0.055$ & $+0.111$ & $+0.052$ (230) \\
VACE               & $+0.006$ & $+0.278$ & $+0.013$ & $+0.002$ & $+0.025$ (13) \\
Qwen-Image-Edit   & $+0.021$ & $+0.169$ & $-0.020$ & $+0.068$ & $+0.012$ (338) \\
OmniGen2          & $+0.011$ & $+0.498$ & $-0.008$ & $+0.022$ & $+0.005$ (16) \\
\bottomrule
\end{tabular}
\caption{\textbf{Face-swap keypoint change versus identity gain.}
Per clip, $\Delta$kpt is \gap minus base keypoint error against the input subject
(lower is better), and $\Delta$ID is \gap minus base AdaFace similarity to the
reference image given to the model. The quartile columns report the average $\Delta$kpt within the lowest and highest quarters of $\Delta$ID. The last
column reports the average over clips whose identity similarity changes by less
than $0.05$, with the corresponding count in parentheses. }
\label{tab:supp_fs_kpt_id}
\end{table}

\subsection{Denoising-step interventions}
\label{sec:supp_knob}

\paragraph{Source of the keypoint cost.} \gap raises attention to all reference tokens with a single offset, so it may also transfer the reference's expression and head pose. We test this on the $1040$ head-swap test clips. With MediaPipe, we extract a $51$-dimensional blendshape vector (expression) and Euler angles (head pose, in degrees) from the output, the reference image, and the target frame.

\emph{Pull toward the reference} is how much closer the \gap output is to the reference than the base output is, minus the same quantity computed with an unrelated reference image in place of the true one; negative values mean the output moves toward this particular reference. \emph{Drift from the target} is the corresponding increase in distance to the target frame. We also report the Spearman correlation between the reference--target gap and the per-clip keypoint cost. Intervals are $95\%$ bootstrap intervals clustered by source video.

\Cref{tab:supp_attr} shows that on the three approaches whose keypoint error rises, JoyAI-Video-Edit, FLUX.2-klein, and FLUX.2-klein-base, the output moves toward the reference in both expression ($-.063$ to $-.260$) and head pose ($-0.24^\circ$ to $-3.66^\circ$), with intervals excluding zero, and drifts away from the target frame. Clips with a larger reference--target gap have a larger keypoint cost ($\rho=+.18$ to $+.44$). On Qwen-Image-Edit, whose keypoint error does not change, expression moves slightly toward the reference ($-.069$), but head pose does not change significantly and the correlation is near zero. The cost therefore appears where the pull toward the reference is large.

\begin{table}[H]\centering\scriptsize\setlength{\tabcolsep}{4pt}
\begin{tabular}{lcccc}
\toprule
& \multicolumn{2}{c}{Pull toward reference} & & \\
Approach & Expression & Head pose ($^\circ$) & Drift from target (expr.\ / pose) & $\rho$ (expr.\ / pose) \\
\midrule
JoyAI-Video-Edit  & $-.063$ [$-.079$, $-.047$] & $-0.24$ [$-0.44$, $-0.06$] & $+.095$ / $+1.0^\circ$ & $+.19$ / $+.22$ \\
FLUX.2-klein      & $-.159$ [$-.188$, $-.134$] & $-1.81$ [$-2.13$, $-1.55$] & $+.062$ / $+1.9^\circ$ & $+.29$ / $+.44$ \\
FLUX.2-klein-base & $-.260$ [$-.288$, $-.230$] & $-3.66$ [$-4.36$, $-3.08$] & $+.058$ / $+2.4^\circ$ & $+.18$ / $+.25$ \\
\midrule
Qwen-Image-Edit   & $-.069$ [$-.111$, $-.028$] & $-0.24$ (n.s.) & $+.021$ / $+0.2^\circ$ (n.s.) & $+.00$ / $-.03$ (n.s.) \\
\bottomrule
\end{tabular}
\caption{\textbf{Pull of head-swap outputs toward the reference's expression and pose.} $1040$ head-swap test clips. Pull toward reference: change in the output's distance to the reference (\gap minus base), minus the same change for an unrelated reference; negative means closer. Drift from target: change in distance to the target frame. $\rho$: Spearman correlation between the reference--target gap and the per-clip keypoint cost. Qwen-Image-Edit, whose keypoint error does not change, is shown for comparison. Brackets: $95\%$ intervals clustered by source video; n.s.: interval includes zero.}
\label{tab:supp_attr}
\end{table}

We next test whether the geometric cost can be reduced by delaying the attention correction. The full denoising trajectory is preserved; the correction is disabled only during selected initial steps.

As shown in \cref{tab:supp_knob}, skipping the first step reduces keypoint error from $0.085$ to $0.052$ on JoyAI-Video-Edit and from $0.153$ to $0.101$ on FLUX.2-klein. These improvements are accompanied by reductions in identity similarity from $0.644$ to $0.613$ and from $0.790$ to $0.753$, respectively.

FLUX.2-klein-base uses a longer, $50$-step schedule, for which skipping only the first step has little effect. Skipping the first ten steps (on $200$ randomly sampled test clips) reduces keypoint error from $0.236$ to $0.183$, but also lowers identity similarity from $0.799$ to $0.735$.

To compare identity at near-base keypoint error, we select on the validation split, for each of these three approaches, the highest-identity setting among those swept whose keypoint error is at most $.005$ above the base (the preservation tolerance): skip step 1 with $\gamma_{\mathrm{edit}}=0.4$ on FLUX.2-klein and $0.6$ on JoyAI-Video-Edit, and the bias applied only for $\sigma\le0.9$ on FLUX.2-klein-base. On the test set, these settings keep identity gains of $+.032$, $+.106$, and $+.107$ over the base at keypoint error $.007$, $.009$, and $.001$ above it (\cref{tab:supp_knob}).

Thus, delaying the correction can partially preserve input geometry, but at the cost of weaker identity transfer. Because this trade-off depends on the approach and sampling schedule, we retain correction at all steps to prioritize recognizable reference identity.

\begin{table}
\centering
\setlength{\tabcolsep}{4pt}
\begin{tabular}{lccc}
\toprule
HeadSwapBench & ID$\uparrow$ & kpt$\downarrow$ & LPIPS$\downarrow$ \\
\midrule
\multicolumn{4}{@{}l}{\emph{FLUX.2-klein (4-step schedule)}} \\
base & 0.726 & \textbf{0.097} & 0.222 \\
\gap ($\gamma_{\mathrm{edit}}{=}0.6$, all steps) & \textbf{0.790} & 0.153 & \textbf{0.217} \\
\gap, skip step 1 & 0.753 & 0.101 & 0.221 \\
\gap ($\gamma_{\mathrm{edit}}{=}0.4$), skip step 1 & 0.758 & 0.104 & --- \\
\midrule
\multicolumn{4}{@{}l}{\emph{JoyAI-Video-Edit (2-step schedule)}} \\
base & 0.507 & \textbf{0.043} & 0.218 \\
\gap ($\gamma_{\mathrm{edit}}{=}0.6$, all steps) & \textbf{0.644} & 0.085 & \textbf{0.215} \\
\gap, skip step 1 & 0.613 & 0.052 & 0.217 \\
\midrule
\multicolumn{4}{@{}l}{\emph{FLUX.2-klein-base (50-step schedule)}} \\
base & 0.588 & \textbf{0.169} & 0.326 \\
\gap ($\gamma_{\mathrm{edit}}{=}0.6$, all steps) & \textbf{0.799} & 0.236 & \textbf{0.261} \\
\gap, skip step 1$^\dagger$ & 0.798 & 0.235 & 0.264 \\
\gap, skip first $10/50$ steps$^\dagger$ & 0.735 & 0.183 & 0.320 \\
\gap, bias only for $\sigma\le0.9$ & 0.695 & 0.170 & --- \\
\bottomrule
\end{tabular}
\caption{\textbf{Disabling correction at early denoising steps.} Results on $1040$ head-swap test clips. ``Skip'' disables the attention correction at the indicated steps; denoising still proceeds normally. LPIPS is measured over the whole frame; --- marks runs for which it was not computed. $^\dagger$Measured on $200$ randomly sampled test clips. Bold marks the best value within each approach.}
\label{tab:supp_knob}
\end{table}

\begin{figure*}[t]\centering
\includegraphics[width=\textwidth]{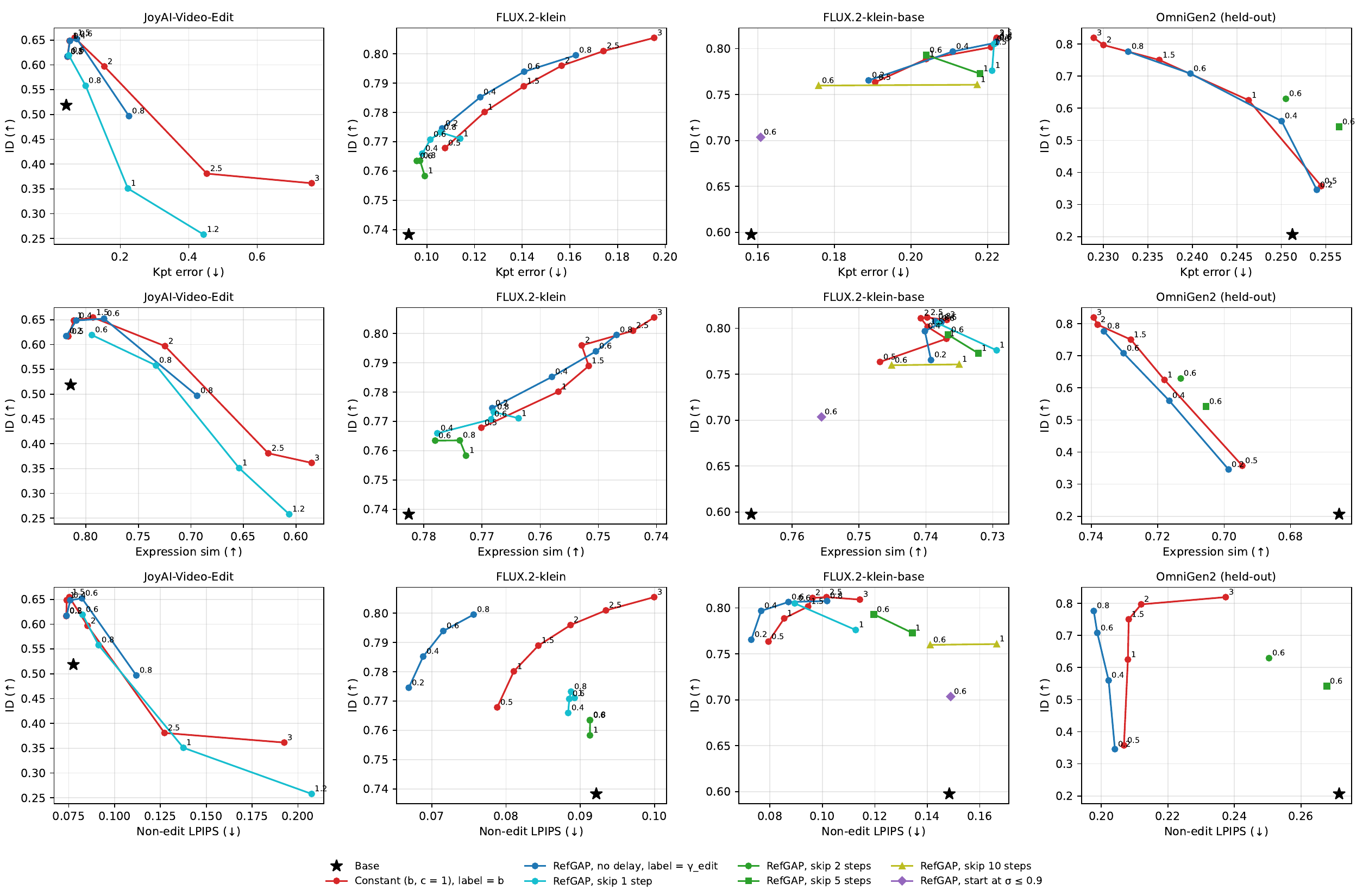}
\caption{\textbf{Identity--pose trade-off of \gap and the shared constant on head swap.} Rows show identity versus keypoint error, expression similarity, and non-edit LPIPS; the expression axis is reversed so that the upper left is better in every row. \gap labels give $\gamma_{\mathrm{edit}}$; colors give the delay setting. Constant labels give $b$, with $c=1$. The star marks the base. $n=200$ clips ($197$ on FLUX.2-klein-base and $198$ on OmniGen2).}
\label{fig:supp_pareto}
\end{figure*}

\paragraph{Identity--pose frontier.} \Cref{fig:supp_pareto} places \gap and the shared constant of \cref{app:shared_const} on the same identity--keypoint and identity--expression planes, on $200$ randomly sampled head-swap test clips. We sweep four approaches: JoyAI-Video-Edit, FLUX.2-klein, and FLUX.2-klein-base, where \gap raises head-swap keypoint error the most (\cref{tab:main}), and the held-out OmniGen2. On the other three, \gap changes head-swap keypoint error by at most $.009$ (LoomVideo $-.009$, VACE $+.002$, Qwen-Image-Edit $.000$), so there is little pose cost to trade, and their slow generation makes a full sweep costly. For \gap we vary $\gamma_{\mathrm{edit}}\in\{0.2,0.4,0.6,0.8\}$ and the number of skipped steps; for the constant we vary $b\in\{0.5,\dots,3\}$ with $c=1$.

The two curves largely overlap. At matched keypoint error, interpolated along the \gap frontier, identity differs from the constant's by $-.011$ to $+.008$ on JoyAI-Video-Edit, FLUX.2-klein, and FLUX.2-klein-base. On OmniGen2, stronger settings lower keypoint error, so there is no pose cost to trade. At matched identity, \gap's LPIPS$_{\mathrm{ne}}$ is lower by $.010$ to $.016$ on the two FLUX.2 variants and by $.007$ to $.010$ on OmniGen2 (intervals excluding zero), and the two are equal on JoyAI-Video-Edit.

Lowering $\gamma_{\mathrm{edit}}$ is a better way to trade identity for pose than delaying the correction: on JoyAI-Video-Edit, $\gamma_{\mathrm{edit}}=0.4$ gives keypoint error $.053$ at identity $.648$, whereas skipping one step at $\gamma_{\mathrm{edit}}=0.6$ gives $.050$ at $.619$. This agrees with \cref{tab:abl_fixed}: \gap's identity--pose trade-off is comparable to that of a tuned constant, without a per-approach strength sweep; against constants with $c=1$, it also preserves the non-edit region better.

\subsection{Background replacement failure}

Background replacement differs from head swapping, face swapping, and virtual try-on: the reference specifies the surrounding scene rather than the person or their appearance, while the person must be preserved. This shift in the role of the reference may affect how an approach responds to stronger reference conditioning.

On Qwen-Image-Edit, the shared setting reduces DINO from 0.299 to 0.042. Restricting correction to the second half does not recover fidelity on the diagnostic clips. A fixed edit-side bias of $b=2.0$ also produces a similar degradation, indicating that the failure is not specific to adaptive edit-side strength (see \cref{tab:supp_qwen_bg}).

\begin{table}[H]\centering\scriptsize\setlength{\tabcolsep}{4pt}
{%
\begin{tabular}{llc|cc|c}
\toprule
 & & & \multicolumn{2}{c|}{diagnostic clips ($n=30$)} & all clips ($n=150$) \\
setting & layers & edit-side strength & DINO $\uparrow$ & subject LPIPS $\downarrow$ & DINO $\uparrow$ \\
\midrule
base & --- & --- & .473 & .0384 & .299 \\
\gap (shared setting) & all & online, $\gamma_{\mathrm{edit}}=0.6$ & .112 & .0242 & .042 \\
\gap, second half & $30$--$59$ & online, $\gamma_{\mathrm{edit}}=0.6$ & .106 & .0127 & --- \\
fixed edit-side bias & all & constant, $b=2.0$ & .040 & .0174 & .037 \\
\bottomrule
\end{tabular}}
\caption{\textbf{Qwen-Image-Edit background replacement under alternative settings.} The diagnostic clips are the first $30$ of the $150$-clip list, not selected on outcome. }
\label{tab:supp_qwen_bg}
\end{table}

Other approaches improve on this task, so the task difference alone does not explain the failure. The results suggest an approach-dependent limitation when transferring the correction from person-related edits to scene replacement; its underlying cause remains unresolved.

\subsection{Rare head-swap collapses}

\begin{figure*}[htbp]\centering
\includegraphics[width=\textwidth]{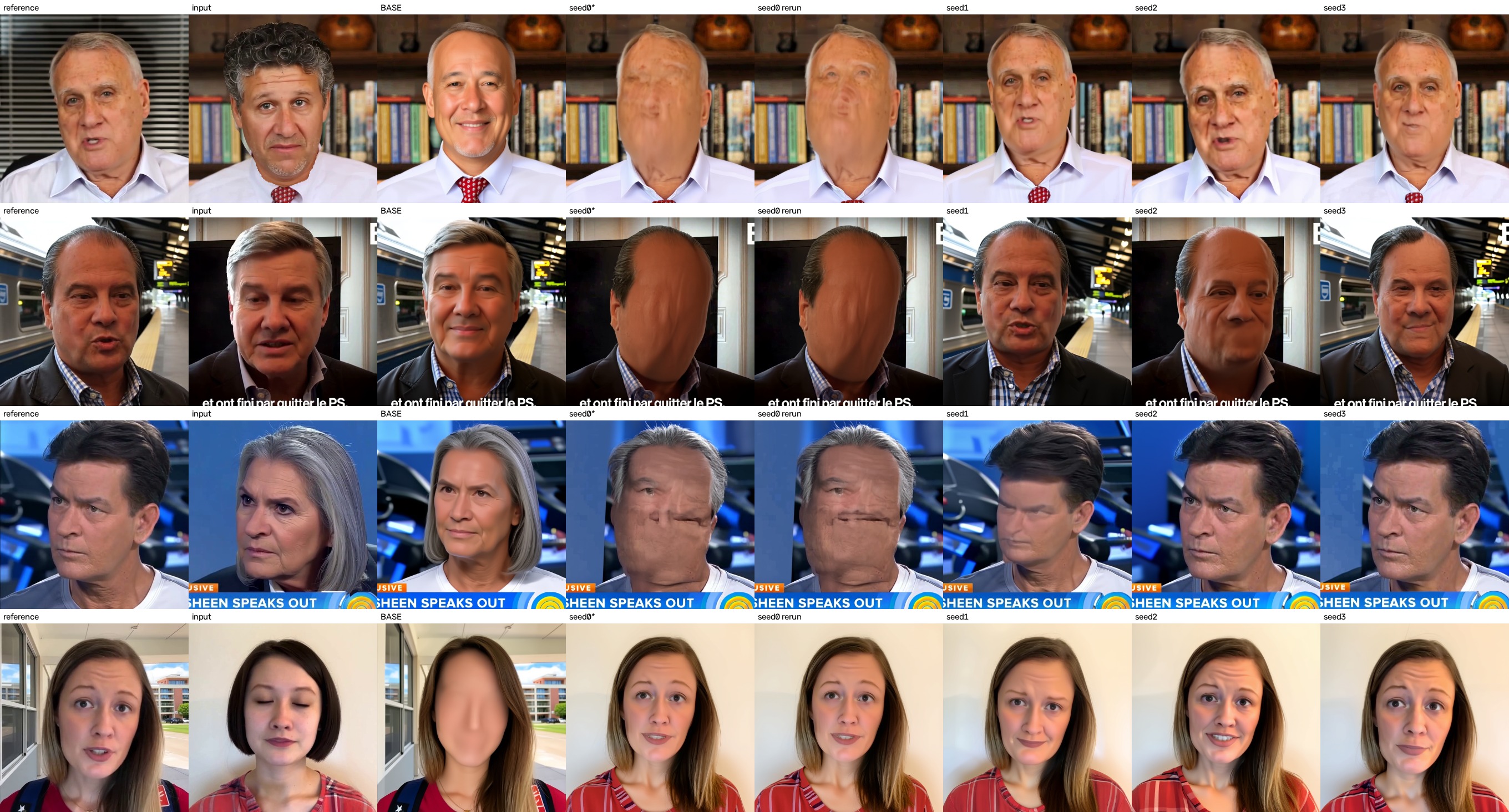}
\caption{\textbf{Seed sensitivity of OmniGen2 head-swap outputs.} Columns show the reference, input, base, \gap at seed $0$, a repeat at the same seed, and three alternative seeds. Repeated seeds reproduce the failures, while alternative seeds can recover plausible faces. }
\label{fig:supp_seedgrid}
\end{figure*}

OmniGen2 occasionally produces missing or heavily smeared faces in both the base and \gap settings. Screening followed by manual inspection identifies $3/1040$ such outputs for the base and $8/1040$ for \gap, indicating an existing failure mode that becomes more frequent under the correction. As shown in \cref{fig:supp_seedgrid}, changing the sampling seed can recover plausible faces, while repeating the same seed reproduces the failure.

\section{Additional experiments}
\label{app:extra}

This section compares \gap with a shared constant edit/keep pair, evaluates it on object replacement and multi-person head swapping, compares it with SaaS and KV-Edit, and relates the measured reference mass to the required strength. \gap uses the deployed coefficients and layer rule without re-tuning throughout.

\subsection{Shared constant offsets}
\label{app:shared_const}

\paragraph{Setup.}
\Cref{tab:abl_fixed} replaces only the edit-side strength with a constant chosen per approach. Here both sides are constant and shared across approaches: the edit-region offset is $+b$ and the keep-region offset is $-c$, applied with the same masks and the same layer rule as \gap.

We select the pair with the rule used for $(\gamma_{\mathrm{edit}},\gamma_{\mathrm{keep}})$ (\cref{app:moved}), applied to a $5\times5$ grid, $b\in\{1,1.5,2,2.5,3\}$ and $c\in\{0,0.5,1,2,3\}$, on the 40-clip head-swap validation split of the four development approaches at their rule-selected layers (\cref{tab:supp_shared_grid}; per-approach values in \cref{tab:supp_shared_grid_full}). A pair is feasible if it raises identity and keeps $\Delta\mathrm{LPIPS}_{\mathrm{ne}}\leq0.005$ on all four approaches. This excludes every $b\geq2$, where JoyAI-Video-Edit exceeds the criterion, and $c=0$, where both FLUX.2 variants exceed it.

Among feasible pairs, $b=1.5$ gives higher identity than $b=1$ on every approach at every $c$. At $b=1.5$, raising $c$ from $0.5$ to $1$ lowers $\Delta\mathrm{LPIPS}_{\mathrm{ne}}$ on all four approaches while changing identity by at most $.001$; raising it further to $2$ lowers JoyAI-Video-Edit's identity by $.002$ and does not lower its $\Delta\mathrm{LPIPS}_{\mathrm{ne}}$. We therefore select $(b,c)=(1.5,1)$. Held-out approaches play no part in the selection.

\begin{table}[H]\centering\scriptsize\setlength{\tabcolsep}{4pt}
\begin{tabular}{c|ccccc}
\toprule
$b\,\backslash\,c$ & $0$ & $0.5$ & $1$ & $2$ & $3$ \\
\midrule
$1$ & .778 \ding{55}\,(K, B) & .784 & .781 & .781 & .780 \\
$1.5$ & .798 \ding{55}\,(K, B) & .800 & \textbf{.799} & .798 & .798 \\
$2$ & .786 \ding{55}\,(J, K, B) & .786 \ding{55}\,(J, K) & .785 \ding{55}\,(J) & .787 \ding{55}\,(J) & .784 \ding{55}\,(J) \\
$2.5$ & .736 \ding{55}\,(J, K, B) & .735 \ding{55}\,(J, K) & .736 \ding{55}\,(J, K) & .733 \ding{55}\,(J) & .733 \ding{55}\,(J) \\
$3$ & .738 \ding{55}\,(J, L, K, B) & .738 \ding{55}\,(J, L, K) & .739 \ding{55}\,(J, L, K) & .742 \ding{55}\,(J, L, K) & .741 \ding{55}\,(J, L, K) \\
\bottomrule
\end{tabular}
\caption{\textbf{Selecting the shared constant pair on validation.} Mean head-swap validation identity over the four development approaches ($40$ clips) for edit offset $+b$ and keep offset $-c$ at the rule-selected layers. \ding{55}~marks pairs that violate $\Delta\mathrm{LPIPS}_{\mathrm{ne}}\leq0.005$ on the listed approaches (J: JoyAI-Video-Edit, L: LoomVideo, K: FLUX.2-klein, B: FLUX.2-klein-base). Bold: the selected pair.}
\label{tab:supp_shared_grid}
\end{table}

\begin{table}[H]\centering\scriptsize\setlength{\tabcolsep}{4pt}
\begin{tabular}{cc|cccc}
\toprule
$b$ & $c$ & JoyAI-Video-Edit & LoomVideo & FLUX.2-klein & FLUX.2-klein-base \\
\midrule
$1$ & $0$ & .685 / $-.0046$ & .699 / $-.0001$ & .883 / $+.0068$ \ding{55} & .844 / $+.0466$ \ding{55} \\
$1$ & $0.5$ & .687 / $-.0045$ & .698 / $-.0004$ & .883 / $-.0002$ & .866 / $-.0392$ \\
$1$ & $1$ & .684 / $-.0040$ & .697 / $-.0004$ & .883 / $-.0030$ & .859 / $-.0524$ \\
$1$ & $2$ & .685 / $-.0042$ & .694 / $-.0013$ & .883 / $-.0042$ & .861 / $-.0540$ \\
$1$ & $3$ & .685 / $-.0040$ & .692 / $-.0023$ & .883 / $-.0045$ & .860 / $-.0546$ \\
\midrule
$1.5$ & $0$ & .703 / $-.0042$ & .724 / $-.0003$ & .887 / $+.0097$ \ding{55} & .877 / $+.0718$ \ding{55} \\
$1.5$ & $0.5$ & .707 / $-.0047$ & .724 / $-.0004$ & .887 / $+.0020$ & .880 / $-.0327$ \\
$1.5$ & $1$ & .706 / $-.0049$ & .724 / $-.0005$ & .888 / $-.0010$ & .879 / $-.0440$ \\
$1.5$ & $2$ & .704 / $-.0048$ & .723 / $-.0013$ & .887 / $-.0025$ & .879 / $-.0532$ \\
$1.5$ & $3$ & .704 / $-.0040$ & .721 / $-.0027$ & .888 / $-.0029$ & .878 / $-.0552$ \\
\midrule
$2$ & $0$ & .626 / $+.0102$ \ding{55} & .746 / $+.0002$ & .889 / $+.0132$ \ding{55} & .882 / $+.0819$ \ding{55} \\
$2$ & $0.5$ & .625 / $+.0101$ \ding{55} & .746 / $-.0002$ & .889 / $+.0050$ \ding{55} & .882 / $-.0116$ \\
$2$ & $1$ & .626 / $+.0109$ \ding{55} & .746 / $-.0006$ & .890 / $+.0014$ & .881 / $-.0448$ \\
$2$ & $2$ & .633 / $+.0097$ \ding{55} & .745 / $-.0012$ & .890 / $-.0003$ & .881 / $-.0501$ \\
$2$ & $3$ & .623 / $+.0098$ \ding{55} & .743 / $-.0027$ & .890 / $-.0007$ & .881 / $-.0509$ \\
\midrule
$2.5$ & $0$ & .416 / $+.0491$ \ding{55} & .761 / $+.0045$ & .889 / $+.0207$ \ding{55} & .878 / $+.0868$ \ding{55} \\
$2.5$ & $0.5$ & .416 / $+.0502$ \ding{55} & .760 / $+.0039$ & .889 / $+.0119$ \ding{55} & .876 / $-.0031$ \\
$2.5$ & $1$ & .417 / $+.0500$ \ding{55} & .760 / $+.0024$ & .889 / $+.0074$ \ding{55} & .876 / $-.0352$ \\
$2.5$ & $2$ & .412 / $+.0500$ \ding{55} & .761 / $+.0016$ & .888 / $+.0043$ & .873 / $-.0431$ \\
$2.5$ & $3$ & .413 / $+.0515$ \ding{55} & .760 / $-.0000$ & .888 / $+.0036$ & .873 / $-.0443$ \\
\midrule
$3$ & $0$ & .417 / $+.1144$ \ding{55} & .776 / $+.0150$ \ding{55} & .889 / $+.0269$ \ding{55} & .870 / $+.0837$ \ding{55} \\
$3$ & $0.5$ & .417 / $+.1148$ \ding{55} & .776 / $+.0141$ \ding{55} & .888 / $+.0180$ \ding{55} & .871 / $-.0078$ \\
$3$ & $1$ & .421 / $+.1152$ \ding{55} & .775 / $+.0128$ \ding{55} & .889 / $+.0146$ \ding{55} & .872 / $-.0260$ \\
$3$ & $2$ & .434 / $+.1147$ \ding{55} & .774 / $+.0108$ \ding{55} & .888 / $+.0115$ \ding{55} & .872 / $-.0300$ \\
$3$ & $3$ & .427 / $+.1135$ \ding{55} & .774 / $+.0094$ \ding{55} & .888 / $+.0099$ \ding{55} & .874 / $-.0313$ \\
\bottomrule
\end{tabular}
\caption{\textbf{Per-approach validation results for the shared constant grid.} Cells show identity / $\Delta\mathrm{LPIPS}_{\mathrm{ne}}$ on the $40$-clip head-swap validation split at the rule-selected layers. \ding{55}~marks $\Delta\mathrm{LPIPS}_{\mathrm{ne}}>0.005$.}
\label{tab:supp_shared_grid_full}
\end{table}

We evaluate the selected pair on the $1{,}040$ head-swap test clips of all seven approaches (\cref{tab:supp_shared_const}). LPIPS$_{\mathrm{ne}}$ is LPIPS to the ground truth outside the dilated head mask; a higher value means the non-edit region is damaged more. Intervals are clustered by source video.

\begin{table}[H]\centering\scriptsize\setlength{\tabcolsep}{3pt}
\begin{tabular}{l|ccc|ccc|cc}
\toprule
& \multicolumn{3}{c|}{\gap} & \multicolumn{3}{c|}{Shared constant $(1.5,1)$} & \multicolumn{2}{c}{Constant $-$ \gap\ [95\% CI]} \\
Approach & ID $\uparrow$ & LPIPS$_{\mathrm{ne}}$ $\downarrow$ & Kpt.\ $\downarrow$ & ID $\uparrow$ & LPIPS$_{\mathrm{ne}}$ $\downarrow$ & Kpt.\ $\downarrow$ & $\Delta$ID & $\Delta$LPIPS$_{\mathrm{ne}}$ \\
\midrule
JoyAI-Video-Edit$^\ast$ & .644 & .079 & .085 & .644 & \textbf{.072} & .074 & $+.001$ [$-.002$, $+.004$] & $-.007$ [$-.008$, $-.006$] \\
LoomVideo & \textbf{.665} & .136 & .098 & .647 & .133 & .100 & $-.018$ [$-.023$, $-.013$] & $-.003$ [$-.006$, $-.001$] \\
FLUX.2-klein & .790 & \textbf{.071} & .153 & .788 & .083 & .152 & $-.002$ [$-.005$, $-.000$] & $+.012$ [$+.007$, $+.017$] \\
FLUX.2-klein-base & .799 & \textbf{.087} & .236 & .795 & .093 & .234 & $-.004$ [$-.007$, $-.001$] & $+.007$ [$+.001$, $+.014$] \\
\midrule
VACE$^\dagger$ & .729 & .047 & .200 & \textbf{.736} & .047 & .195 & $+.008$ [$+.007$, $+.009$] & $+.001$ [$+.001$, $+.001$] \\
Qwen-Image-Edit$^\dagger$ & .723 & \textbf{.075} & .179 & \textbf{.738} & .129 & .194 & $+.015$ [$+.008$, $+.022$] & $+.054$ [$+.043$, $+.066$] \\
OmniGen2$^\dagger$ & .693 & \textbf{.192} & .258 & \textbf{.745} & .202 & .249 & $+.052$ [$+.039$, $+.066$] & $+.010$ [$+.006$, $+.014$] \\
\bottomrule
\end{tabular}
\caption{\textbf{\gap versus a shared constant edit/keep pair on head swapping ($1{,}040$ test clips).} $^\dagger$~held-out approaches. $^\ast$~Both arms run at $504^2$ on $1{,}039$ clips. Bold marks the better ID or LPIPS$_{\mathrm{ne}}$ where the two differ by at least $.005$. Base LPIPS$_{\mathrm{ne}}$: JoyAI-Video-Edit .075, LoomVideo .133, FLUX.2-klein .090, FLUX.2-klein-base .158, VACE .048, Qwen-Image-Edit .289, OmniGen2 .262.}
\label{tab:supp_shared_const}
\end{table}

\paragraph{Results.}
Identity is within $.004$ of \gap on JoyAI-Video-Edit and the two FLUX.2 variants. On LoomVideo the constant loses identity ($-.018$): the shared $b=1.5$ is capped because JoyAI-Video-Edit rules out every $b\geq2$ on validation, whereas \gap sets the strength online. On the three held-out approaches the constant has higher identity ($+.008$ to $+.052$). On Qwen-Image-Edit and OmniGen2 this gain comes with higher LPIPS$_{\mathrm{ne}}$ ($+.054$ and $+.010$); the gap on Qwen-Image-Edit is large ($.129$ versus $.075$). LPIPS$_{\mathrm{ne}}$ is also higher on both FLUX.2 variants, slightly lower on JoyAI-Video-Edit and LoomVideo ($-.007$ and $-.003$), and nearly equal on VACE.

\paragraph{Near-copies of the reference.}
LPIPS$_{\mathrm{ne}}$ measures how far the non-edit region departs from the ground truth, but not whether the output has been replaced by the reference. We count an output as a near-copy when its whole-frame LPIPS (AlexNet) to the reference is below $0.2$, with both images resized to $512^2$. The test uses only the output and the reference. Image approaches use the edited middle frame; video approaches use the middle output frame. A near-copy reaches high identity without editing the input, so we also report identity with near-copies scored $0$, as for outputs without a detected face (\cref{app:multiperson}). Intervals use a paired bootstrap clustered by source video ($B=5000$).

\begin{table}[H]
\centering
\scriptsize
\setlength{\tabcolsep}{0.5pt}
\resizebox{\textwidth}{!}{%
\begin{tabular}{l|ccc|cc}
\toprule
& \multicolumn{3}{c|}{Near-copy rate (\%) $\downarrow$} & \multicolumn{2}{c}{ID, near-copies scored $0$ $\uparrow$} \\
Approach & Base / \gap / const. & \gap $-$ base [95\% CI] & Const.\ $-$ \gap\ [95\% CI] & Base / \gap / const. & Const.\ $-$ \gap\ [95\% CI] \\
\midrule
JoyAI-Video-Edit$^\ast$ & 1.7 / 1.5 / 1.7 & $-0.2$ [$-0.7$, $+0.2$] & $+0.2$ [$+0.0$, $+0.5$] & .495 / .631 / .631 & $-.001$ [$-.004$, $+.003$] \\
LoomVideo & 5.6 / 7.5 / 6.2 & $+1.9$ [$+0.9$, $+3.2$] & $-1.2$ [$-2.3$, $-0.3$] & .566 / .609 / .598 & $-.010$ [$-.019$, $-.002$] \\
FLUX.2-klein & 7.9 / 7.0 / 9.4 & $-0.9$ [$-2.8$, $+1.3$] & $+2.4$ [$+1.0$, $+4.0$] & .673 / .730 / .708 & $-.022$ [$-.038$, $-.009$] \\
FLUX.2-klein-base & 10.9 / 6.4 / 7.0 & $-4.4$ [$-7.3$, $-1.1$] & $+0.6$ [$+0.0$, $+1.2$] & .534 / .744 / .736 & $-.008$ [$-.014$, $-.003$] \\
\midrule
VACE$^\dagger$ & 2.1 / 2.3 / 2.3 & $+0.2$ [$+0.0$, $+0.5$] & $+0.0$ [$+0.0$, $+0.0$] & .390 / .709 / .716 & $+.007$ [$+.006$, $+.009$] \\
Qwen-Image-Edit$^\dagger$ & 57.7 / 9.1 / 22.3 & $-48.6$ [$-55.0$, $-41.4$] & $+13.2$ [$+10.6$, $+16.0$] & .153 / .645 / .560 & $-.084$ [$-.108$, $-.062$] \\
OmniGen2$^\dagger$ & 5.6 / 5.3 / 10.3 & $-0.3$ [$-2.4$, $+2.1$] & $+5.0$ [$+3.6$, $+6.6$] & .188 / .650 / .663 & $+.013$ [$-.004$, $+.029$] \\
\bottomrule
\end{tabular}%
}
\caption{\textbf{Near-copies of the reference on head swapping ($1{,}040$ test clips).} An output is a near-copy when its whole-frame LPIPS to the reference is below $0.2$. Rate differences are in percentage points. $^\dagger$~held-out approaches. $^\ast$~All arms run at $504^2$ on $1{,}039$ clips.}
\label{tab:supp_leak}
\end{table}

\gap lowers the near-copy rate on Qwen-Image-Edit by $48.6$ points ($57.7\%\to9.1\%$) and on FLUX.2-klein-base by $4.4$ points. It leaves the rate unchanged on FLUX.2-klein, OmniGen2, JoyAI-Video-Edit, and VACE (intervals include zero; on VACE, $24$ versus $22$ clips), and raises it by $1.9$ points on LoomVideo. On VACE, \gap and the constant flag the same $24$ clips. With near-copies scored $0$, \gap's identity remains above the base on all seven approaches.

The constant produces more near-copies than \gap on all four image approaches ($+0.6$ to $+13.2$ points; borderline on FLUX.2-klein-base) and $1.2$ points fewer on LoomVideo. With near-copies scored $0$, \gap has higher identity than the constant on FLUX.2-klein, FLUX.2-klein-base, Qwen-Image-Edit, and LoomVideo ($.008$ to $.084$). On OmniGen2, the constant's raw advantage of $.052$ shrinks to $+.013$ [$-.004$, $+.029$]; on VACE the constant remains $.007$ higher, and JoyAI-Video-Edit is tied. Keypoint error and expression similarity are slightly better for the constant on JoyAI-Video-Edit, VACE, and OmniGen2 (by at most $.010$) and better for \gap on Qwen-Image-Edit.

\paragraph{Threshold sensitivity.}
\Cref{tab:supp_leak_tau} repeats the analysis with thresholds $0.15$ and $0.25$ on the same outputs. The direction of every near-copy-rate difference is unchanged: the constant makes more near-copies on the four image approaches (lower bound $+0.0$ on FLUX.2-klein-base at all thresholds) and fewer on LoomVideo. With near-copies scored $0$, \gap's identity is higher than the constant's on FLUX.2-klein, FLUX.2-klein-base, and Qwen-Image-Edit at all thresholds; on LoomVideo the difference is not significant at $0.25$. On OmniGen2 the identity difference depends on the threshold: the constant remains higher at $0.15$ ($+.028$ [$+.014$, $+.042$]), the difference is not significant at $0.2$, and it is $.000$ at $0.25$. Part of the constant's identity advantage on OmniGen2 therefore comes from near-copies, but not all of it at the strictest threshold.

\begin{table}[H]\centering\scriptsize\setlength{\tabcolsep}{4pt}
\begin{tabular}{l|ccc}
\toprule
Approach & $\tau=0.15$ & $\tau=0.2$ & $\tau=0.25$ \\
\midrule
\multicolumn{4}{l}{\textit{Near-copy rate, constant $-$ \gap\ (points)}} \\
JoyAI-Video-Edit$^\ast$ & $+0.1$ & $+0.2$ & $0.0$ \\
LoomVideo & $-0.9$ [$-1.9$, $+0.0$] & $-1.2$ [$-2.3$, $-0.3$] & $-1.6$ [$-2.8$, $-0.5$] \\
FLUX.2-klein & $+1.8$ [$+0.8$, $+2.9$] & $+2.4$ [$+1.0$, $+4.0$] & $+2.1$ [$+1.0$, $+3.3$] \\
FLUX.2-klein-base & $+0.4$ [$+0.0$, $+0.9$] & $+0.6$ [$+0.0$, $+1.2$] & $+1.2$ [$+0.0$, $+2.9$] \\
\midrule
VACE$^\dagger$ & $0.0$ & $0.0$ & $+0.1$ \\
Qwen-Image-Edit$^\dagger$ & $+11.8$ [$+9.5$, $+14.5$] & $+13.2$ [$+10.6$, $+16.0$] & $+15.7$ [$+12.7$, $+18.9$] \\
OmniGen2$^\dagger$ & $+3.0$ [$+1.9$, $+4.2$] & $+5.0$ [$+3.6$, $+6.6$] & $+6.7$ [$+5.1$, $+8.7$] \\
\midrule
\multicolumn{4}{l}{\textit{ID with near-copies scored $0$, constant $-$ \gap}} \\
JoyAI-Video-Edit$^\ast$ & $.000$ & $-.001$ & $+.001$ \\
LoomVideo & $-.012$ [$-.020$, $-.004$] & $-.010$ [$-.019$, $-.002$] & $-.007$ [$-.017$, $+.002$] \\
FLUX.2-klein & $-.018$ [$-.029$, $-.008$] & $-.022$ [$-.038$, $-.009$] & $-.019$ [$-.030$, $-.009$] \\
FLUX.2-klein-base & $-.007$ [$-.012$, $-.003$] & $-.008$ [$-.014$, $-.003$] & $-.014$ [$-.029$, $-.004$] \\
\midrule
VACE$^\dagger$ & $+.007$ & $+.007$ & $+.007$ \\
Qwen-Image-Edit$^\dagger$ & $-.077$ [$-.099$, $-.056$] & $-.084$ [$-.108$, $-.062$] & $-.103$ [$-.128$, $-.079$] \\
OmniGen2$^\dagger$ & $+.028$ [$+.014$, $+.042$] & $+.013$ [$-.004$, $+.029$] & $.000$ [$-.019$, $+.017$] \\
\bottomrule
\end{tabular}
\caption{\textbf{Near-copy threshold sensitivity on head swapping ($1{,}040$ test clips).} $\tau$ is the LPIPS threshold to the reference; $\tau=0.2$ is used in \cref{tab:supp_leak}. Intervals are omitted for JoyAI-Video-Edit and VACE, where all differences are at most $0.2$ points and $.007$. $^\dagger$~held-out approaches. $^\ast$~$1{,}039$ clips, as in \cref{tab:supp_leak}.}
\label{tab:supp_leak_tau}
\end{table}

\paragraph{Conclusion.}
The shared pair selected by the same rule, $(1.5,1)$, matches \gap's identity on most approaches, but it is too weak on LoomVideo. On the image approaches, it produces more near-copies of the reference; once these are scored as failures, its identity advantage on OmniGen2 shrinks to between $+.028$ and $.000$ depending on the threshold, and it reverses on Qwen-Image-Edit, where its LPIPS$_{\mathrm{ne}}$ is also much higher. On the video approaches, \gap does not reduce near-copies and adds $1.9$ points on LoomVideo. \gap reaches comparable identity with coefficients fixed once.

\subsection{Object replacement}
\label{app:object}

\paragraph{Image approaches.}
We run the subject-replacement task of DreamEditBench~\citep{li2023dreamedit} on the four image approaches ($212$ pairs, SAM3~\citep{carion2025sam3segmentconcepts} masks). Background LPIPS compares the output with the source outside the dilated source-object mask; a higher value means the background is damaged more. Subject DINO compares the output with the reference subject. Both use all pairs, and intervals are clustered by reference subject ($22$ subjects).

\begin{table}[H]\centering\scriptsize\setlength{\tabcolsep}{4pt}
\begin{tabular}{l|cc|cc}
\toprule
Approach & Bg.\ LPIPS $\downarrow$ & $\Delta$ [95\% CI] & Subject DINO $\uparrow$ & $\Delta$ [95\% CI] \\
\midrule
FLUX.2-klein & .449$\to$.397 & $-.052$ [$-.069$, $-.035$] & .603$\to$.626 & $+.023$ [$+.006$, $+.042$] \\
FLUX.2-klein-base & .560$\to$.320 & $-.239$ [$-.310$, $-.169$] & .596$\to$.639 & $+.043$ [$+.011$, $+.076$] \\
Qwen-Image-Edit & .481$\to$.305 & $-.175$ [$-.238$, $-.116$] & .626$\to$.634 & $+.008$ [$-.018$, $+.035$] \\
OmniGen2 & .667$\to$.622 & $-.045$ [$-.073$, $-.019$] & .583$\to$.627 & $+.044$ [$+.020$, $+.070$] \\
\bottomrule
\end{tabular}
\caption{\textbf{Object replacement on DreamEditBench (image approaches, $212$ pairs).} Cells show base$\to$\gap and the paired change.}
\label{tab:supp_object_image}
\end{table}

\gap lowers background LPIPS on all four approaches, and every interval excludes zero. Subject DINO rises on all four, significantly on all but Qwen-Image-Edit.

\paragraph{DreamSwapV.}
We also run all seven approaches on the $167$ subject instances of DreamSwapV-Benchmark~\citep{wang2026dreamswapv}. Metrics use every fourth frame. Subject DINO compares the object-box crop with the reference subject. Intervals are clustered by source video. \Cref{fig:supp_object} shows examples.

Subject DINO rises on six of the seven approaches, with intervals excluding zero (\cref{tab:supp_object_video}), and does not change significantly on Qwen-Image-Edit. On the three video approaches, background LPIPS rises slightly ($+.002$ to $+.006$; LoomVideo $.093\to.099$).

\begin{table}[H]\centering\scriptsize\setlength{\tabcolsep}{4pt}
\begin{tabular}{l|cc|cc}
\toprule
Approach & Subject DINO $\uparrow$ & $\Delta$ [95\% CI] & Bg.\ LPIPS $\downarrow$ & $\Delta$ [95\% CI] \\
\midrule
JoyAI-Video-Edit & .163$\to$.217 & $+.054$ [$+.039$, $+.070$] & .148$\to$.152 & $+.004$ [$+.002$, $+.005$] \\
LoomVideo & .150$\to$.164 & $+.014$ [$+.004$, $+.025$] & .093$\to$.099 & $+.006$ [$+.004$, $+.008$] \\
FLUX.2-klein & .295$\to$.329 & $+.033$ [$+.022$, $+.046$] & .061$\to$.061 & $+.000$ [$-.002$, $+.002$] \\
FLUX.2-klein-base & .398$\to$.463 & $+.065$ [$+.043$, $+.087$] & .104$\to$.109 & $+.005$ [$+.001$, $+.009$] \\
VACE & .281$\to$.305 & $+.025$ [$+.004$, $+.045$] & .067$\to$.068 & $+.002$ [$+.000$, $+.003$] \\
Qwen-Image-Edit & .276$\to$.266 & $-.009$ [$-.038$, $+.014$] & .067$\to$.058 & $-.010$ [$-.020$, $-.001$] \\
OmniGen2 & .246$\to$.379 & $+.133$ [$+.093$, $+.172$] & .300$\to$.282 & $-.017$ [$-.034$, $-.002$] \\
\bottomrule
\end{tabular}
\caption{\textbf{Object replacement on DreamSwapV-Benchmark ($167$ subject instances).} Cells show base$\to$\gap and the paired change; $\Delta$ is computed before rounding. Values are not comparable with \cref{tab:supp_object_image}, which uses a different benchmark and sample subset.}
\label{tab:supp_object_video}
\end{table}

\subsection{Multi-person head swapping}
\label{app:multiperson}

We evaluate on $97$ real multi-person clips cropped from $92$ VFHQ videos~\citep{xie2022vfhq}, using face detection to select segments with multiple people, and swap one head per clip. The prompts of both base and \gap state the position of the target person, so the base also knows which head to swap; VACE follows the mask without a prompt change. Target ID compares the target person in the output with the ground-truth identity; an output with no face detected at the target position scores $0$, which lowers the absolute values. Non-target ID compares each other person with their own face in the input. LPIPS$_{\mathrm{ne}}$ is measured outside the target head. \Cref{fig:supp_multiperson} shows examples.

Target ID rises on all seven approaches, with intervals excluding zero on six; on Qwen-Image-Edit the lower bound is $-.000$ (\cref{tab:supp_multiperson}). Non-target ID rises on FLUX.2-klein, FLUX.2-klein-base, and OmniGen2, changes by at most $.004$ on JoyAI-Video-Edit, LoomVideo, and VACE, and drops by $.026$ on Qwen-Image-Edit. LPIPS$_{\mathrm{ne}}$ rises by $.0052$ on JoyAI-Video-Edit, just above the $.005$ preservation threshold, and by at most $.0022$ elsewhere. JoyAI-Video-Edit has low absolute target ID ($.034\to.092$) because it rarely replaces the head in these scenes.

\begin{table}[H]\centering\scriptsize\setlength{\tabcolsep}{5pt}
\begin{tabular}{l|cc|cc}
\toprule
Approach & Target ID $\uparrow$ & $\Delta$ [95\% CI] & $\Delta$Non-target ID $\uparrow$ & $\Delta$LPIPS$_{\mathrm{ne}}$ $\downarrow$ \\
\midrule
JoyAI-Video-Edit & .034$\to$.092 & $+.058$ [$+.041$, $+.076$] & $-.004$ & $+.0052$ \\
LoomVideo & .138$\to$.193 & $+.055$ [$+.032$, $+.080$] & $-.002$ & $+.0022$ \\
FLUX.2-klein & .522$\to$.597 & $+.075$ [$+.058$, $+.094$] & $+.021$ & $-.0008$ \\
FLUX.2-klein-base & .640$\to$.707 & $+.067$ [$+.020$, $+.117$] & $+.031$ & $-.0046$ \\
VACE & .083$\to$.232 & $+.150$ [$+.111$, $+.191$] & $-.002$ & $+.0003$ \\
Qwen-Image-Edit & .121$\to$.173 & $+.052$ [$-.000$, $+.104$] & $-.026$ [$-.055$, $-.004$] & $+.0002$ \\
OmniGen2 & .091$\to$.374 & $+.282$ [$+.221$, $+.344$] & $+.088$ & $-.0840$ \\
\bottomrule
\end{tabular}
\caption{\textbf{Multi-person head swapping ($97$ real clips, $92$ source videos).} Both arms' prompts state the target position. Cells show base$\to$\gap or the paired change.}
\label{tab:supp_multiperson}
\end{table}

On Qwen-Image-Edit, non-target identity changes by $-.026$ [$-.055$, $-.004$]. Of the $95$ clips with a non-target face, $6$ drop by more than $.1$ and $2$ rise by more than $.1$; in most of the largest drops the editor redraws a non-target person while leaving the target unchanged. With a prompt that does not name the target's position, the change is $-.078$ [$-.127$, $-.036$], so it shrinks when the person to edit is specified. \gap changes reference attention only for the given mask and does not determine which person the editor chooses to redraw.

\subsection{Comparison with SaaS and KV-Edit}
\label{app:saas_kv}

\paragraph{Setup.}
SaaS~\citep{zhou2025saas} rescales attention to a text sub-instruction inside its region so that it matches attention to the input image. We adapt it to reference-conditioned editing by treating the reference image as the under-attended condition. For queries inside the edit mask (the same mask used by \gap), we add $\ln(\xi\alpha)$ to the logits of the reference tokens, where $\alpha$ is the ratio of attention mass on input-image tokens to that on reference tokens, pooled over the steered layers. Following the official implementation, step $0$ only collects statistics, and the bias is applied from step $1$ through the first $40\%$ of the denoising steps. Our port differs from the official code in three ways: it steers the second half of each approach's blocks (official: layers $8$--$31$ of OmniGen's $32$), it uses our edit mask instead of an Otsu-thresholded attention mask, and $\alpha$ pools the mass over all earlier steps instead of only the previous step. For JoyAI-Video-Edit only step $1$ is biased, so the last difference does not arise. No keep-side correction is applied. VACE and LoomVideo provide no separate group of input-image tokens, so the rule is undefined for them.

We also select $\xi$ with the rule used for $(\gamma_{\mathrm{edit}},\gamma_{\mathrm{keep}})$ (\cref{app:moved}): on the 40-clip validation split of the three development approaches where the rule applies, we take the $\xi\in\{1,2,4,8\}$ with the highest mean identity among those keeping $\Delta\mathrm{LPIPS}_{\mathrm{ne}}\leq0.005$ on all three. $\xi=4$ exceeds the criterion on FLUX.2-klein-base ($+.015$) and $\xi=8$ on JoyAI-Video-Edit ($+.0084$), so $\xi=2$ is selected (mean identity $.706$ versus $.677$ at $\xi=1$).

KV-Edit~\citep{zhu2025kvedit} replaces the keys and values of keep-region tokens with those of the source. We evaluate it alone and combined with the edit side of \gap. All results use the $1{,}040$ head-swap test clips.

\begin{table}[H]\centering\scriptsize\setlength{\tabcolsep}{4pt}
\begin{tabular}{l|ccccc}
\toprule
Setting & JoyAI-Video-Edit$^\ast$ & FLUX.2-klein & FLUX.2-klein-base & Qwen-Image-Edit & OmniGen2 \\
\midrule
Base & .507 / .075 & .726 / .090 & .588 / .158 & .609 / .289 & .207 / .262 \\
SaaS rule, $\xi=1$ (default) & .560 / .075 & .723 / .090 & .570 / .155 & .587 / .288 & .214 / .263 \\
SaaS rule, $\xi=2$ (selected) & .578 / .077 & .740 / .090 & .602 / .160 & .658 / .290 & .286 / .264 \\
\gap, edit side only & --- & .783 / .100 & .776 / .293 & .802 / .338 & --- \\
\gap & .644 / .079 & .790 / .071 & .799 / .087 & .723 / .075 & .693 / .192 \\
\midrule
KV-Edit & .508 / .067 & .752 / .036 & .632 / .042 & .426 / .041 & .218 / .052 \\
KV-Edit background + \gap edit side & .642 / .078 & .791 / .040 & .799 / .061 & .776 / .045 & .706 / .059 \\
\bottomrule
\end{tabular}
\caption{\textbf{Comparison with SaaS and KV-Edit on head swapping ($1{,}040$ test clips).} Cells show ID $\uparrow$ / LPIPS$_{\mathrm{ne}}$ $\downarrow$. $^\ast$~All JoyAI-Video-Edit arms run at $504^2$ on $1{,}040$ clips.}
\label{tab:supp_saas_kv}
\end{table}

\paragraph{Results: SaaS.}
\Cref{tab:supp_saas_kv} reports all arms. At its default $\xi=1$, the SaaS rule changes identity by $-.022$ to $+.053$ and keeps LPIPS$_{\mathrm{ne}}$ within $.003$ of the base; on Qwen-Image-Edit it lowers identity ($-.022$ [$-.043$, $-.003$]). With the selected $\xi=2$, it raises identity by $+.014$ to $+.079$, again within $.003$ of the base LPIPS$_{\mathrm{ne}}$, but stays $.051$ to $.407$ below \gap\ (all 95\% CIs exclude $0$). \gap raises identity by $+.064$ to $+.486$ on the same approaches. Without any preservation mechanism, the edit side of \gap alone raises LPIPS$_{\mathrm{ne}}$ on all three approaches where it was run.

\paragraph{Results: KV-Edit.}
KV-Edit locks the background: at every step it resets background latents to the noised source and replaces background keys and values with cached source ones, so its LPIPS$_{\mathrm{ne}}$ approaches that of pasting the source back. Alone, it keeps the original head on Qwen-Image-Edit, where identity drops by $.183$ [$.154$, $.211$]. Where the base rewrites the background (FLUX.2-klein, FLUX.2-klein-base, Qwen-Image-Edit, OmniGen2), combining it with the edit side of \gap matches or exceeds the identity of \gap at lower LPIPS$_{\mathrm{ne}}$, so there it can replace the keep side. This replacement has costs:
\begin{itemize}
\item It doubles the forward passes per step.
\item It requires the output to be token-aligned with the input, which excludes LoomVideo, whose output is resampled to $24$\,fps.
\item It adds nothing where the approach already preserves the background (JoyAI-Video-Edit: LPIPS$_{\mathrm{ne}}$ $.078$ versus $.079$); VACE, which conditions on the masked source, was not run.
\item It fixes all content outside the mask, which rules out edits that must change it.
\item On Qwen-Image-Edit, it raises keypoint error ($.213$ versus $.179$) and changes expression similarity by $-.017$ [$-.023$, $-.011$].
\end{itemize}

\subsection{Measured mass and required strength}
\label{app:mass_strength}

\begin{figure*}[t]\centering
\includegraphics[width=0.8\textwidth]{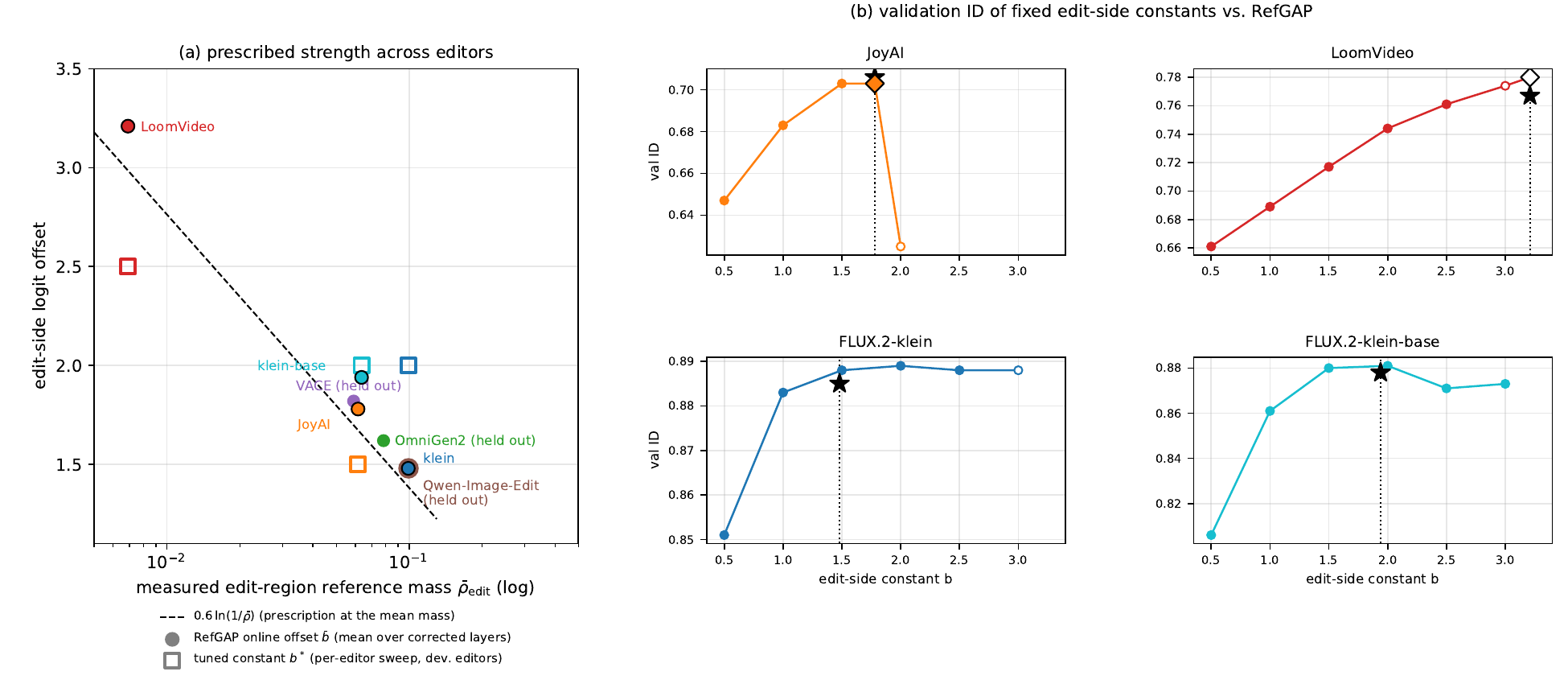}
\caption{\textbf{Measured reference mass and edit-side strength.} (a)~Edit-side offset against the measured edit-region reference mass. Filled circles: \gap's online offset $\bar b$, averaged over corrected layers. Open squares: the per-approach tuned constant $b^\ast$ (development approaches only). Dashed line: $0.6\ln(1/\bar\rho)$. (b)~Validation identity of a fixed edit-side constant $b$ on each development approach, with the same layers and online keep-side correction; open markers violate $\Delta\mathrm{LPIPS}_{\mathrm{ne}}\leq.005$. The star marks \gap, placed at its mean offset $\bar b$ (dotted line); the diamond marks a constant set to $\bar b$ (JoyAI and LoomVideo).}
\label{fig:mass_strength}
\end{figure*}

\Cref{fig:mass_strength}a shows that the offset \gap derives online differs across approaches, from $1.48$ on FLUX.2-klein and Qwen-Image-Edit to $3.21$ on LoomVideo. LoomVideo has the lowest measured mass ($\bar\rho_{\mathrm{edit}}=.0069$) and also needs the largest per-approach tuned constant ($2.5$, \cref{app:fixedb}).

\Cref{fig:mass_strength}b places \gap on each development approach's fixed edit-side sweep (\cref{tab:supp_matrix}). Without a sweep, \gap's validation identity is within $.004$ of the best admissible constant on the two FLUX.2 variants and above it on JoyAI-Video-Edit ($.706$ versus $.703$) and LoomVideo ($.767$ versus $.761$).

A constant set to \gap's mean offset is not equivalent. On JoyAI-Video-Edit ($b=1.78$) it gives identity $.703$ and $\Delta\mathrm{LPIPS}_{\mathrm{ne}}=-.0009$, versus $.706$ and $-.0022$ for \gap. On LoomVideo ($b=3.21$) it reaches $.780$ but raises $\Delta\mathrm{LPIPS}_{\mathrm{ne}}$ to $+.018$, violating the preservation criterion, whereas \gap gives $.767$ at $+.003$. How the online offset is distributed over layers and steps therefore matters, not only its mean.

\section{Qualitative results}
\label{app:showcase}

\begin{figure}[H]
\centering
\includegraphics[width=0.9\textwidth]{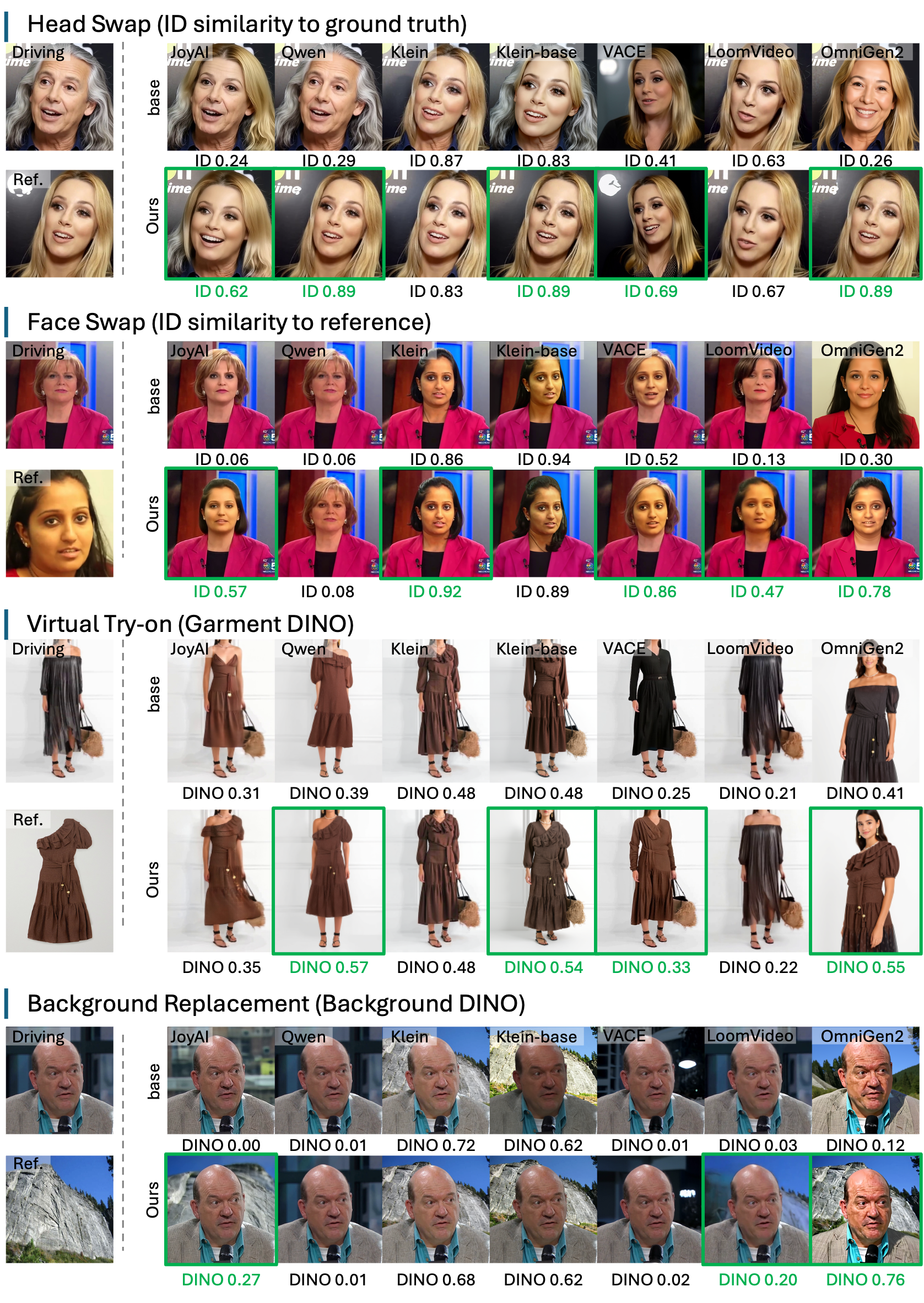}
\caption{\textbf{Complete seven-approach comparison for the main-paper examples.} Each block shows the input, reference, and base/\gap outputs, with reference-fidelity scores beneath. Green borders indicate gains $\geq 0.05$.}
\label{fig:supp_whole}
\end{figure}

\begin{figure}[H]
\centering
\includegraphics[width=\textwidth]{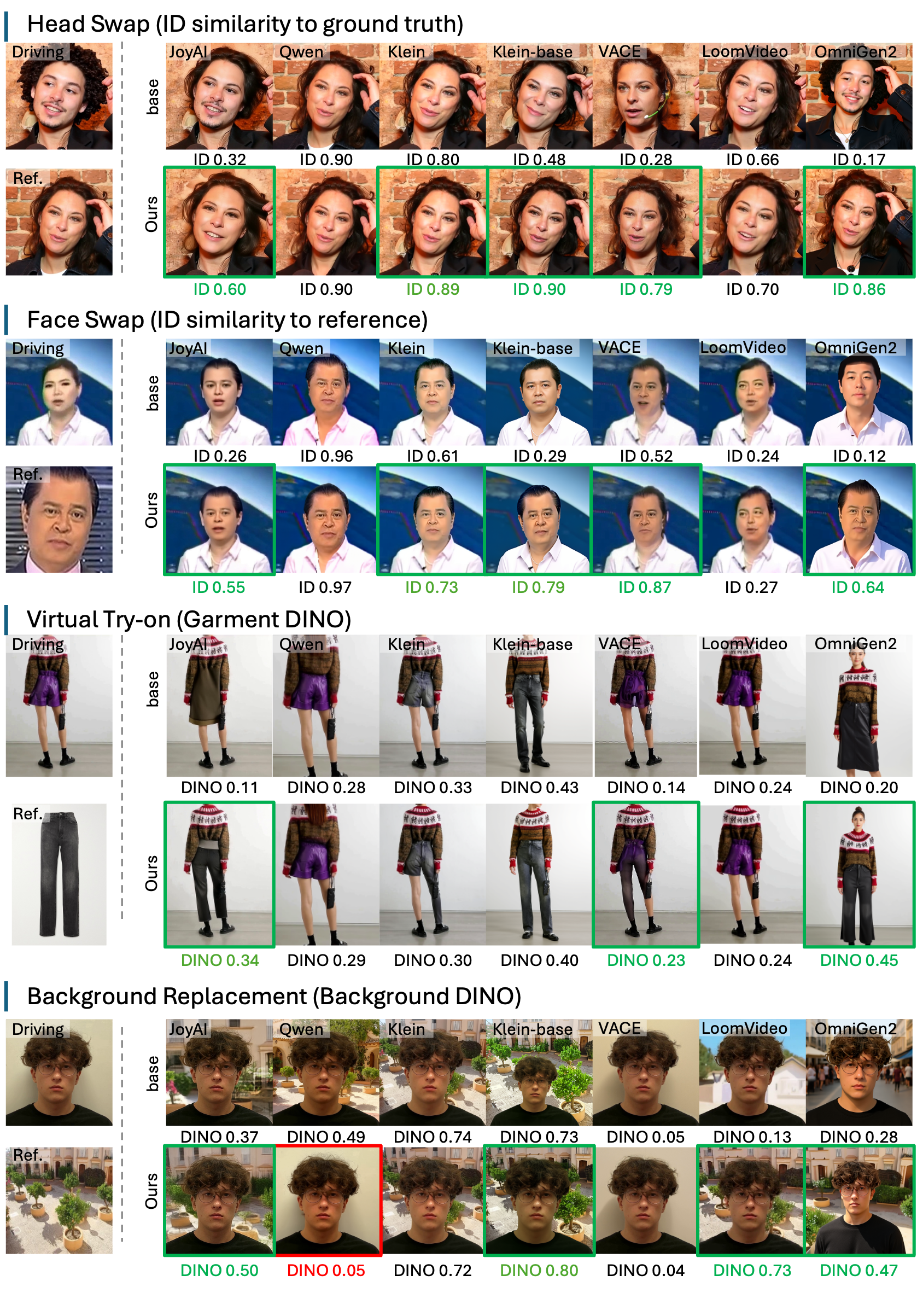}
\caption{\textbf{Additional qualitative examples across seven approaches and four tasks.} The layout and annotations follow \cref{fig:supp_whole}.}
\label{fig:more_example}
\end{figure}

\begin{figure}[H]
\centering
\includegraphics[width=\textwidth]{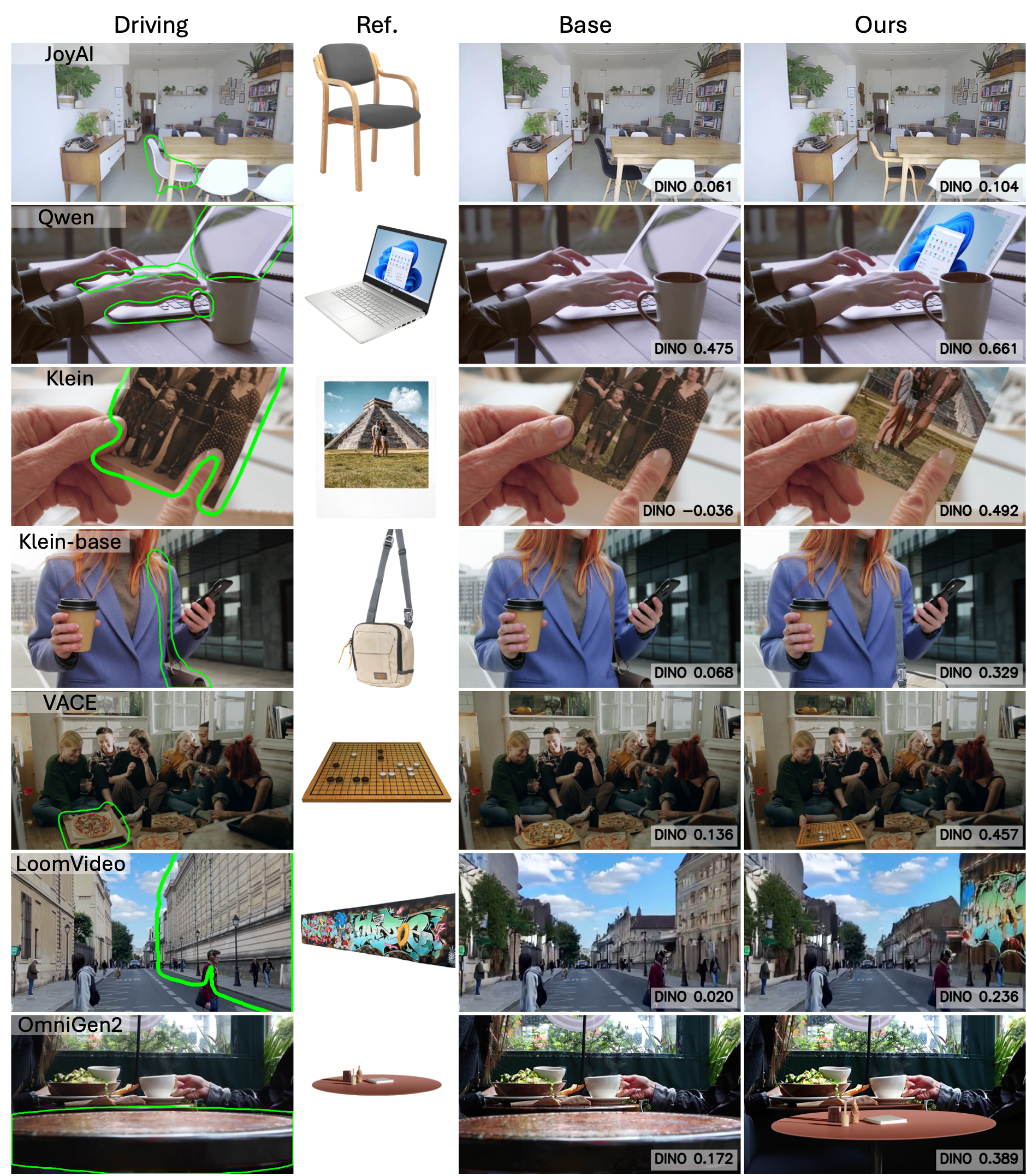}
\caption{\textbf{Object replacement on DreamSwapV-Benchmark across seven approaches.} Columns show the driving frame with the source-object mask (green), the reference subject, and base/\gap outputs, with subject DINO on each output.}
\label{fig:supp_object}
\end{figure}

\begin{figure}[H]
\centering
\includegraphics[width=\textwidth]{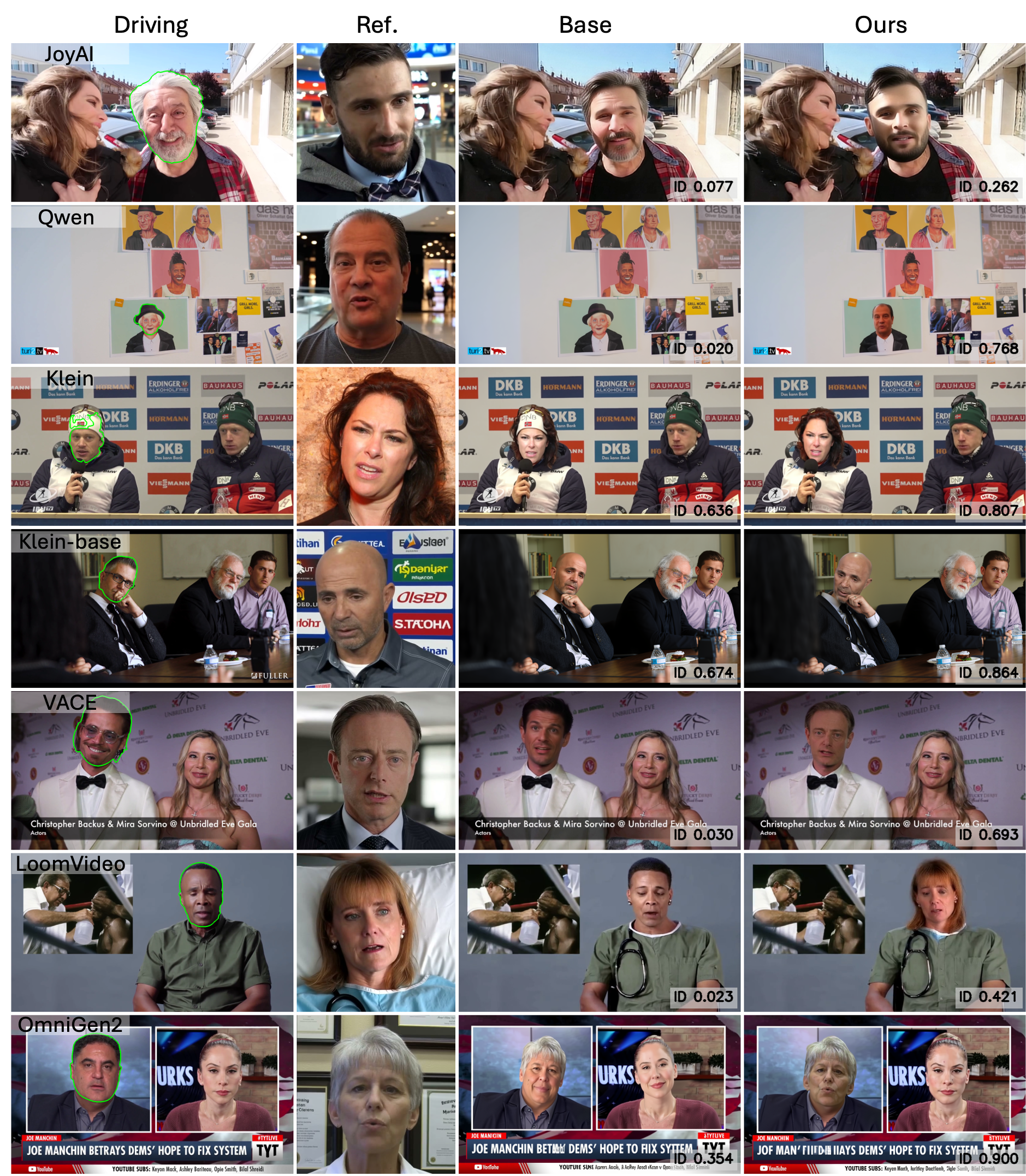}
\caption{\textbf{Multi-person head swapping across seven approaches.} Columns show the driving frame with the target-head mask (green), the reference, and base/\gap outputs, with target ID on each output. Only the outlined person is swapped.}
\label{fig:supp_multiperson}
\end{figure}

\Cref{fig:supp_whole} extends the examples in \cref{fig:result} to all seven approaches. \Cref{fig:more_example} provides an additional set of examples across the same four tasks. \Cref{fig:supp_object,fig:supp_multiperson} show object replacement on DreamSwapV and multi-person head swapping (\cref{app:extra}).

\end{document}